\documentclass[11pt,a4paper]{article}

\usepackage[english]{babel}

\usepackage[letterpaper,top=2cm,bottom=2cm,left=3cm,right=3cm,marginparwidth=1.75cm]{geometry}

\usepackage{amsmath}
\usepackage{graphicx}
\usepackage[authoryear, round, sort&compress]{natbib}
\usepackage[dvipsnames]{xcolor} 
\usepackage[colorlinks=true, linkcolor=black, citecolor=Periwinkle, urlcolor=Periwinkle]{hyperref}

\usepackage{tabularx}
\usepackage{booktabs}
\usepackage{array}
\usepackage{multirow}
\usepackage{ragged2e}
\usepackage[most]{tcolorbox}
\usepackage{newtxtext, newtxmath}
\usepackage{tocloft}
\usepackage{setspace}
\usepackage{pifont}
\newcommand{\cmark}{\textcolor{teal}{\ding{51}}}
\newcommand{\xmark}{\textcolor{red!60}{\ding{55}}}

\tcbset{
  abstractstyle/.style={
    colback=blue!3!white,
    colframe=blue!20!white,
    arc=8pt,
    boxrule=0.5pt,
    left=16pt,
    right=16pt,
    top=14pt,
    bottom=14pt,
  }
}

\newtcolorbox{summarybox}[1]{
  colback=blue!5!white,
  coltitle=blue!75!black,
  fonttitle=\bfseries\footnotesize,
  fontupper=\footnotesize,
  arc=6pt,
  boxrule=0pt,
  enhanced,
  title=#1,
  attach title to upper,
  after title={\par\smallskip},
  left=6pt,
  right=6pt,
  top=6pt,
  bottom=6pt,
}

\usepackage[T1]{fontenc}
\usepackage[utf8]{inputenc}
\usepackage{lmodern}
\usepackage{authblk}
\usepackage{microtype}
\usepackage{mdframed}
\usepackage{fontawesome5}
\usepackage{amsmath}
\usepackage{tikz}
\usetikzlibrary{backgrounds}

\makeatletter
\renewcommand\AB@affilsep{\par}
\makeatother

\definecolor{abstractbg}{HTML}{E8F2FC}
\definecolor{abstractborder}{HTML}{B8CFE8}
\definecolor{linkblue}{HTML}{1A4A8A}
\definecolor{linkbg}{HTML}{D6E8F8}
\definecolor{titledark}{HTML}{0D1B3E}
\definecolor{subtlegray}{HTML}{5A6A8A}
\definecolor{metadark}{HTML}{3A4A6A}

\newmdenv[
  backgroundcolor=abstractbg,
  linecolor=abstractborder,
  linewidth=0.5pt,
  innerleftmargin=12pt, innerrightmargin=12pt,
  innertopmargin=10pt,  innerbottommargin=10pt,
  skipabove=6pt, skipbelow=0pt,
  roundcorner=4pt
]{abstractbox}

\newcommand{\ghlink}[1]{%
  \textcolor{linkblue}{\small\url{#1}}%
}

\title{
{\LARGE\bfseries\color{titledark}
\raisebox{-0.32cm}{\includegraphics[height=1.5cm]{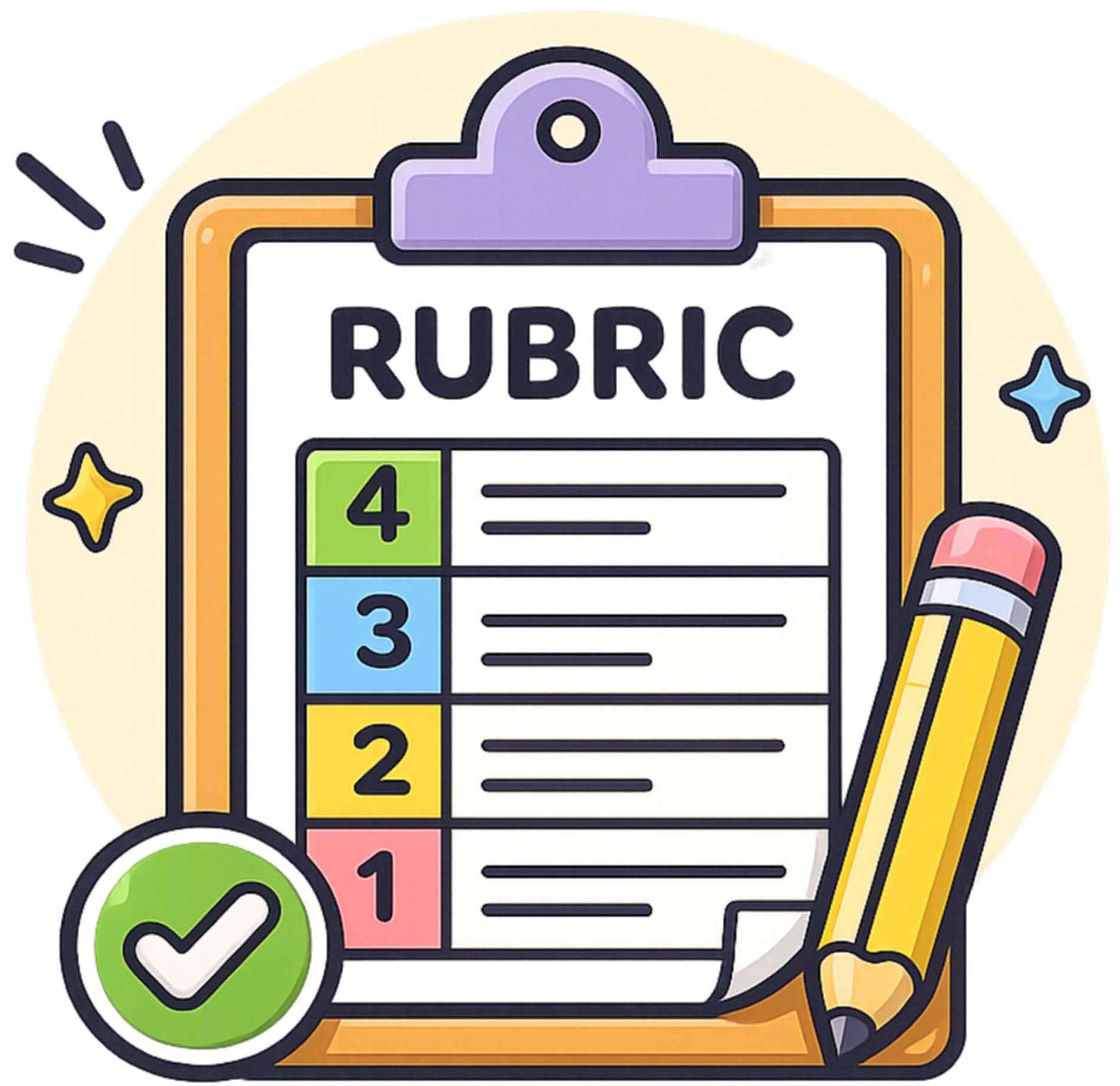}}
\hspace{-0.1cm}
From Holistic Evaluation to Structured Criteria:\\[4pt]
Rubrics Across the Evolving LLM Landscape}\\[8pt]
}

\author[1]{{\small\textbf{\color{titledark}Hao Chen}}}
\author[1]{{\small\textbf{\color{titledark}Ziyu Han}}}
\author[2,{\faEnvelope}]{{\small\textbf{\color{titledark}Yukun Yan}}}
\author[1]{{\small\textbf{\color{titledark}Qingfu Zhu}}}
\author[2]{{\small\textbf{\color{titledark}Maosong Sun}}}
\author[1,{\faEnvelope}]{{\small\textbf{\color{titledark}Wanxiang Che}}}

\affil[1]{{\small Research Center for Social Computing and Interactive Robotics,}}
\affil[ ]{{\small Harbin Institute of Technology}}
\affil[2]{{\small Department of Computer Science and Technology, Institute for AI,}}
\affil[ ]{{\small Tsinghua University}}
\date{}

\begin{document}
\maketitle
\vspace{-18pt}
\begin{tcolorbox}[abstractstyle]
\noindent
\textbf{Abstract:} As Large Language Models (LLMs) advance from task-specific systems toward open-ended autonomous agents, the mechanisms used to evaluate and guide their behavior must evolve accordingly. This work introduces the rubric as a unifying framework that systematically captures this evolution, characterizing rubrics as a dynamic response to successive LLM paradigm shifts, which recurs across otherwise independent efforts in evaluation, reinforcement learning, and safety alignment. We define rubrics as explicit sets of criteria that transform complex quality judgments into structured and actionable standards, and demonstrate that their recurrence across these diverse research threads is not coincidental. We systematically organize and categorize dispersed research efforts across the LLM lifecycle under a unified perspective, comprehensively examining how rubrics manifest and evolve at each critical stage. Rubrics manifest at three progressively deeper levels of impact. At the evaluative level, they decompose holistic judgments into verifiable dimensions, enabling reliable and interpretable assessment. At the training level, they serve as dense feedback signals that provide process-level guidance where scalar reward models and outcome-based approaches fall short.  At the agentic-intrinsic level, rubrics are autonomously internalized and evolved by agents themselves, emerging as self-generated standards that drive open-ended capability growth rather than serving as externally imposed constraints. We further assess rubric reliability from multiple perspectives, covering generation quality, execution fidelity, theoretical constraints, and security threats, before overviewing rubric-based benchmarks and applications across diverse domains. By rendering assessment transparent and decomposable, rubrics serve as a consistent grounding mechanism that translates human value expectations into machine-learnable signals. We argue that as models grow more capable and autonomous, rubrics will serve as the enduring bridge between human intentions and machine behavior, underpinning the next generation of reliable and aligned AI systems.
\end{tcolorbox}

\vspace{10pt}
\noindent
{\small\faEnvelope\enspace\textbf{Corresponding Authors}}\\[4pt]
{\small\faCalendar[regular]\enspace\textbf{Date:}\enspace
  \textcolor{black}{May 31, 2026 (v1);\enspace July 1, 2026 (v2)}}\\[4pt]
{\small\faGithub\enspace\textbf{GitHub:}\enspace
  \ghlink{https://github.com/AI9Stars/LLM-Rubrics-Survey}}
\newpage
\tableofcontents
\newpage

\section{Introduction}
The rapid evolution of Large Language Models (LLMs)~\citep{singh2025gpt5,guo2025deepseekr1} has fundamentally expanded the scope of machine autonomy, moving beyond simple text generation toward complex autonomous reasoning and long-horizon agentic tasks~\citep{guo2024multiagent, wu2025agenticreasoning}. As these models become integrated into increasingly open-ended domains, the role of evaluation and supervision mechanisms has transitioned from a mere benchmarking tool to a critical component for ensuring reliability and alignment~\citep{kinniment2024evaluating}. However, this expansion of capability has exposed a critical gap in the current ecosystem, where the development of such mechanisms consistently trails behind the models they are intended to monitor~\citep{yehudai2026surveyevaluation}.

Existing approaches to model evaluation typically rely on scalar reward models or holistic LLM-as-a-judge frameworks~\citep{ouyang2022rlhf,bai2022constitutionalai,zheng2023mtbench}. These methods often collapse multifaceted qualitative judgments into coarse signals, failing to provide the granular feedback or interpretable reasoning necessary for targeted refinement~\citep{kim2024prometheus}. Such limitations become particularly severe in scenarios lacking deterministic ground truths, where traditional programmatic verification is unfeasible and outcome-oriented rewards remain blind to spurious intermediate logic~\citep{lightman2024let, yuan2025Miracle}. As model complexity grows, this gap becomes increasingly difficult to bridge. The central challenge for the field lies in deriving robust, fine-grained feedback signals for intricate behaviors that defy simple correctness metrics.

In response to these limitations, independent efforts in open-ended evaluation, reinforcement learning, and safety alignment have gradually converged on a shared design principle of making quality criteria explicit and structured~\citep{Hashemi_2024llmrubric, gunjal2025rar, mu2024RBR}. Recent methodologies have shifted toward decomposing high-level alignment into interpretable principles and pairing qualitative dimensions with structured reasoning paths to enhance reliability~\citep{lambert2024rewardbench, rezaei2025onlinerubric, li2026rubrichub}. Furthermore, both theoretical insights and empirical evidence suggest that decomposing instructions into independently verifiable checklist items consistently outperforms scalar reward models~\citep{viswanathan2025rlcf,zhang2026chasingtail}. These diverse efforts, while varying in application, point toward the unified concept of a rubric, defined as an explicit and fine-grained set of criteria that transforms complex quality judgments into structured and actionable standards. By rendering the basis of evaluation transparent and decomposable, rubrics facilitate both reliable assessment and targeted model improvement across increasingly sophisticated tasks.

Nonetheless, the significance of the rubric extends well beyond its role as a mere evaluation instrument. As LLMs evolve through successive paradigm shifts, the rubric manifests at three progressively deeper levels of impact. At the evaluative level, it transforms subjective, holistic judgments into fine-grained and verifiable criteria, enabling reliable and interpretable assessment~\citep{arora2025healthbench, akyrek2025prbench}. Moving to the supervisory level, the rubric functions as a dense training signal that overcomes the inherent limitations of verifiable rewards in open-ended domains, providing the process-level guidance that outcome-based approaches often lack~\citep{huang2025rubicon,mahmoud2026rewardhacking}. At its most profound as an agentic-intrinsic level, the rubric emerges dynamically from the model's own training behaviors, co-evolving with model capability to drive self-improvement rather than remaining an externally imposed constraint~\citep{li2026evolm}. This progression reveals that the recurrence of the rubric is not coincidental, reflecting its role as an evolving anchor that translates human value expectations into machine-learnable signals~\citep{bai2022traininghelpful}. As LLM capabilities expand, the rubric remains the consistent grounding mechanism across the full trajectory of LLM evolution.

To this end, this work provides a comprehensive and unified perspective on the systematic role of rubrics in governing and steering Large Language Model behavior. Our main contributions are summarized as follows.
\begin{itemize}
    \item We are the first to examine the evolution of rubrics through the lens of the co-evolution between rubrics and LLM paradigms, tracing their development across the stages of pretraining, reinforcement learning, reasoning, agentic, and self-evolving systems. This perspective reveals how rubrics progressively transform from evaluation instruments into supervisory signals and ultimately into endogenous mechanisms for self-improvement.
    \item We unify dispersed research efforts across the LLM lifecycle under a single coherent framework, comprehensively examining how rubrics manifest at each critical stage from construction to deployment, while rigorously analyzing their reliability from multiple perspectives, ranging from practical failure modes to fundamental theoretical constraints.
    \item We provide the most comprehensive overview of LLM rubrics to date, including a systematic synthesis of rubric-based benchmarks and real-world applications across diverse domains, an analysis of emerging trends, and a research roadmap toward scalable supervision and self-evolving intelligent systems.
\end{itemize}

\begin{figure}[t]
\centering
\includegraphics[width=\linewidth]{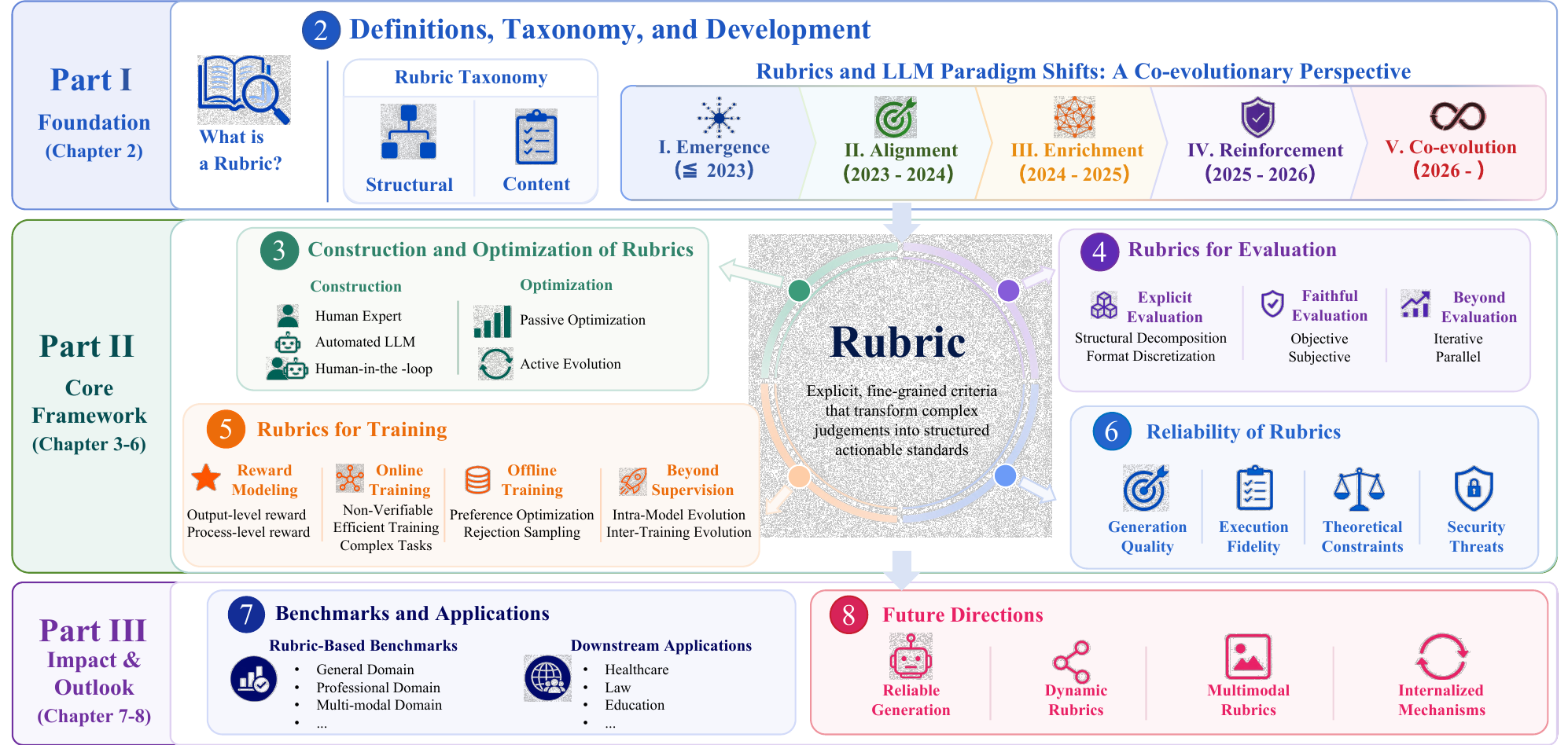}
\caption{Conceptual Organization of This Survey.}
\label{fig:intro}
\end{figure}

As illustrated in Figure~\ref{fig:intro}, this work is organized into three interconnected parts. Part I (§~\ref{section2}) establishes the conceptual foundations of rubrics by introducing their definitions, taxonomy, and developmental trajectories. Part II (§~\ref{section3}--§~\ref{section6}) presents the core methodological framework of rubric research, covering rubric construction and optimization, rubric-based evaluation, rubric-driven training, and reliability analysis from multiple perspectives. Part III (§~\ref{section7}--§~\ref{section8}) highlights rubric-based benchmarks and downstream applications, and concludes by discussing challenges and future research directions.

\section{What Is a Rubric? Definitions, Taxonomy, and Paradigm Shifts}\label{section2}
LLM evaluation demands criteria that are simultaneously explicit, decomposable, and reproducible, properties that neither scalar reward models nor unstructured human feedback can provide. Rubrics address this gap by operationalizing implicit quality judgments into structured, independently verifiable standards that can be systematically applied and consistently reproduced across evaluators.

This chapter formalizes the concept of rubrics in the LLM context through three lenses: a definitional framework grounded in four core properties (§~\ref{2.1Definition}), a two-dimensional taxonomy organizing rubrics by structural form and evaluative content (§~\ref{2.2Taxonomy}), and a historical analysis offering a co-evolutionary perspective on how successive LLM paradigm shifts have shaped the development of rubrics (§~\ref{2.3Co-evolutionary}).

\subsection{Definition}\label{2.1Definition}
The concept of rubric, originating from educational measurement, serves to operationalize implicit quality judgments into structured, reproducible, and interpretable evaluation criteria~\citep{brookhart2013create}. In the LLM landscape, while the objects of evaluation have expanded to model-generated text, code, and reasoning chains , and evaluators have transitioned to LLM-as-Judge systems , the core logic remains constant: transforming implicit intent into explicit and operationalized protocols. 

\begin{figure}[htbp]
\centering
\includegraphics[width=0.85\linewidth]{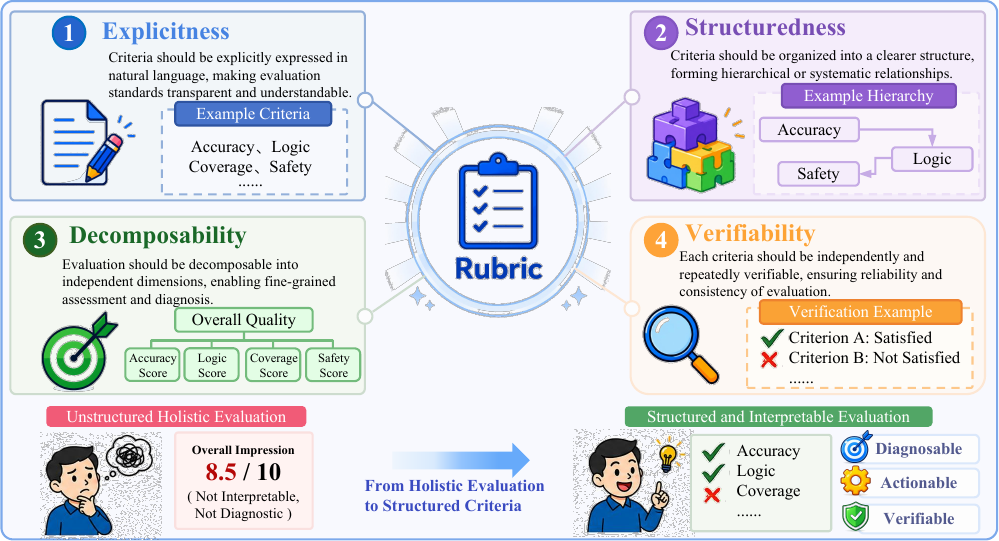}
\caption{Key Properties of an Effective Rubric.}
\label{fig:Definition}
\end{figure}

As shown in Figure~\ref{fig:Definition}, we define a rubric in the LLM context as a structured set of explicit criteria for assessing model outputs, characterized by four core properties: \textbf{explicitness} (criteria are articulated in natural language), \textbf{structuredness} (criteria are organized through explicit relationships), \textbf{decomposability} (criteria are partitioned into mutually independent units), and \textbf{verifiability} (criteria are designed for independent and reproducible evaluation). These properties collectively distinguish rubrics from holistic scalar rewards and unstructured natural language feedback.

\subsection{Taxonomy}\label{2.2Taxonomy}
Rubrics vary considerably in both form and evaluative foundation, reflecting the diversity of tasks, models, and evaluation goals they are designed to serve. To bring order to this landscape, we organize existing rubric designs along two complementary axes: structural taxonomy, which captures how rubrics are formally organized, and content taxonomy, which captures what they evaluate.

\subsubsection{Structural Taxonomy}
Rubrics exhibit structural variations across two orthogonal dimensions: the evaluation level and the evaluation granularity. The former defines the stage of the model's output being assessed, while the latter determines the resolution of the scoring criteria. These dimensions are detailed below and summarized in Table~\ref{tab:rubric_taxonomy}.

\textbf{Evaluation level.} The divergence in evaluation levels aligns with the ongoing debate between Outcome Reward Models (ORM) and Process Reward Models (PRM)~\citep{lightman2024let}. \textit{Output-level rubrics} direct assessment toward the model's final response~\citep{liu2023geval, arora2025healthbench}. Conversely, \textit{process-level rubrics} delve into intermediate reasoning steps, evaluating the quality of each individual derivation or action~\citep{jia2026autorubricr1v, chen2026rmr1}. The motivation for process-level rubrics lies in detecting “spurious reasoning”, where a model arrive at a correct answer through a logically flawed trajectory~\citep{yuan2025Miracle}.

\begin{table}[h]
\centering
\footnotesize
\caption{Structural taxonomy of rubrics.}
\label{tab:rubric_taxonomy}
\begin{tabularx}{\columnwidth}{>{\bfseries}l 
  >{\RaggedRight\arraybackslash\hsize=0.9\hsize}X 
  >{\RaggedRight\arraybackslash\hsize=1.1\hsize}X 
  >{\RaggedRight\arraybackslash\hsize=1.0\hsize}X}
\toprule
 & \textbf{Holistic} & \textbf{Analytic} & \textbf{Atomic} \\
\midrule
Output-level
  & MT-Bench~(\citeyear{zheng2023mtbench}), AlpacaEval~(\citeyear{dubois2025alpacaeval}), etc.
  & G-Eval~(\citeyear{liu2023geval}), PRBench~(\citeyear{akyrek2025prbench}), CDRRM~(\citeyear{liu2026cdrrm}), etc.
  & RBR~(\citeyear{mu2024RBR}), TICK~(\citeyear{cook2024tick}), HealthBench~(\citeyear{arora2025healthbench}), RLCF~(\citeyear{viswanathan2025rlcf}), etc. \\
\addlinespace[0.3em]
Process-level
  & Let's Verify Step by Step~(\citeyear{lightman2024let}), Math-Shepherd~(\citeyear{wang2024mathshepherd}), etc.
  & RM-R1~(\citeyear{chen2026rmr1}), AdaRubric~(\citeyear{ding2026adarubric}), Critic Rubrics~(\citeyear{wang2026criticrubric}), etc.
  & AutoRubric-R1V~(\citeyear{jia2026autorubricr1v}), RLCER~(\citeyear{sheng2026RLCER}), CaRR~(\citeyear{zhang2026CaRR}), etc. \\
\bottomrule
\end{tabularx}
\end{table}

\textbf{Evaluation granularity.} Drawing on three granularity levels from educational assessment~\citep{hunter1996use, brookhart2018appropriate}, we extend and adapt this continuum to the LLM setting. \textit{Holistic rubrics} yield a single overall judgment without decomposing quality into sub-dimensions~\citep{zheng2023mtbench}. \textit{Analytic rubrics} decompose quality into independently scored dimensions, enabling dimension-level diagnosis and interpretable feedback~\citep{liu2023geval, fan2025sedareval}. \textit{Atomic rubrics} reduce each criterion to a minimal binary proposition, where the evaluator answers “whether” rather than “how much”, a level that emerges naturally where instruction-following and factual accuracy admit objective per-item verification~\citep{viswanathan2025rlcf}. Although holistic, analytic, and atomic approaches are both traditionally regarded as rubric forms in educational assessment, this work primarily focuses on the latter two levels. We view holistic rubrics as a conceptual precursor to rubric-based evaluation rather than a fully structured set of assessment criteria.

\subsubsection{Content Taxonomy}
While structural coordinates define the form of a rubric, they do not distinguish their evaluative foundations. \textit{G-Eval}, \textit{CDRRM}, and \textit{PRBench} are all $Output \times Analytic$, yet they ground their criteria in task instructions, model outputs, and external knowledge, respectively. We therefore complement the structural taxonomy with a content taxonomy organizing rubrics by their evaluative anchor into three classes: Task-Grounded, Behavior-Grounded and Knowledge-Grounded.

\textbf{Task-Grounded} rubrics are grounded in the requirements of the task itself, directly verifiable against the current task, and valid within its scope. Representative types include \textit{task-constraint} rubrics, which extract explicitly verifiable constraints from instructions~\citep{mu2024RBR, cook2024tick, viswanathan2025rlcf}; \textit{quality-dimension} rubrics, which assess multiple quality facets of task outputs on graded scales~\citep{liu2023geval, Hashemi_2024llmrubric, fan2025sedareval}, and \textit{implicit-expectation} rubrics capturing evaluation dimensions users expect but never explicitly state~\citep{wadhwa2025evalagent, sharma2025researchrubrics}.

\textbf{Behavior-Grounded} rubrics are constructed from or applied to the model's existing outputs and behaviors, with their validity depending on external observation or verification. Representative types include \textit{process-critical} rubrics that diagnose reasoning trajectories~\citep{wang2026criticrubric, fan2026agent-RRM}, \textit{error-inductive} rubrics distilled from historical failure patterns~\citep{wan2026DeepVerifier, sanders2026generatingdatadriven}, and \textit{comparative-discriminative} rubrics designed to distinguish differences between candidate responses~\citep{liu2026cdrrm, xie2026autorubric}.

\textbf{Knowledge-Grounded} rubrics anchor their criteria in human-predefined principles and domain-specific sources that remain consistent across diverse tasks. Representative types include \textit{fact-checking} rubrics cross-referencing response content against external knowledge bases~\citep{ma2025search-Gen-V, zhang2026CaRR}, and \textit{domain-norm} rubrics derived from established professional standards such as legal statutes or financial guidelines~\citep{akyrek2025prbench, lee2026LEGIT}.

\subsection{Rubrics and LLM Paradigm Shifts: A Co-evolutionary Perspective}\label{2.3Co-evolutionary}
\begin{figure}[t]
\centering
\includegraphics[width=\linewidth]{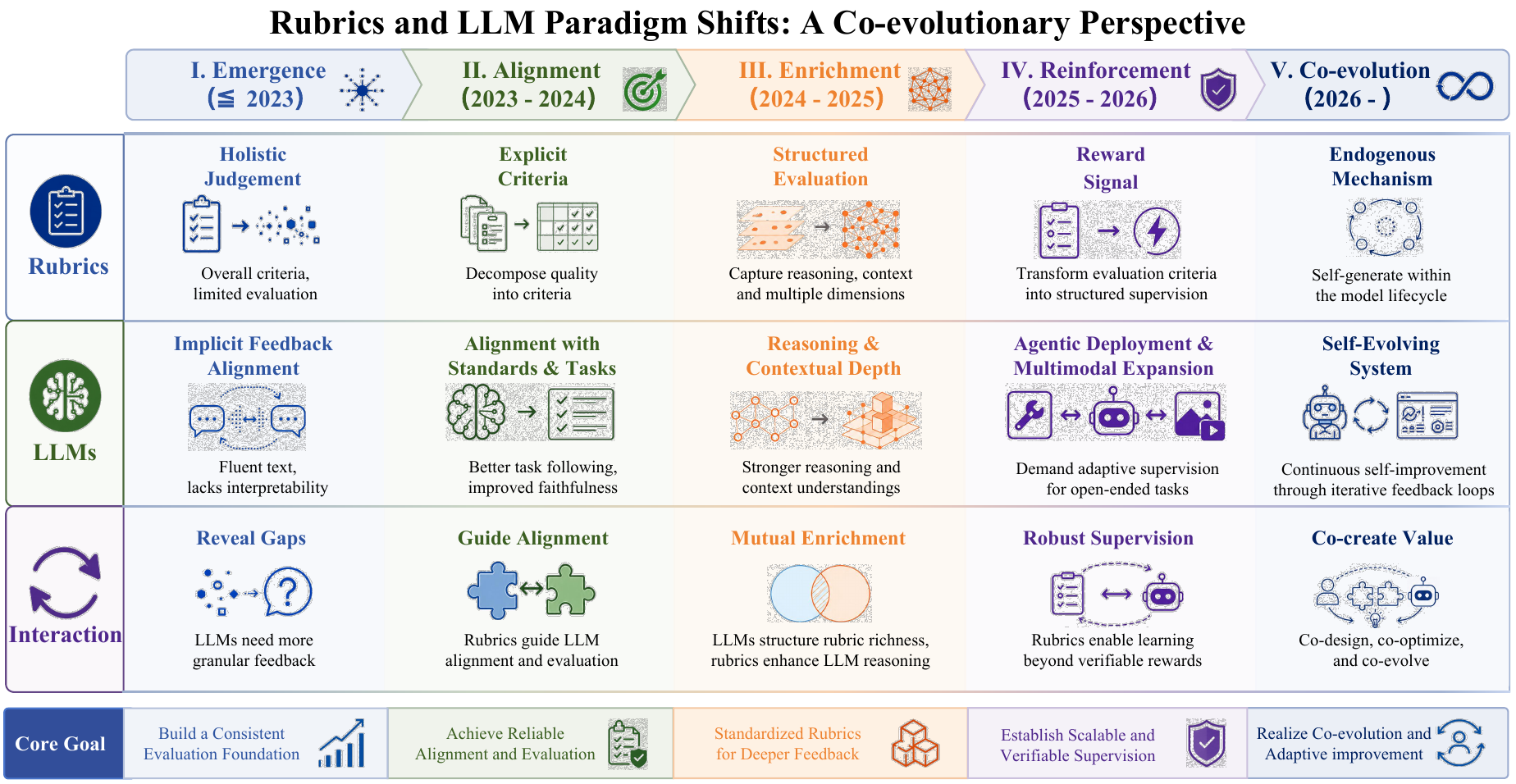}
\caption{A co-evolutionary framework illustrating the reciprocal development of rubrics and LLMs.}
\label{fig:Co-evolutionary}
\end{figure}

The evolution of rubrics has been tightly coupled with successive paradigm shifts in LLM development. As language models progressed from instruction following to reasoning, autonomous agents, and ultimately self-evolving systems, the demands on evaluation and supervision expanded accordingly. In response, rubrics continuously evolved from structured evaluation criteria to alignment guides, reasoning scaffolds, executable reward signals, and eventually endogenous supervisory mechanisms. As illustrated in Figure~\ref{fig:Co-evolutionary}, this progression forms a co-evolutionary feedback loop: advances in model capabilities drive the development of increasingly sophisticated rubrics, while richer rubric-based supervision, in turn, provides the evaluative and optimization foundations for the next generation of intelligent systems.

\textbf{Phase I (Up to 2023): From Holistic Judgments to Explicit Criteria.}
Early LLM development and evaluation largely relied on holistic judgments and scalar preference signals, where quality was treated as an indivisible outcome. Although RLHF~\citep{ouyang2022rlhf} demonstrated the effectiveness of preference-based supervision, scalar rewards often failed to capture the multifaceted nature of model quality and offered little insight into why a model succeeded or failed. Meanwhile, the rapid growth of reasoning~\citep{wei2023cot} and instruction-following tasks, together with increasingly capable evaluators such as GPT-4~\citep{openai2024gpt4}, exposed the limitations of coarse-grained assessment and created both the demand and the technical feasibility for structured evaluation. Consequently, rubrics evolved from implicit human standards into explicit and interpretable evaluation criteria, enabling more transparent, reproducible, and fine-grained assessment of model behavior. By decomposing complex objectives into multiple dimensions, rubrics also made evaluation outcomes easier to interpret, compare, and improve, establishing a structured interface between human expectations and model behavior. Representative efforts, including Constitutional AI~\citep{bai2022constitutionalai}, G-Eval~\citep{liu2023geval}, and MT-Bench~\citep{zheng2023mtbench}, progressively transformed previously tacit notions of quality into explicit evaluation dimensions and standardized assessment protocols. During this phase, rubrics primarily served as structured evaluation instruments rather than optimization objectives, laying the conceptual and methodological foundation for subsequent research.

\textbf{Phase II (2023--2024): Rubrics as Guides for Reliable Alignment.}
As LLMs expanded from closed-form benchmarks to increasingly open-ended scenarios, the central challenge shifted from producing fluent outputs to reliably aligning model behavior with human expectations. RLVR~\citep{shao2024rlvr} demonstrated the effectiveness of verifiable rewards in domains such as mathematics and coding, while simultaneously revealing a verification gap in open-ended tasks where correctness could not be automatically determined. This limitation transformed rubrics from evaluation instruments into alignment scaffolds, enabling complex objectives to be decomposed into explicit and interpretable supervisory criteria. Rather than merely assessing model outputs, rubrics increasingly served as executable specifications that guided both optimization and inference. Structured criteria became essential for emerging paradigms such as test-time scaling~\citep{jaech2024openaio1} and agentic workflows requiring fine-grained behavioral constraints~\citep{ma2024agentic}. At the same time, researchers began to recognize the limitations of static criteria. \citeauthor{gupta2025carmo} showed that fixed rubrics struggle to approximate complex reward functions, motivating richer and more adaptive rubric formulations, while analytic rubric frameworks demonstrated that multidimensional criteria could reliably approximate human judgments~\citep{Hashemi_2024llmrubric}. During this phase, rubrics evolved from evaluation criteria into alignment mechanisms, translating abstract human values into actionable supervisory signals for both optimization and inference.

\textbf{Phase III (2024--2025): Mutual Enrichment Between Rubrics and Reasoning.}
The emergence of reasoning-centric LLMs shifted evaluation from judging final answers toward assessing the reasoning process itself. As GRPO and large-scale reasoning models~\citep{guo2025deepseekr1} became increasingly prevalent, outcome-based supervision proved insufficient for evaluating long reasoning trajectories, contextual understanding, and intermediate decisions. Consequently, rubrics evolved into process-aware evaluation frameworks that decompose complex reasoning into explicit cognitive dimensions and provide interpretable feedback throughout the reasoning process~\citep{galvansosa2025rubrikscube}. Beyond evaluating reasoning, rubrics also began to actively support it by enabling structured feedback, reflection, and iterative refinement. Conversely, increasingly capable reasoning models facilitated the automatic generation, refinement, and adaptation of more sophisticated rubrics, creating a mutually reinforcing cycle between reasoning and evaluation. Representative benchmarks such as HealthBench~\citep{arora2025healthbench} modeled expert reasoning in specialized domains, while rubric-guided agents extended structured evaluation to long-form generation and writing tasks~\citep{wadhwa2025evalagent}. At the same time, systematic studies of scoring bias~\citep{Dineen_2025QALIGN,li2025leveraging} underscored that increasingly sophisticated rubrics must also be robust and trustworthy to serve as reliable supervision. This phase marked the emergence of a feedback-rich ecosystem in which reasoning continuously improved rubrics, while rubrics increasingly shaped reasoning itself.

\textbf{Phase IV (2025--2026): Rubrics as Reward Signals for Scalable Supervision.}
The rise of autonomous agents transformed rubrics from alignment guides into executable reward signals for scalable supervision. As LLMs increasingly operated in open-ended environments where explicit correctness signals were unavailable, rubric-based supervision emerged as a practical alternative to conventional reward modeling. To mitigate reward hacking, Checklist-as-Reward~\citep{viswanathan2025rlcf} decomposed complex objectives into verifiable criteria, while Rubicon~\citep{huang2025rubicon} scaled this paradigm through large repositories of reusable rubric-based rewards. Meanwhile, online rubric engineering frameworks enabled the automatic extraction and continual refinement of evaluation criteria from preference data~\citep{rezaei2025onlinerubric}. As rubric-based supervision matured, its applicability naturally expanded beyond text-only settings. The rapid emergence of multimodal foundation models~\citep{bai2025qwen25vl} exposed the limitations of outcome-based rewards and coarse-grained metrics, further motivating rubrics as structured reward interfaces that decompose multimodal capabilities into interpretable dimensions for fine-grained supervision~\citep{li2026ueval}. These advances enabled scalable supervision across increasingly diverse domains, including multimodal reasoning~\citep{jia2026autorubricr1v}, scientific discovery~\citep{goel2025aicoscientists}, and other agentic applications. During this phase, rubrics evolved into reusable and executable reward abstractions, enabling scalable supervision in environments where explicit reward functions are difficult or impossible to specify.

\textbf{Phase V (2026--): Endogenous Mechanisms and Self-Evolving Systems.}
The latest frontier moves beyond externally designed evaluation frameworks toward endogenous rubric generation within the model lifecycle itself. Rather than being manually authored, rubrics are increasingly generated, refined, and adapted through interactions among generation, evaluation, optimization, and deployment processes, forming closed-loop supervisory mechanisms. Recent work has extended this paradigm to unified cross-modal alignment~\citep{kong2026omni}, while simultaneously exposing the reliability limitations of machine-generated rubrics, including substantial degradation in rubric quality and evaluator consistency~\citep{zhang2026rubricbench}. These challenges have motivated growing interest in generator--evaluator co-evolution~\citep{xu2026rubricARM}, dynamic rubric adaptation, and autonomous governance mechanisms. At the same time, the identification of intrinsic rubric failures~\citep{qi2026rift} and self-preference biases~\citep{pombal2026selfpreferencebias} highlights that trustworthy self-supervision remains a fundamental challenge. Looking ahead, rubrics are expected to evolve from standalone evaluation artifacts into persistent supervisory infrastructure that spans the entire model lifecycle, supporting planning, reasoning, evaluation, optimization, and deployment within a unified framework. Rather than being periodically updated by human experts, future rubrics may continuously adapt to changing tasks, environments, and model capabilities through ongoing interactions with data, feedback, and experience. Such self-evolving mechanisms could ultimately enable autonomous closed-loop governance, where intelligent systems jointly improve both their behaviors and the criteria used to evaluate them, establishing rubrics as endogenous components for scalable, adaptive, and trustworthy AI systems.

Together, these phases reveal that the evolution of rubrics is not merely a consequence of advances in LLMs, but an essential driver of their continued progress. As language models evolve from instruction following to reasoning, autonomous agents, and self-improving systems, rubrics have continuously expanded their role—from explicit evaluation criteria and alignment guides to reasoning scaffolds, executable reward signals, and ultimately endogenous supervisory mechanisms. This co-evolution reflects a broader paradigm shift in which supervision itself becomes increasingly structured, adaptive, and self-evolving. Rather than serving solely as external evaluation tools, rubrics are emerging as foundational components of intelligent systems, shaping how future models are evaluated, optimized, and ultimately improved.

\begin{summarybox}{Summary}
This chapter establishes the conceptual foundations of rubrics from three complementary perspectives, leading to the following key conclusions.
\begin{itemize}
\item[\raisebox{-0.15ex}{\includegraphics[width=1.2em]{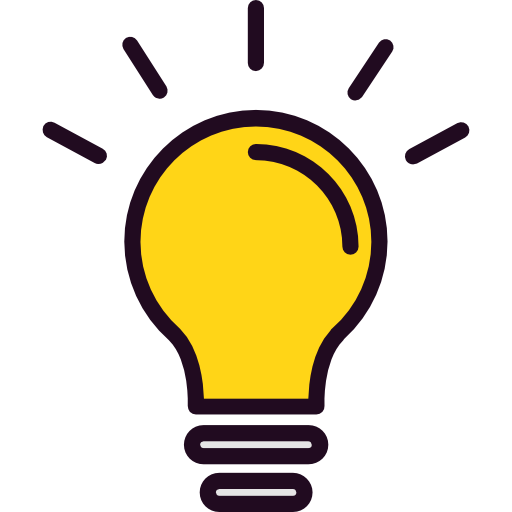}}]
Formally, a rubric is characterized by four defining properties: \emph{explicitness}, \emph{structuredness}, \emph{decomposability}, and \emph{verifiability}, which distinguish it from holistic evaluation.

\item[\raisebox{-0.15ex}{\includegraphics[width=1.2em]{image/lightbulb.png}}]
Taxonomically, rubrics can be organized along two structural dimensions and three content anchors, providing a unified framework for understanding and comparing diverse rubric designs.

\item[\raisebox{-0.15ex}{\includegraphics[width=1.2em]{image/lightbulb.png}}]
Historically, the evolution of rubrics reflects the co-evolution of supervision and model capabilities: as alignment challenges become increasingly complex, rubrics evolve from evaluation criteria to alignment guides, reasoning scaffolds, reward signals, and ultimately self-evolving supervisory mechanisms.
\end{itemize}
\end{summarybox}

Having established the conceptual foundations of rubrics, we now shift from what rubrics are to how they are developed and applied throughout the LLM lifecycle. As illustrated in Figure~\ref{fig:rubric-taxonomy-linked}, existing research can be organized into four tightly connected dimensions: rubric construction and optimization, rubric-powered evaluation, rubric-powered training, and rubric reliability. Together, these dimensions span the complete lifecycle of rubric-based systems, from rubric creation and refinement to their deployment, optimization, and validation. More importantly, they provide a unified perspective that connects research traditionally scattered across different stages of the LLM lifecycle. Viewed through this lens, rubrics emerge as a common abstraction underlying evaluation, alignment, training, and continual improvement, reflecting a deeper insight that translating human expectations into machine-executable criteria remains a central challenge throughout the evolution of LLMs.
\input{image/rubric_survey_taxonomy}
\section{How Are Rubrics Constructed and Optimized?}\label{section3}
Rubric construction and optimization constitute the foundation of the entire rubric lifecycle: a rubric must first be built and maintained at sufficient quality before it can serve evaluation or alignment purposes. This prerequisite is non-trivial, as empirical evidence demonstrates that low-quality rubrics do not merely fail to provide useful signals but can actively mislead reward models and degrade evaluation accuracy~\citep{shen2026RRD}.

This chapter addresses this across two layers: rubric construction (§~\ref{3.1Construction}), covering how human intent is translated into structured criteria across varying degrees of automation; and rubric optimization (§~\ref{3.2Optimization}), covering how rubrics improve from passive refinement driven by external signals to active co-evolution with model capabilities.

\subsection{Construction: How Rubrics Are Built}\label{3.1Construction}
\begin{figure}[t]
\centering
\includegraphics[width=\linewidth]{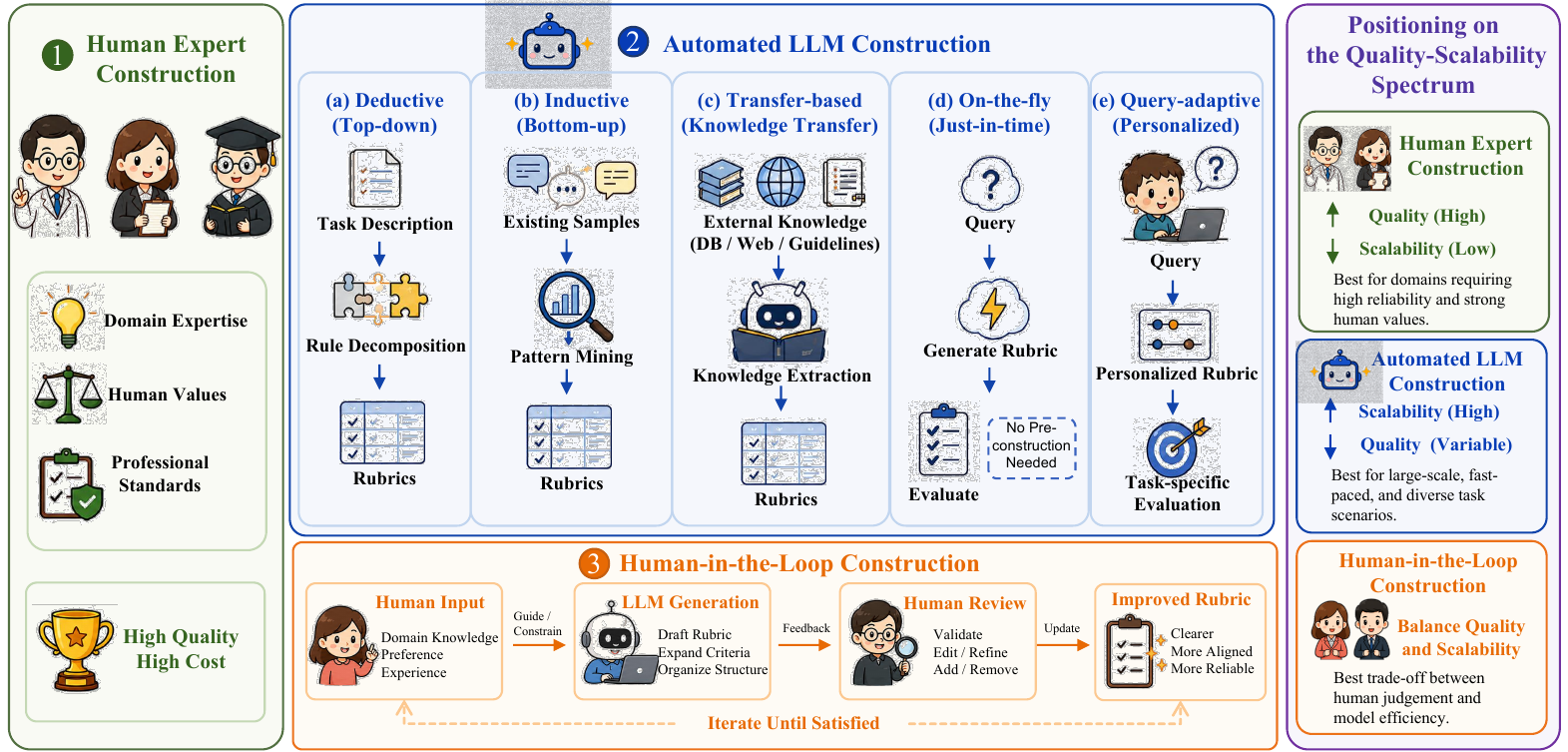}
\caption{Rubric Construction Paradigms and Their Positioning on the Quality–Scalability Spectrum.}
\label{fig:Construction}
\end{figure}

Rubric construction is fundamentally a process of knowledge externalization, transforming implicit evaluation standards into explicit, actionable criteria. As illustrated in Figure~\ref{fig:Construction}, existing approaches can be broadly categorized into three paradigms according to the degree of human involvement: human expert construction, automated LLM construction, and human-in-the-loop construction, each reflecting a different trade-off between rubric quality and scalability.

\subsubsection{Human Expert Construction}
Human expert construction provides irreplaceable domain priors and value alignment, and has been most systematically deployed in high-stakes professional domains. HealthBench~\citep{arora2025healthbench} engaged 262 physicians to write 48,562 criteria for 5,000 medical dialogues. In the legal and financial domains, PRBench~\citep{akyrek2025prbench} and ProfBench~\citep{wang2025profbench} provide large-scale expert-written criteria without LLM assistance. For broader professional coverage, XpertBench~\citep{liu2026xpertbench} provides  weighted checkpoints per task across seven domains, and ExpertLongBench~\citep{ruan2025expertbench} supplies task-specific rubrics for long-form generation tasks in law, clinical care and so on. Beyond task-level annotation, \citet{han2026aesrm} manually defines three orthogonal video aesthetics dimensions with 15 fine-grained criteria, transforming subjective aesthetic judgment into a programmatically verifiable framework. The fundamental limitation of this pathway is its prohibitive cost, which directly motivates the automated construction methods.

\subsubsection{Automated LLM Construction}
Automated construction methods aim to leverage LLMs' language understanding and generation capabilities to automatically extract or generate structured rubric, without large-scale human annotation. In this section, we organize existing work into five sub-pathways based on their underlying knowledge source and generation strategy.

\textbf{Deductive: Generating from Task Descriptions.} Deductive methods decompose task descriptions or meta-rules into fine-grained criteria through logical inference, differing primarily in the granularity and structure of decomposition. At the coarsest level, \citet{cook2024tick} and~\citet{viswanathan2025rlcf} directly extract flat checklists from user instructions, binding criteria naturally to each instruction. Going deeper, RubricHub~\citep{li2026rubrichub} introduces multi-model aggregation and a difficulty evolution mechanism to capture the fine-grained gap between good and excellent responses. Taking this to the extreme, Qworld~\citep{gao2026qworld} recursively expands each question into a hierarchical tree of binary criteria, ensuring every leaf node is an unambiguous judgment. At scale, ARES~\citep{li2026ares} co-generates instance-specific weighted rubrics alongside 
question-answer pairs directly from raw pretraining documents, ensuring each rubric is tailored to its associated question.

\textbf{Inductive: Extracting from Samples.} Inductive methods distill criteria from existing samples rather than task descriptions, and can be further grouped by their data source. In the \textit{trajectory aggregation} direction, AutoRubric-R1V~\citep{jia2026autorubricr1v} extracts common reasoning steps from multiple successful trajectories, naturally filtering out coincidental paths without human annotation; In the \textit{contrast-driven} direction, Auto-Rubric~\citep{xie2026autorubric}, OpenRubrics~\citep{liu2026openrubric}, CDRRM~\citep{liu2026cdrrm}, C2~\citep{kawabata2026c2}, and ROPD~\citep{fang2026rubricbasedonpolicy} each analyze differences between preferred and rejected responses to distill discriminative criteria, differing in whether they frame this as constrained optimization, contrastive generation, or closed-loop quality control. In the \textit{failure-reflection} direction, \citet{sanders2026generatingdatadriven} and~\citet{wan2026DeepVerifier} take the opposite stance, building negative error pattern libraries from historical failures rather than characterizing what good responses should contain. Beyond these response-level sources, RaR~\citep{gunjal2025rar} derives multi-category rubrics from reference answers; MIRA~\citep{wang2026mira} 
discovers source-specific rubrics through self-anchoring for mid-training data selection; and PARL~\citep{qiu2026parl} induces personalized rubrics from user interaction histories to capture stable individual preferences.

\textbf{Transfer-based: Migrating from External Knowledge.} Transfer-based methods mine implicit evaluation standards from existing external knowledge structures rather than task descriptions or sample data. From \textit{structured repositories}, ResearchQA~\citep{yifei2025researchqa} extracts rubric entries from academic survey papers, while~\citet{lee2026LEGIT} converts court judgment structures into hierarchical rubric checkpoints with built-in legal authority. From \textit{unstructured sources}, EvalAgent~\citep{wadhwa2025evalagent}, QuRL~\citep{wei2026qurl}, and TechImage-Bench~\citep{ni2026techimagebench} mine task-specific criteria from web documents, public resources, and multimodal context respectively. RubricRAG~\citep{dhole2026rubricrag} further takes a \textit{retrieval-augmented} approach, using existing rubrics from related queries as few-shot guidance for generating new criteria. Beyond static repositories, DR-rubric~\citep{mei2026DRRubric} and DeepRubric~\citep{zhu2026deeprubric} ground rubric construction in active retrieval: the former through iterative agentic search, the latter by back-synthesizing criteria from a multi-hop evidence tree.

\textbf{On-the-fly Generation: Generating Without Pre-construction.} Unlike the methods above that pre-build rubric libraries, on-the-fly generation produces task-specific criteria instantaneously at evaluation or training time. Some methods generate rubrics before scoring, using them as explicit reasoning anchors: \citet{chen2026rmr1} generates sample-level rubrics prior to judgment, while CARMO~\citep{gupta2025carmo} dynamically generates context-aware criteria for each query to ground the scoring process, and Think-with-Rubrics~\citep{yu2026thinkwithrubrics} internalizes rubric generation into the reasoning chain itself, having the model generate task-specific rubrics before producing a response so that quality standards become prior constraints on generation. Others generate rubrics as self-proposed verification standards: CaT~\citep{jayalath2026cat} derives binary auditable criteria from pseudo-reference answers, and~\citet{feng2025sage} locks generated rubrics as fixed standards across all answer pairs to mitigate context preference bias. DeltaRubric~\citep{liu2026deltarubric} extends this paradigm to multimodal settings, dynamically generating instance-specific rubrics that capture visual differences in spatial position and object attributes before verifying each criterion step by step. \citet{Dineen_2025QALIGN} takes a more structured approach, generating hierarchically gated evaluation programs from constitutional principles.

\textbf{Query-adaptive Generation: Input-specific Criterion Generation.} Query-adaptive methods occupy a middle ground between pre-construction and on-the-fly generation, customizing rubrics for each individual query rather than applying generic static standards. AdaRubric~\citep{ding2026adarubric} dynamically generates orthogonal scoring dimensions for each agent task. \citet{lv2026learningquery} trains a query-specific rubric generator via GRPO, demonstrating that RL is necessary for effective rubric generation on complex tasks. DR Tulu~\citep{shao2025drtulu} and \citet{huang2026PoP} further integrate rubric generation into online RL training, using variance filtering and pre-training conditioning to prevent proliferation and reward hacking respectively. However, QUBRIC~\citep{zhang2026qubric} identifies a deeper structural bottleneck: open-ended queries inherently produce under-constrained rubrics, and rubric quality cannot be improved in isolation without co-designing the query itself. Rubric-as-Experts~\citep{xu2026rubricasexpert} extends this adaptive principle to translation quality evaluation, generating case-specific MQM rubrics that replace generic error taxonomies with instance-adaptive standards.

\subsubsection{Human-in-the-Loop Construction}
Human-in-the-loop construction seeks a balance between the quality of manual construction and the efficiency of automated methods, focusing human input on the highest-value steps while delegating scalable execution to LLMs. The most direct instantiation is the “expert refinement, model scaling” paradigm: \citet{shi2025humanintheloop} propose LLMs generate draft rubrics that human experts refine once into gold standards, after which LLM judges scale up evaluation, finding that shared rubrics significantly improve inter-human agreement.

Beyond one-time refinement, human-in-the-loop methods vary in how they integrate human judgment. ARCANE~\citep{masters2025arcane} elicits implicit stakeholder preferences through dialogue and converts them into dynamically updatable rubrics, avoiding retraining when preferences shift. \citet{shah2026casespecific} take a validation-oriented approach, having clinicians author rubrics that are then verified by LLM scoring agents, achieving clinician-level agreement at a fraction of the cost. CLR-voyance~\citep{nagar2026clrvoyance} takes a data-driven stance, inducting rubric criteria directly from real patient outcome data rather than expert priors. RubricsTree~\citep{zhang2026rubricstree} similarly grounds criteria in real user queries, but adopts an evolutionary protocol where a physician-led panel iteratively refines atomic Boolean rubrics as clinical understanding deepens. ReviewGrounder~\citep{li2026reviewgrounder} instead diversifies the human input itself, synthesizing rubrics from official guidelines, paper content, and human review comments to balance authority, relevance, and practical experience.

\subsection{Optimization: How Rubrics Improve}\label{3.2Optimization}
Once a rubric has been constructed, continuously improving its quality is another central challenge. Existing optimization work falls into two categories: passive optimization and active evolution.

\subsubsection{Passive Optimization}
Passive optimization methods treat rubrics as optimization targets, driving rubric improvement by introducing external quality signals. Some methods approach this from the \textbf{generation perspective}: OptimSyn~\citep{fan2026Optimsyn} uses each synthetic sample's causal contribution to model improvement as a reward signal for rubric generation, transforming rubric design into a mathematically optimizable objective. Others focus on \textbf{diagnosing and repairing structural defects} in existing rubrics: \citet{chu2026CARO} treats scoring errors as directional clusters rather than random noise, identifying dominant patterns via confusion matrices and generating pattern-specific repair patches; \citet{shen2026RRD} recursively decompose overly coarse rubrics into finer sub-criteria while filtering out misdirected and redundant ones; SibylSense~\citep{xu2026SibylSense} further tackles the problem of rubric saturation by maintaining a memory bank that continuously replaces exhausted criteria, coupled with adversarial policy updates that actively probe for new quality dimensions; AMARIS~\citep{wu2026amarismemory} extends this memory-augmented paradigm by maintaining a persistent memory bank that accumulates diagnostic information across the entire training history, driving rubric evolution toward identifying the most critical unresolved failure patterns. A third line of work targets \textbf{human preference alignment}: \citet{qiu2026Proxy-GRM} train a lightweight proxy to predict preference rankings from rubrics alone, using prediction accuracy as a direct quality signal, while~\citet{harada2025Reflect-and-Revise} iteratively revise rubrics against human scoring mismatches, finding that even a minimal initial rubric converges to match manually written ones; SVR~\citep{sun2026SVR} reformulates rubric construction as a max-margin problem, mining contrastive features from preference pairs to close the gap between self-generated and human-authored rubrics; and Feedback-to-Rubrics~\citep{yoshida2026feedbacktorubrics} drives rubric refinement through annotation prediction errors, identifying coverage gaps by comparing rubric-conditioned predictions against expert inline comments.

\subsubsection{Active Evolution}
Active evolution fundamentally redefines the role of rubrics from static external standards to dynamic mechanisms deeply coupled with the training process, falling into two forms: rubric self-evolution and rubric-model co-evolution.

\textbf{Rubric Self-evolution.} In this direction, rubrics emerge from actual model or user behavior rather than pre-specified designs, with works differing in what drives the evolution. iRULER~\citep{bai2026iruler} grounds evolution in user writing practice, recursively applying rubrics to evaluate and refine rubrics themselves. CoReflect~\citep{li2026coreflect} and OnlineRubrics~\citep{rezaei2025onlinerubric} instead derive evolution from model behavior: the former closes a bidirectional loop between dialogue generation and rubric refinement, while the latter distills new rubric entries by analyzing differences between current and reference policy outputs at each training step. Taking a more controlled stance, InfiMed-ORBIT~\citep{wang2025infimedorbit} incrementally increases rubric difficulty throughout training to prevent early collapse, and DR Tulu~\citep{shao2025drtulu} maintains a dynamic balance by generating both positive and negative rules online to simultaneously encourage exploration and suppress reward hacking.

\textbf{Rubric-Model Co-evolution.} In this direction, rubric quality and model capability reinforce each other under a shared optimization objective, with works differing in how tightly the two sides are coupled. EvoLM~\citep{li2026evolm} establishes the core loop: a rubric generator and a response generator share parameters within a single model, where better rubrics elicit better responses, which in turn demand more discriminative rubrics. Rubric-ARM~\citep{xu2026rubricARM} formalizes this intuition within a reinforcement learning framework, showing theoretically that alternating the two optimization targets reduces gradient variance compared to joint updates; RUBRIC-ARROW~\citep{jiang2026rubricarrow} instantiates the same alternating principle without relying on frontier LLMs, jointly training a small rubric generator and a rubric-conditioned judge from preference data alone. While these works maintain a generator-judge separation, RLCER~\citep{sheng2026RLCER} and~\citet{ye2025selfrewardingrubric} collapse the distinction entirely, having a single policy model act simultaneously as rubric generator and scorer so that generation and evaluation capabilities improve together. RLAC~\citep{wu2025rlac} takes the most adversarial stance: rather than cooperative co-training, the judge is explicitly trained to expose weaknesses in generated content, forcing improvement through competition rather than collaboration. EvoRubrics~\citep{ding2026evorubrics} extends this adversarial dynamic to curriculum construction, having the rubric generator continuously produce harder and more discriminative criteria as the policy improves, with the trained generator further transferring to new tasks without external supervision. Beyond parameter-sharing architectures, \citet{wang2026generating} close the co-evolution loop at the rubric generation level itself, using a meta-judge to produce rubric preference pairs that fine-tune the generator without any human annotation.

\begin{summarybox}{Summary}
This chapter organizes existing work along two dimensions: how rubrics are constructed across varying degrees of human involvement, and how they are subsequently improved from passive refinement to active evolution, from which three core conclusions can be drawn.
\begin{itemize}
    \item[\raisebox{-0.15ex}{\includegraphics[width=1.2em]{image/lightbulb.png}}] Human expertise remains the gold standard for rubric quality, but its prohibitive cost makes it impractical as a routine approach, pushing the field toward increasingly automated paradigms.
    \item[\raisebox{-0.15ex}{\includegraphics[width=1.2em]{image/lightbulb.png}}] Automated construction has diversified rapidly, yet the core risk has shifted from rubric absence to rubric harm: low-quality rubrics can actively degrade model judgment, making quality control an integral part of the construction pipeline.
    \item[\raisebox{-0.15ex}{\includegraphics[width=1.2em]{image/lightbulb.png}}] The boundary between rubric and model is increasingly dissolving: rather than being imposed as static external standards, rubrics are progressively generated, refined, and evolved through model behavior itself, foreshadowing their emergence as endogenous components of intelligent systems.
\end{itemize}
\end{summarybox}
\section{How Do Rubrics Power Evaluation?}\label{section4}
Traditional evaluation pipelines rely on holistic scoring that suffers from three fundamental limitations: interpretive opacity, which collapses multidimensional quality into a single undifferentiated score with no articulable rationale; inconsistent execution, where judges are susceptible to systematic biases driven by non-semantic cues rather than actual quality; and static passivity, where evaluation signals terminate at scoring rather than feeding back into model improvement. Rubrics address all three simultaneously. By decomposing quality into explicit, independently verifiable criteria, rubrics make evaluation interpretable, harder to distort, and actionable beyond the scoring step itself.

This chapter traces how rubrics restructure the evaluation process: from establishing explicit criteria through structural and format-level decomposition (§~\ref{4.1Explicit}), to enforcing faithful and unbiased execution through objective traceability and subjective bias suppression (§~\ref{4.2Faithful}), concluding with the most consequential extension where rubric-based evaluation signals transcend passive assessment to drive active inference-time refinement and candidate selection (§~\ref{4.3Beyond}).

\subsection{Explicit Evaluation: Deconstruction and Representation of Criteria}\label{4.1Explicit}
\begin{figure}[t]
\centering
\includegraphics[width=\linewidth]{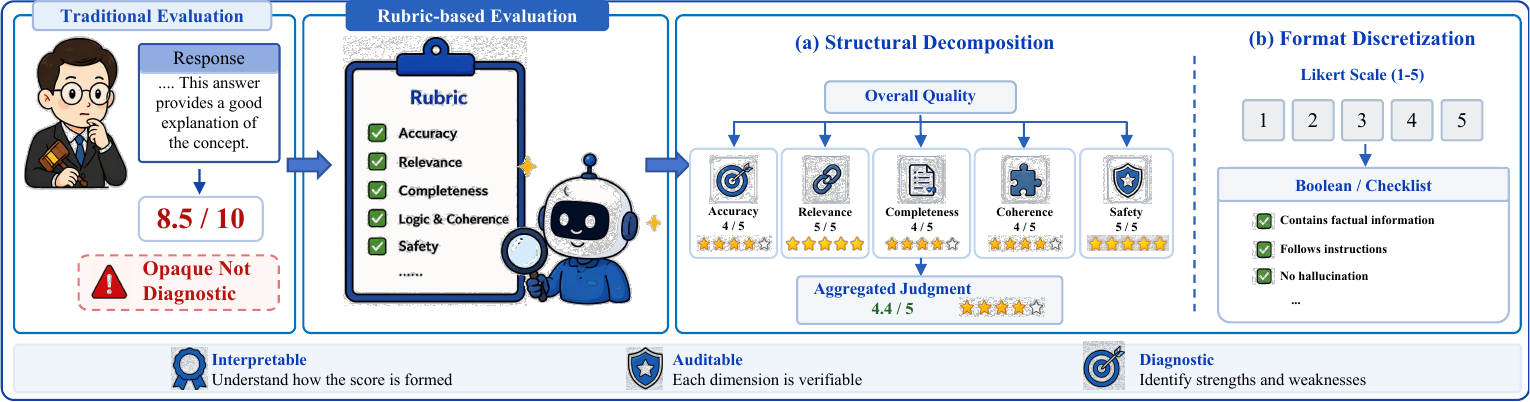}
\caption{Rubric-Based Evaluation through Structural Decomposition and Format Discretization.}
\label{fig:Explicit_Evaluate}
\end{figure}

LLM evaluation has long relied on single scalar scores. While G-Eval~\citep{liu2023geval} demonstrated the promise of criterion-guided LLM judging, it also exposed the fundamental limitations of holistic scoring: a single scalar cannot explain its own rationale or support fine-grained diagnosis. To address this limitation, rubric-based evaluation progressively shifts from holistic judgments toward more explicit and interpretable assessment processes~\citep{zhang2026llmevallogic}. As illustrated in Figure~\ref{fig:Explicit_Evaluate}, this section examines two complementary directions toward this goal: structurally decomposing holistic judgments into assessable analytic dimensions, and discretizing the scoring format within each dimension from continuous scales into verifiable checks.

\subsubsection{Structural Decomposition}
Analytic evaluation converts implicit quality judgments into explicit, independently assessable dimensions, along two complementary strategies.

\textbf{Fixed-dimension decomposition} establishes a fixed set of orthogonal dimensions applicable across instances. LLM-RUBRIC~\citep{Hashemi_2024llmrubric} is the foundational work: rather than chasing a single ground-truth score, it models each rater's individualized judgment by aggregating per-dimension distributions through a calibration network, capturing inter-rater subjectivity rather than discarding it as noise. \citet{rao2026autorubric} further operationalizes this paradigm as a unified toolchain supporting binary, scalar, and categorical rubric types with built-in bias mitigations. The same principle extends to specialized domains: AesRM~\citep{han2026aesrm} decomposes video aesthetics into three orthogonal dimensions to prevent inter-dimensional contamination, while MTalk-Bench~\citep{du2025mtalkbench} applies dimension-level scoring across semantic, paralinguistic, and environmental axes in speech dialogue evaluation.

\textbf{Instance-adaptive decomposition} decomposition tailors evaluation criteria to the specific task or domain at hand, enabling more targeted diagnosis. ExpertLongBench~\citep{ruan2025expertbench} structures long-form evaluation as a three-step process, moving from rubric to checklist to comparison against reference outputs, converting subjective judgment into structured information extraction. In educational assessment, \citet{favero2026holistic} evaluate writing-specific trait dimensions such as content organization and argumentative quality independently, substantially improving diagnostic value. In code evaluation, \citet{Pathak_2025} demonstrates that evaluating against question-specific rubrics far outperforms generic criteria, as task-tailored standards capture the precise decision boundaries that domain-agnostic dimensions miss.

However, decomposition is not universally superior. \citet{zhang2026rethinking} finds that on tasks requiring high completeness, holistic judges equipped with detailed rubrics outperform atomic judges, as fine-grained decomposition can fragment completeness reasoning and make global omissions harder to detect. Therefore, the appropriate granularity must be chosen with respect to the task at hand.

\subsubsection{Format Discretization}
Beyond dimensional structure, the scoring format within each dimension critically shapes evaluation consistency. Likert scales are the most common choice, but their inherent subjectivity limits reproducibility, motivating a transformation toward discrete, verifiable checks along two strategies.

\textbf{Boolean verification} converts ambiguous degree judgments into binary yes/no checks, eliminating interpretive ambiguity at its root. \citet{mallinar2026scalable} decompose multi-dimensional Likert rubrics into fine-grained binary criteria and dynamically retains the most relevant subset per query, significantly improving inter-rater agreement. At a larger scale, HealthBench~\citep{arora2025healthbench} deploys this approach across 48,562 physician-authored binary criteria spanning medical dialogue, demonstrating its scalability in high-stakes settings where ambiguity carries impact.

\textbf{Dual-track scoring} goes further by distinguishing positive and negative evidence separately. SedarEval~\citep{fan2025sedareval} structures evaluation as addition items (rewarding correct behaviors) and deduction items (penalizing errors), mimicking human exam grading logic and enabling diagnosis beyond what binary verification allows: distinguishing “answered correctly but penalized” from “answered incorrectly but partially credited”, two failure modes requiring different remediation.

\subsection{Faithful Evaluation: Reliability Reinforcement of Judgment}\label{4.2Faithful}
Having established explicit criteria, this section turns to whether judge models can reliably execute them. Reliable rubric execution faces two distinct challenges: whether scoring decisions are objectively traceable and auditable, and whether judge behavior remains free from subjective distortions introduced by non-semantic cues.

\subsubsection{Objective Evidence Anchoring}
Scoring decisions should be objectively traceable and auditable, verifiable by external parties independent of the judge's reasoning. Two lines of work address this from complementary angles.

\textbf{Structural locking and evidence enforcement} make scoring decisions auditable by design. RULERS~\citep{hong2026rulers} compiles criteria into versioned immutable bundles, requires judges to cite auditable evidence for every scoring decision, and applies post-hoc calibration to align score distributions with human annotations. DeCE~\citep{yu2025DeCE} takes a complementary approach, splitting evaluation into orthogonal precision and recall workflows to prevent inter-criteria interference, outperforming pointwise LLM scoring in expert alignment.

\textbf{Reliability measurement and validity testing} move from execution to verification. \citet{amin2026Co-Creation} propose a six-dimensional reliability protocol, substantially more informative than conventional single-metric approaches, while~\citet{huynh2026quantify} show that rubric design choices have asymmetric effects: representative examples improve human-autorater consistency, whereas excessive complexity reduces it; and \citet{lim2026reliableexpress} further demonstrate that safety judges remain brittle to rubric phrasing variations, proposing a reliable-to-expressive curriculum that progressively exposes judges to diverse rubric formulations to improve cross-rubric consistency. RubricEval~\citep{pan2026rubriceval} goes further as the first criterion-level meta-evaluation benchmark, directly verifying judge accuracy at the granularity of individual rubric items. \citet{chen2026criterion} raise a deeper question: even faithful execution may not suffice if scores fail to predict their intended outcomes, finding that criteria tied to user trust predict real-world conversion rates while technical capability dimensions do not.

\subsubsection{Subjective Bias Suppression}\label{4.2.2biassuppression}
Even when scoring decisions are traceable, judge behavior may still be distorted by subjective biases, where non-semantic cues are internalized as spurious preference signals. Research has identified multiple such bias sources, each requiring targeted mitigation.

\textbf{Bias identification} has produced a growing taxonomy. On \textit{scoring format biases}, \citet{li2026evaluatingscoringbias} identifies rubric order bias, score ID bias, and reference score bias, each showing that superficial prompt-level features systematically distort scores; position bias in score selection, where judges favor options at particular list positions, has also been documented \citet{xu2026ipointwise}. On \textit{self-referential biases}, judges have been found to favor outputs from their own model family even under fully objective criteria, marking wrong answers as satisfying at rates up to 50\% \cite{pombal2026selfpreferencebias}. On \textit{contextual instability}, Sage~\citep{feng2025sage} documents systematic criterion drift across answer combinations, Curse of Knowledge~\citep{li2025curseknowledge} identifies criterion gap and entanglement biases finding paradoxically that larger reasoning models are more vulnerable, and~\citet{weng2026BeyondAccuracy} reveals that safety judge verdicts drift with minor rubric wording changes independently of actual behavior. Beyond these instance-level biases, TRACE~\citep{mittal2026TRACE} takes a \textit{structural perspective}, mapping LLM-human misalignment across coding modalities and identifying 35 significant misalignment sources that aggregate accuracy metrics conceal.

\textbf{Bias mitigation} operates at two levels. At the \textit{training level}, FairJudge~\citep{yang2026fairjudge} addresses all three bias classes through a curriculum paradigm where SFT instills criterion compliance, DPO optimizes against non-semantic sensitivity, and GRPO enforces cross-mode consistency. At the \textit{inference level}, rubric-locking reduces inconsistency with no training required~\citep{feng2025sage}; boundary-focused exemplar selection narrows adjacent-level confusion~\citep{chu2026GUIDE}; and consensus-voting deferral handles low-confidence predictions~\citep{deng2025rubricconditioned}. More broadly, ~\citet{li2025leveraging} reveal that rubric granularity must match task type, as detailed rubrics benefit reasoning tasks but hurt coding tasks when irrelevant criteria are introduced.

\subsection{Beyond Evaluation: Capabilities Extension via Test-Time Scaling}\label{4.3Beyond}
\begin{figure}[t]
\centering
\includegraphics[width=\linewidth]{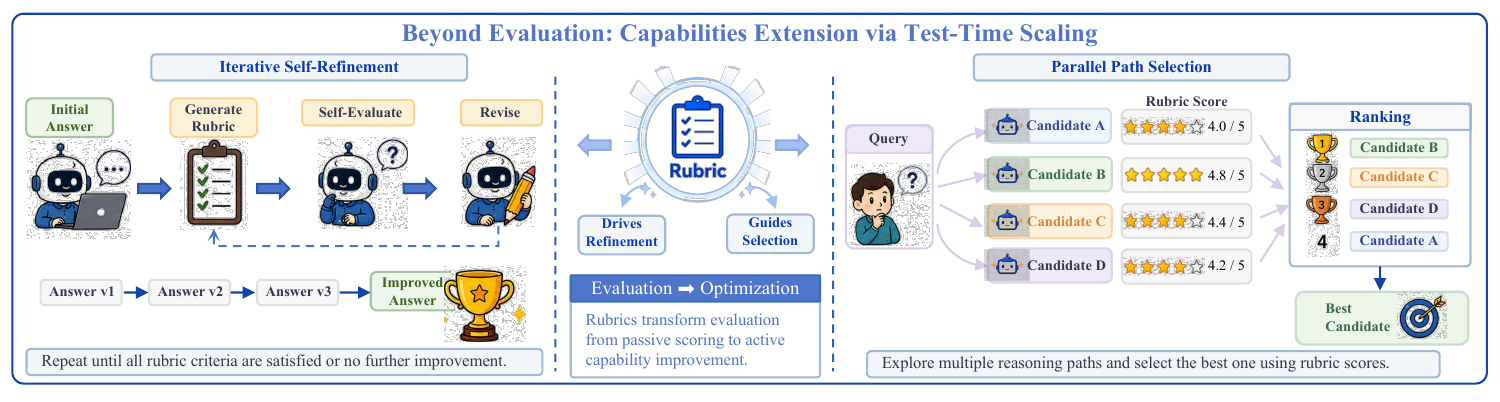}
\caption{Rubric-Guided Test-Time Scaling via Iterative Refinement and Path Selection.}
\label{fig:Beyond_Evaluation}
\end{figure}

This section examines how rubric-based evaluation signals transcend their traditional scoring role and become active drivers of inference-time improvement. As illustrated in Figure~\ref{fig:Beyond_Evaluation}, rubrics can support capability extension through two complementary mechanisms: iterative self-refinement driven by structured feedback, and parallel path selection that uses rubrics as lightweight verifiers to identify higher-quality reasoning trajectories.

\subsubsection{Iterative Self-Refinement}
Rubric signals can drive output improvement without additional training. By injecting structured evaluation feedback into the generation loop, models use self-generated criteria to iteratively refine outputs at inference time.

\textbf{Single-model self-evaluation} uses rubric-based self-assessment to drive iterative refinement without external supervision. TICK~\citep{cook2024tick} decomposes each instruction into a yes/no checklist and uses the resulting signal to drive iterative refinement, with the added finding that providing the checklists to human evaluators substantially improves inter-annotator agreement, confirming that rubric structure standardizes evaluation cognition for both humans and models. iRULER~\citep{bai2026iruler} extends this with three-dimensional feedback and applies the mechanism recursively via a rubric-of-rubrics meta-evaluation, enabling rubrics to co-evolve with writing practice. Think-with-Rubrics~\citep{yu2026thinkwithrubrics} takes this further by integrating rubric generation into the reasoning context itself: the model first generates task-specific rubrics as prior constraints and produces responses guided by them, internalizing quality standards as part of the reasoning process rather than treating them as external signals. These approaches share a common assumption that rubrics operate at the response level. Co-ReAct~\citep{kang2026coreact} challenges this by repositioning rubrics as pre-execution normative constraints: at each ReAct decision point, a step-level rubric conditioned on the current partial trajectory specifies what the next action must satisfy, shifting rubrics from post-hoc evaluators to action-selection guides. \citet{wang2026learnableassessmentskill} and Eval-Skill~\citep{yue2026rubricsexplorat} further treat evaluation competence as accumulative, distilling reusable scoring knowledge across tasks rather than regenerating criteria from scratch each time. VISTA~\citep{long2025vista} similarly applies iterative rubric-guided refinement to video generation, where critic agents evaluate structured criteria across visual, audio, and contextual dimensions and synthesize feedback to refine generation prompts across rounds. 

\textbf{Multi-agent rubric-guided grounding} applies structured criteria to coordinate specialized agents toward substantive outputs. ReviewGrounder~\citep{li2026reviewgrounder} decomposes peer review into drafting and grounding stages, where rubrics guide literature retrieval and evidence integration, demonstrating that rubric-structured guidance can compensate for model scale. DeepVerifier~\citep{wan2026DeepVerifier} extends this to deep research agents, inductively deriving rubrics from historical failure trajectories and deploying a three-agent pipeline to iteratively refine outputs, yielding consistent accuracy gains on deep research benchmarks. DuMate~\citep{yan2026DuMate-DeepResearch} further integrates rubrics as online reasoning scaffolds in a multi-agent deep research system, dynamically generating task-specific quality criteria at each synthesis step to anchor evidence integration and adaptively determine when retrieval is sufficient to terminate.

\subsubsection{Parallel Path Selection}
Rubrics can simultaneously score multiple candidates to select the best, exploiting the asymmetry that verification is cheaper than regeneration. \citet{raghavendra2026agenticrubrics} generate repository-context-aware rubrics and score candidate patches without code execution, with rubric scores highly aligned with actual test outcomes. Ctx2Skill~\citep{si2026ctx2skill} applies rubrics as lightweight inference-time probes in a three-role self-play cycle to autonomously discover and refine context-specific skills, with no parameter updates required. RubricRefine~\citep{levine2026rubricrefine} extends this to tool-use agents, generating rubrics from task requirements and tool documentation to perform pre-execution compliance checking, intercepting semantic defects before execution. \citet{ye2026rubricguidedprocess} applies a similar principle to model selection, using rubric-based process rewards to decide at each reasoning step whether to route to a stronger model for regeneration, providing richer routing signals than uncertainty-based heuristics.

\begin{summarybox}{Summary}
This chapter organizes existing work along three dimensions, explicit criteria construction, faithful execution, and inference-time extension, from which three core conclusions can be drawn.
\begin{itemize}
    \item[\raisebox{-0.15ex}{\includegraphics[width=1.2em]{image/lightbulb.png}}] The shift from holistic to analytic evaluation is not merely a formatting choice: explicit, verifiable criteria fundamentally change what evaluation can diagnose, enabling fine-grained attribution that scalar scores cannot provide.
    \item[\raisebox{-0.15ex}{\includegraphics[width=1.2em]{image/lightbulb.png}}] Faithful execution cannot be assumed. Judge models exhibit systematic biases across format, self-reference, and contextual instability, and reliability depends as much on how criteria are executed as on how they are designed.
    \item[\raisebox{-0.15ex}{\includegraphics[width=1.2em]{image/lightbulb.png}}] Evaluation signals are no longer terminal outputs. As rubric-based judgments are increasingly incorporated into inference-time refinement and decision-making loops, evaluation itself begins to function as an active supervisory mechanism rather than a passive measurement process.
\end{itemize}
\end{summarybox}
\section{How Do Rubrics Power Training?}\label{section5}
Traditional alignment pipelines rely on scalar reward models that suffer from three fundamental limitations: informational sparsity, which collapses multidimensional quality into a single undifferentiated score; expressive limitation, which cannot capture the complex multi-constraint nature of open-ended tasks; and vulnerability to reward hacking, which actively misleads optimization once a model learns to exploit the reward model's blind spots. Rubrics address all three simultaneously. By decomposing quality into explicit, independently verifiable criteria, rubrics provide high-resolution supervision, make evaluation harder to game, and enable practitioners to diagnose failure modes rather than merely observe them. 

This chapter traces how rubrics have entered the LLM training pipeline: from theoretical foundations (§~\ref{5.1Theoretical}) to reward signal design at output and process levels (§~\ref{5.2Reward}), through online RL training across open-ended, long-horizon, and multimodal tasks (§~\ref{5.3Online}), to offline training via preference optimization and rejection sampling (§~\ref{5.4Offline}), concluding with the most consequential shift where rubrics evolve from externally imposed standards into endogenous mechanisms that co-evolve with the model itself (§~\ref{5.5Beyond}).

\subsection{Theoretical Grounding: Why Rubrics Outperform Scalar Rewards}\label{5.1Theoretical}
\begin{figure}[t]
\centering
\includegraphics[width=\linewidth]{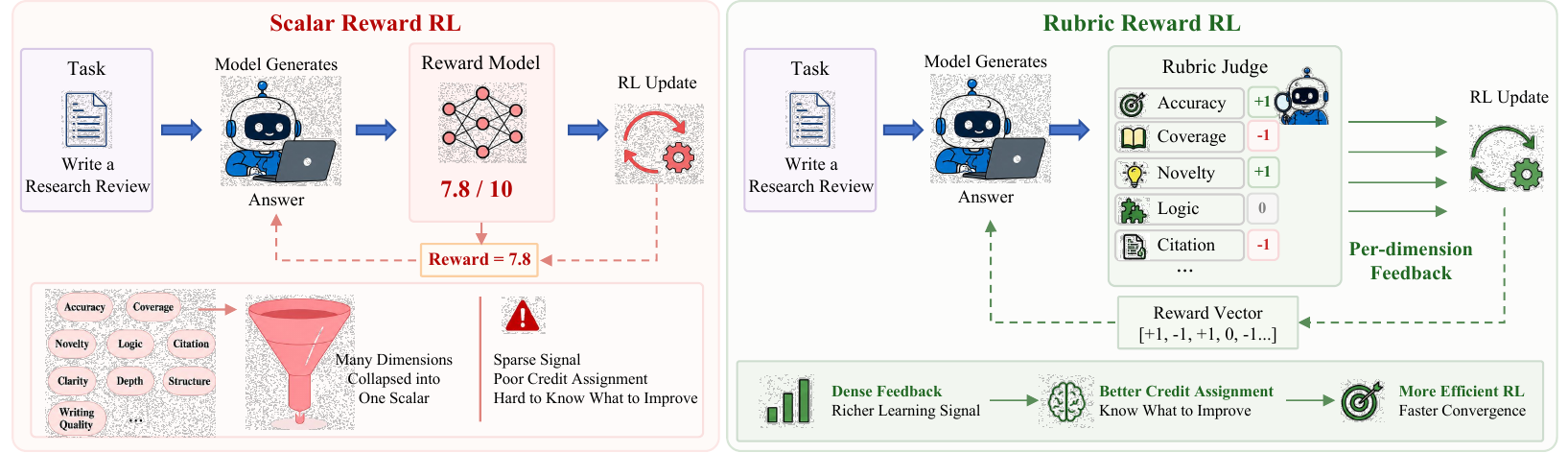}
\caption{Scalar Reward RL and Rubric Reward RL as Training Signals for LLM Optimization.}
\label{fig:Rubric_RL}
\end{figure}

Before examining how rubrics are deployed in practice, it is worth establishing why they constitute a theoretically superior training signal. Several independent lines of work converge on the conclusion that scalar rewards are not merely suboptimal but structurally inadequate for the demands of modern LLM training. Figure~\ref{fig:Rubric_RL} illustrates this distinction: while scalar rewards collapse multiple evaluation dimensions into a single score, rubric rewards preserve dimension-level feedback, yielding richer learning signals and more effective credit assignment.

\textbf{Why scalar rewards fail.} \citet{zhang2026chasingtail} prove that reward over-optimization originates specifically in the high-reward tail: scalar reward models cannot reliably distinguish excellent responses from merely good ones, causing policies to overfit to distributional artifacts rather than genuine quality. Errors concentrated in this region cause win rates to collapse as KL divergence increases, a structural failure rather than an incidental one. \citet{gupta2025carmo} provides the mathematical confirmation: for any finite fixed set of evaluation criteria, there always exists a true reward function the rubric completely fails to capture. No matter how carefully a scalar reward is constructed, there will always be quality dimensions it structurally cannot express. Together, these results establish that scalar rewards fail in practice precisely because they must fail in theory.

\textbf{Why rubrics are better.} OpenRubrics~\citep{liu2026openrubric} and CDRRM~\citep{liu2026cdrrm} argue for a fundamental reframing: traditional RLHF trains models to learn a ranking, whereas rubric-based training trains models to learn the evaluative basis for that ranking, a distinction that matters because learned rankings are opaque and brittle while learned criteria are interpretable and compositional. OpenRubrics operationalizes this through contrastive rubric generation, extracting discriminative constraints directly from preference pairs. CDRRM instantiates the same idea by mining high-discriminability differences between strong and weak responses and codifying them as rubric rules. Beyond reframing the learning target, rubrics also enforce structural properties that scalar rewards cannot. \citet{lianget2025generative} demonstrate that a rubric-conditioned preference tree, where each branch corresponds to an independent criterion, compels logically grounded justifications: responses that score correctly through flawed reasoning are penalized because every branch must be independently verifiable. This anti-hacking property emerges from the rubric's decomposed structure itself, requiring no additional training objective.

\subsection{Rubric Reward Modeling: Designing the Signal}\label{5.2Reward}
Given the theoretical motivation, the practical question becomes: how should rubric-based reward signals be designed and operationalized? We distinguish two complementary levels at which rubrics intervene in the training signal: output-level rewards, which evaluate the final response as a whole, and process-level rewards, which penetrate the intermediate reasoning steps that produce it. Within each level, we organize the discussion along a consistent axis of increasing granularity, moving from coarse holistic judgments toward finer-grained, more targeted signals.

\subsubsection{Output-Level Reward}
Output-level rubric rewards evaluate the final response against explicit criteria. The discussion follows increasing granularity from holistic scoring to decomposed aggregation, pairwise comparison, and finally causal grounding.

\textbf{From holistic to decomposed scoring.} Holistic LLM scoring collapses all quality dimensions into one undifferentiated signal, providing no information about which aspects of the response are strong or weak. The most direct remedy is explicit decomposition: RaR~\citep{gunjal2025rar} categorizes rubric types into mandatory, important, bonus, and penalty items with importance weights; RLCF~\citep{viswanathan2025rlcf} decomposes per-instruction quality into checklists combining AI-judged and programmatically verifiable constraints; and RBR~\citep{mu2024RBR} represents safety behaviors as fine-grained combinable propositions, substantially reducing over-refusal while maintaining helpfulness. $\text{RLR}^3$~\citep{yu2026RLR3} pushes decomposition to the verifiability level: each rubric criterion is routed to either a deterministic verifier or an LLM judge depending on its nature, with score remapping and hierarchical aggregation preserving reward discriminability throughout training.

\textbf{From independent scoring to pairwise comparison.} Decomposed scoring still evaluates each response in isolation, which limits its ability to identify genuinely discriminative quality differences. \citet{jia2026openrs} address this by conditioning criteria on the semantic differences between two candidates rather than the properties of one, with ablations confirming that the comparison mechanism rather than rubric content drives the gains.

\textbf{From multi-dimensional aggregation to stability.} Combining multiple rubric dimensions introduces a new problem: reward instability during RL training. PAPO~\citep{tan2026papo} traces this to GRPO's joint normalization across all dimensions and resolves it through decoupled advantage normalization, performing within-group normalization independently per dimension before aggregation.

\textbf{From surface features to causal grounding.} Even well-designed rubric rewards remain vulnerable if the reward model latches onto spurious correlates rather than genuine quality drivers. CROME~\citep{srivastava2025crome} addresses this by using rubrics as causal intervention variables, generating samples that enforce sensitivity to causal dimensions and invariance to spurious ones. Rubicon~\citep{huang2025rubicon} tackles the same problem at scale through a library of over 10,000 criteria with veto mechanisms and dedicated defense rubrics, while a structural approach conditions the judge on source documents inaccessible to the policy~\citep{bhattarai2026rubricgroundedrl}.

\subsubsection{Process-Level Reward}
Output-level rewards evaluate whether the final response is good, but say nothing about how it was produced: a model that arrives at a correct answer through flawed reasoning receives the same reward as one whose reasoning is sound. Process-level rubric rewards address this by supervising the intermediate steps themselves, progressing from response-level decoupling through step-level attribution to token-level allocation and trajectory evaluation.

\textbf{From output scoring to response-level decoupling.} The first step is to ensure that evaluation criteria are grounded in the specific instance rather than imposed externally. RM-R1~\citep{chen2026rmr1} proposes Chain-of-Rubrics: the model generates a sample-specific rubric conditioned on the task type before evaluating the response against each criterion, transforming opaque scalar scoring into a structured reasoning process.

\textbf{From response-level to step-level attribution.} Decoupling at the response level still assigns a single reward to the entire output, leaving individual reasoning steps without targeted feedback. AutoRubric-R1V~\citep{jia2026autorubricr1v} pushes supervision into individual steps by extracting common reasoning patterns from successful trajectories as ordered problem-specific rubrics. SRaR~\citep{xie2026SRaR} formalizes this further by mapping each criterion to the specific step responsible for satisfying it, revealing that response-level rewards systematically misattribute credit in both directions. DeepSeekMath-V2~\citep{shao2025deepseekmathv2} extends the same logic to mathematical reasoning, evaluating whether the verifier's logical feedback is itself rubric-justified rather than rewarding answer correctness alone. LongTraceRL~\citep{lin2026longtracerl} further applies entity-level rubric supervision with a positive-only strategy, restricting rubric signals to correct responses to distinguish reasoning quality without incentivizing answer-guessing.

\textbf{From step-level to token-level allocation.} Even at the step level, a rubric constraint may pertain to only a small fraction of tokens yet its score is amortized across all of them. RTT~\citep{xu2026rtt} resolves this by training a token-level relevance discriminator to identify which tokens satisfy each constraint, combining token-level and response-level signals through a unified GRPO objective. RGSD~\citep{rezaei2026RGSD} and RCSD~\citep{gu2026RCSD} take a distillation-based route, conditioning a teacher policy on rubric criteria to provide dense token-level supervision over student trajectories without requiring a separate verifier.

\textbf{From single-response to trajectory-level evaluation.} For long-horizon agentic tasks, the evaluation unit must extend to cover entire multi-step trajectories. Critic Rubrics~\citep{wang2026criticrubric} extracts behavioral rubric features from agent interaction trajectories and jointly models them with sparse outcomes to train a trajectory-level critic. SWE-TRACE~\citep{han2026swetrace} similarly decomposes SWE task quality into observable rubric checkpoints, converting binary test outcomes into dense trajectory-level rewards that supervise the full execution sequence.

\subsection{Rubric-Grounded Online Training: Driving RL Optimization}\label{5.3Online}
With the signal design established, we turn to how rubric-grounded rewards drive online RL training. The central challenge is domain coverage: standard RLVR works well for verifiable tasks with ground-truth answers, but fails for the vast majority of real-world tasks where correctness cannot be programmatically verified. The following subsections trace how rubric-grounded RL progressively expands its reach, adapting training dynamics to make rubric rewards more effective, and scaling to complex tasks where additional structural challenges arise.

\subsubsection{Extending RL to Non-Verifiable Domains}
Rubric-grounded RL extends the reach of reinforcement learning in two complementary directions: strengthening the reward signal in verifiable domains, and enabling RL in domains where verifiable rewards are unavailable.

\textbf{Complementing verifiable rewards.} Rubric-grounded RL strengthen verifiable reward signals even when ground-truth answers exist. Binary outcome rewards capture only whether the final answer is correct, saying nothing about the reasoning quality that produced it. \citet{chen2026improvingdatareward} combine verifiable rewards for closed-form problems with rubric rewards for open-ended questions, finding the mixed approach outperforms either signal alone. \citet{yuan2025Miracle} introduce rubrics that explicitly penalize spurious reasoning steps in mathematical reasoning that binary rewards would accept without question. LongTraceRL~\citep{lin2026longtracerl} similarly complements outcome rewards with entity-level rubric supervision over intermediate reasoning hops, distinguish reasoning quality among correct responses without incentivizing shortcut guessing. These results establish a foundational point: rubric rewards complement rather than merely substitute for verifiable signals.

\textbf{Foundational demonstrations.} The more demanding challenge is extending RL to tasks where no verifiable signal exists. RaR~\citep{gunjal2025rar} addresses settings where reference answers exist but binary correctness cannot capture the full quality spectrum, inducing rubric criteria from references and using weighted aggregation to replace holistic scoring. RLCF~\citep{viswanathan2025rlcf} targets the harder case where no reference exists, extracting per-instruction checklists from task descriptions combining AI-judged and programmatically verifiable constraints, finding it the only method to improve consistently across all evaluated benchmarks. \citet{mehta2026complexconstrain} extends this to complex instruction-following, replacing programmatic verification with expert-authored atomic rubrics as RL reward signals and demonstrating generalization to unseen agent benchmarks across model scales from 4B to 235B.

\textbf{Non-Verifiable domain extensions.} The same principle extends to settings where the source of non-verifiability differs. For writing tasks, quality is either inherently subjective or constrained by multiple simultaneous requirements that binary matching cannot decompose: Writing-Zero~\citep{jia2025writingzero} trains a pairwise generative reward model conditioned on writing principles for creative writing, while ACE-RL~\citep{chen2025acerl} dynamically generates adaptive constraint criteria per instruction for long-form generation. For open-ended question answering, responses are too diverse for binary matching: QuRL~\citep{wei2026qurl} mines case-wise rubrics from web resources as GRPO reward signals. In each case, rubric decomposition makes quality verification tractable precisely because verifiable rewards are structurally unavailable.

\subsubsection{Adapting RL to Efficient Training Dynamics}
Having a rubric-based reward signal is necessary but not sufficient for effective training. Even with well-designed rubric rewards, three interrelated problems can limit how effectively that signal drives learning.

\textbf{Exploration bottlenecks.} Models can only explore within the boundaries of their current capabilities, creating a circular dependency: generating high-quality training samples requires capabilities the model has not yet acquired. RGR-GRPO~\citep{bi2025RGRGRPO} assigns rubrics a dual role as both dense reward signals and offline guidance signals, using rubric-constrained trajectory refinement to produce off-policy samples that expand the solution space beyond online rollouts. RuscaRL~\citep{zhou2026Bottleneck} takes a complementary approach, injecting rubrics into task instructions during exploration and gradually decaying this injection as training progresses, so that the model internalizes quality standards rather than depending on external scaffolding. Where RGR-GRPO expands the solution space through offline refinement, RuscaRL does so through online guidance.

\textbf{Reward sparsity.} Assigning a single aggregated score to an entire response leaves intermediate reasoning steps without targeted feedback. SRaR~\citep{xie2026SRaR} addresses this by attributing each rubric criterion to the specific reasoning step responsible for satisfying it, normalizing per-step scores so that only steps with genuine quality variance generate a learning signal; RTT~\citep{xu2026rtt} pushes granularity further, training a token-level relevance discriminator to convert response-level rubric scores into token-level reward signals, providing targeted feedback precisely where the model's reasoning requires it; and Critic Rubrics~\citep{wang2026criticrubric} decomposes agent trajectories into behavioral rubric features, transforming sparse execution outcomes into dense process-level signals.

\textbf{Reward hacking.} When a policy learns to satisfy rubric criteria superficially, the reward signal actively misleads optimization. Several works address this from distinct angles. AdvancedIF~\citep{he2025advancedif} replaces general-purpose LLM judges with a dedicated rubric verifier fine-tuned on human-annotated data, making it substantially harder for the policy to exploit the judge's blind spots. DR Tulu~\citep{shao2025drtulu} takes a proactive approach, continuously generating negative rules that explicitly capture emerging reward hacking behaviors detected during training, so that the rubric set evolves to close loopholes as the policy discovers them. StitchCUDA~\citep{li2026stitchcuda} addresses hacking through signal combination, pairing rubric rewards with execution-based rewards so that each type of signal constrains the other: satisfying rubric criteria superficially without genuine functional correctness yields no net reward. However, since true response quality is unobservable in practice, understanding what drives hacking in the first place remains difficult. CHERRL~\citep{wang2026CHERRL} addresses this by injecting controlled biases into LLM judges to make hacking observable and locatable, characterizing how different bias types vary in their discoverability and exploitability by the policy.

\subsubsection{Scaling RL to Complex Tasks}
As tasks grow in structural complexity, two additional challenges emerge: credit assignment becomes harder when a single rubric score must cover long sequences or multi-step decisions, and modality heterogeneity requires criteria beyond what text-based rubrics can express.

\textbf{Long-horizon and agentic tasks.} For \textit{deep research}, rubrics enable multi-stage credit assignment by decomposing long trajectories into phase-specific evaluation criteria, with each phase conditioned on self-generated or retrieved rubrics that capture the quality requirements specific to that stage. Representative works include DR Tulu~\citep{shao2025drtulu}, RubricEM~\citep{li2026rubricem}, and CaRR~\citep{zhang2026CaRR}, which address complementary challenges: rubric evolution across dynamic content, memory-augmented trajectory distillation, and citation-grounded evidence verification respectively. The same decomposition principle extends to research planning~\citep{goel2025aicoscientists}, emotional support~\citep{yuan2025kardiar1}, medical consultation~\citep{wang2025infimedorbit}, legal judgment~\citep{su2026judger1}, and specialized technical analysis~\citep{xu2026wafersage}. For \textit{agentic systems}, ATLAS~\citep{gupta2026atlas} applies rubric-based reward decomposition to long-chain tool-use tasks, AskBench~\citep{zhao2026askaskbench} uses rubric-guided RLVR to separate clarification-seeking from response generation, while~\citet{wang2026criticrubric} extracts 24 behavioral rubric features directly from agent interaction trajectories to jointly model intermediate behavior quality and final success probability, and RUBAS~\citep{loye2026RUBAS} extends rubric decomposition to agent safety, structuring tool-use behavior along four dimensions to balance safety and helpfulness under RL optimization.

\textbf{Multimodal tasks.} Text-based rubrics are structurally insufficient for visual tasks where quality depends on spatial relationships and object attributes. Omni-RRM~\citep{kong2026omni} addresses this through a two-layer structure spanning text, image, video, and audio, combining globally shared criteria with modality-specific dimensions. Within \textit{vision-language tasks}, AutoRubric-R1V~\citep{jia2026autorubricr1v} extends process-level rubric supervision to counter spurious reasoning, and RuCL~\citep{chen2026rucl} introduces stratified curriculum learning that adjusts rubric reward weights as model capability advances. For \textit{generative visual tasks}, DeltaRubric~\citep{liu2026deltarubric} dynamically generates instance-specific visual rubrics before independently verifying each criterion, ARR~\citep{tian2026autorubric} externalizes vision-language model preference knowledge as prompt-specific rubrics before pairwise comparison, RubricRL~\citep{feng2025rubricrl} applies prompt-adaptive rubric generation with automatic dimension weighting, and AutoRubric-T2I~\citep{kao2026AutoRubric-T2I} learns discriminative rubrics for text-to-image alignment by synthesizing candidates from preference pair trajectories and pruning them via $\ell_1$-regularized refinement, driving diffusion model training through Flow-GRPO with far less annotated data than conventional reward models. Beyond vision, AnyAudio-Judge~\citep{li2026anyaudiojudge} extends dynamic rubric decomposition to audio instruction following, training a dense reward model via GRPO to supervise downstream RL for audio generation.

\subsection{Rubric-Guided Offline Training: Supervised Policy Improvement}\label{5.4Offline}
A parallel line of work leverages rubrics in offline settings, operating on fixed datasets constructed before training begins. Rubrics intervene in two distinct ways: guiding optimization through DPO-style preference learning, and selecting high-quality trajectories through supervised fine-tuning. Both replace coarse outcome signals with fine-grained rubric-based judgments, differing only in how the training signal is consumed.
\subsubsection{Preference Optimization}
Unlike online RL, where rubrics directly define reward signals, preference optimization leverages rubrics indirectly through the construction and selection of preference pairs. Existing work mainly focuses on improving preference quality and synthesizing controllable preference data.

\textbf{Rubric-based quality control.} The quality of preference pairs is as decisive as their quantity for DPO training. rDPO~\citep{yu2026rdpo} provides the clearest empirical demonstration: rubric-based filtering achieves substantially higher scores on multimodal benchmarks, while outcome-based filtering actually degrades the baseline, directly demonstrating that the filtering mechanism rather than data volume is decisive. C2~\citep{kawabata2026c2} identifies a subtler failure mode where low-quality rubrics actively mislead the reward model, resolving this by learning rubric quality automatically from binary preference data through a cooperative rubric generator paired with a critical verifier.

\textbf{Rubric-conditioned preference synthesis.} Beyond filtering existing data, rubrics enable systematic synthesis of preference pairs with controllable quality gradients. CPT~\citep{gallego2025configurable} uses structured rubrics as conditions for synthesizing preference pairs at different satisfaction  levels, training the model to dynamically adjust outputs based on rubric configurations at inference time without retraining. POP~\citep{huang2026PoP} extends this to a self-contained self-play pipeline where a single LLM simultaneously acts as proposer, solver, and verifier, using pre-training text as the rubric conditioning source to ensure a generation-verification gap that prevents reward hacking.

\subsubsection{Supervised Fine-tuning}
Supervised fine-tuning selects the highest-quality candidates from a pool of model-generated responses for supervised training. Rubrics transform this selection from a binary judgment into a multi-dimensional behavioral assessment.

\textbf{Rubric-guided trajectory selection.} \citet{huang2026Beyond} applies rubric-based selection to agentic SWE tasks, using rubrics encoding desired behavioral patterns to score and select trajectories for fine-tuning, significantly outperforming pure terminal-score selection by capturing richer behavioral signals that binary test outcomes cannot express.

\textbf{Rubric-based knowledge distillation.} A related offline path is knowledge distillation from strong teacher models, where rubrics replace the need for teacher logits or parameter access. ROPD~\citep{fang2026rubricbasedonpolicy} induces prompt-specific rubrics by contrasting teacher and student text outputs, identifying the quality dimensions on which the teacher outperforms the student and using these rubrics to drive policy gradient updates on student rollouts, achieving full black-box compatibility with proprietary models while demonstrating that explicit rubric signals carry no less information than implicit logits. RGSD~\citep{rezaei2026RGSD} and RCSD~\citep{gu2026RCSD} push this further by conditioning a teacher policy directly on rubric criteria to provide dense token-level supervision over student trajectories, eliminating the verifier and replacing sparse trajectory-end rewards with criterion-aware learning signals throughout the sequence.

\subsection{Beyond External Supervision: Rubric as Endogenous Mechanism}\label{5.5Beyond}
\begin{figure}[t]
\centering
\includegraphics[width=0.9\linewidth]{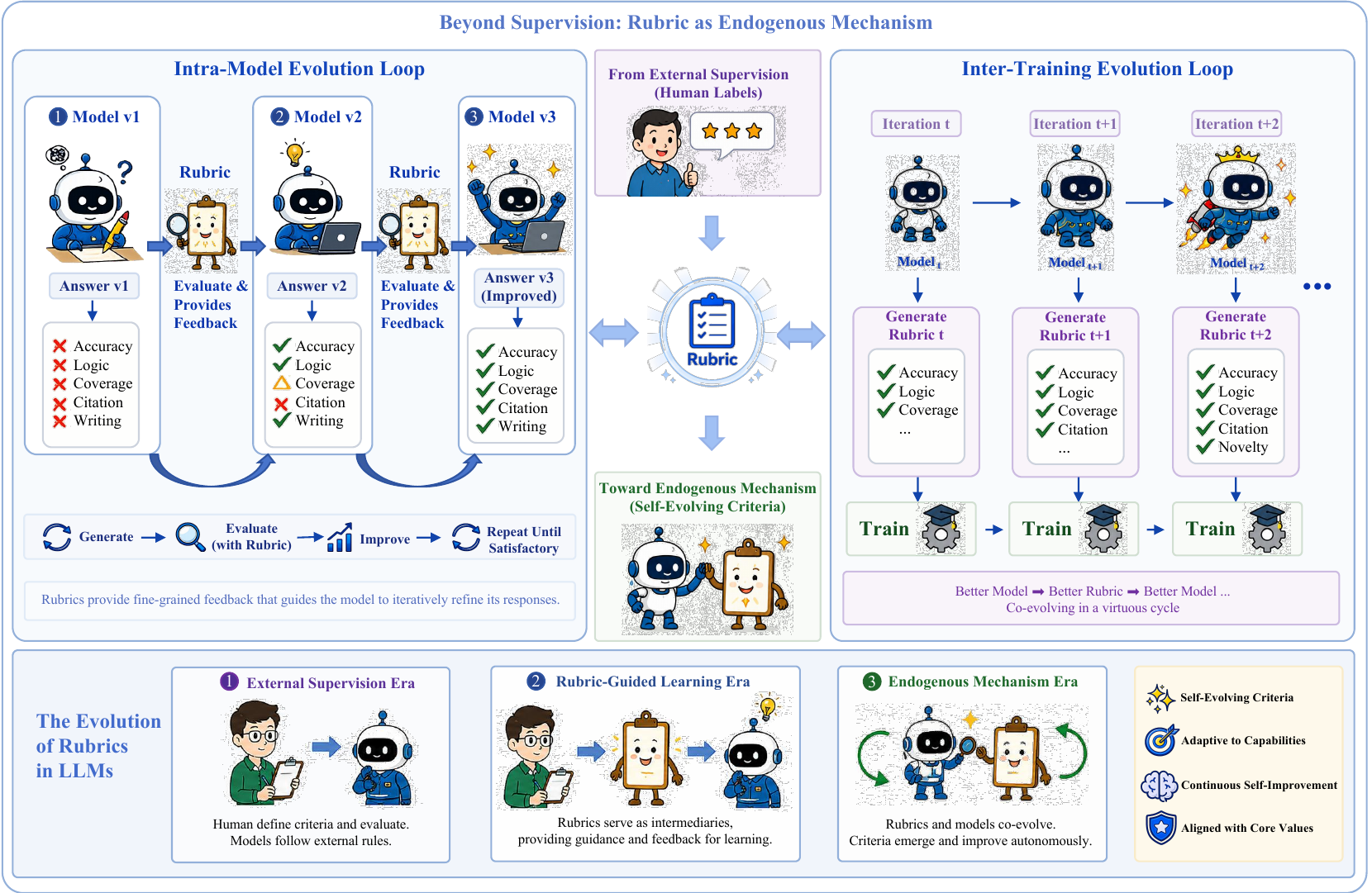}
\caption{The Evolution of Rubrics from External Supervision to Endogenous Mechanisms.}
\label{fig:Beyond_Supervision}
\end{figure}

The approaches overviewed so far share a common assumption: rubrics are externally specified and statically applied throughout training. This section examines the most consequential departure from this assumption, where rubrics are no longer passively consumed as supervision signals but actively generated, applied, and refined by the model itself. As illustrated in Figure~\ref{fig:Beyond_Supervision}, the evolution loop closes as rubrics and model capabilities progressively co-evolve, reducing reliance on external supervisory sources and transforming rubrics from evaluation criteria into endogenous learning mechanisms. We distinguish two levels at which this evolution operates: within a model, where generator and judge roles are jointly internalized and mutually reinforced, and across training iterations, where rubric criteria continuously adapt to the model's evolving capability frontier.

\subsubsection{Intra-Model Evolution Loop}
The most direct form of rubric endogenization is to train a single model to simultaneously generate responses and the rubrics used to evaluate them. Rather than relying on externally supplied criteria, the model develops its own evaluative standards alongside its generative capabilities, with the two reinforcing each other through training. This section traces three realizations of this loop, ordered by the degree of coupling between the generator and judge roles: from fully unified optimization where both roles share parameters, to coordinated alternating optimization where they share objectives but not parameters, to competitive adversarial dynamics where they drive each other through opposition.

\textbf{Cooperative: Unified Optimization.} In this approach, generator and judge share the same set of parameters, so improving one directly improves the other through backpropagation. EvoLM~\citep{li2026evolm}, EvoRubric~\citep{guan2026evorubrics}, and ARCO~\citep{tian2026arco} all instantiate this principle, differing in scope: EvoLM co-evolves a rubric generator and response generator within a single model; EvoRubric extends this to open-ended generation by alternating Reasoner and Rubric Generator roles with a multi-level verification pipeline; and ARCO applies it to multi-step agents through a dual-head shared-backbone architecture where generation and scoring heads co-evolve without proprietary judges. RLCER~\citep{sheng2026RLCER} implements a similar architecture, additionally filtering criteria based on their correlation with answer correctness. \citet{ye2025selfrewardingrubric} demonstrate the same principle at scale, using the model itself as both generator and scorer.

\textbf{Coordinated: Alternating Optimization.} In this approach, generator and judge have independent parameters but share a single optimization objective, updated in alternation rather than jointly to avoid gradient interference. Rubric-ARM~\citep{xu2026rubricARM} instantiates this by alternating between two phases. When the rubric generator is fixed, the judge is trained to maximize preference alignment. When the judge is fixed, the rubric generator is trained to produce criteria that maximize the judge's discriminative power. RUBRIC-ARROW~\citep{jiang2026rubricarrow} follows the same alternating principle but trains entirely from preference data without relying on frontier LLMs, replacing hard Boolean aggregation with soft scoring to recover reward discriminability. Alternating updates are proven to significantly reduce gradient variance and stabilize training compared to joint optimization.

\textbf{Competitive: Adversarial Optimization.} In this approach, generator and judge are trained against each other, with the judge actively seeking weaknesses in the generator's outputs and the generator learning to close those gaps. RLAC~\citep{wu2025rlac} instantiates this by training the judge to identify the most suspicious weaknesses in generated content and calling external verifiers only for these targeted points, while the generator is trained via RL using verification results as rewards. This competitive pressure drives both capabilities forward without any external guidance.

\subsubsection{Inter-Training Evolution Loop}
The inter-training loop operates at a longer timescale than the intra-model loop: rubric criteria themselves evolve across training steps, adapting to the model's shifting capability frontier. Static rubric sets inevitably become non-discriminative as training progresses, degrading the reward signal to noise. We distinguish two strategies: short-term evolution that responds to the model's current state, and long-term evolution that accumulates diagnostic knowledge across the full training history.

\textbf{Short-term Rubric Evolution.} The most immediate response to rubric saturation is to regenerate criteria at each training step based on the model's current behavior. OnlineRubrics~\citep{rezaei2025onlinerubric} does this by dynamically constructing criteria through pairwise comparison of current and reference policy responses, ensuring generated criteria remain genuinely discriminative. CaT~\citep{jayalath2026cat} derives rubrics from the model's own parallel rollouts through a frozen anchor model. DR Tulu~\citep{shao2025drtulu} extends this to deep research by continuously generating positive rules capturing newly explored knowledge and negative rules detecting emerging reward hacking behaviors, with variance-based filtering to prevent rubric proliferation. InfiMed-ORBIT~\citep{wang2025infimedorbit} and RuCL~\citep{chen2026rucl} apply curriculum scheduling, progressively increasing rubric difficulty as model capability advances so that criteria remain at the model's current learning frontier.

\textbf{Long-term Rubric Evolution.} Short-term adaptation responds to the current state but discards diagnostic information, preventing accumulation of evaluation knowledge across training. AMARIS~\citep{wu2026amarismemory} addresses this by maintaining a persistent memory repository that accumulates diagnostic information from each training round, tracking which criteria lose discriminative power and which dimensions the policy persistently fails, grounding rubric modifications in the complete training history rather than just the current state. SibylSense~\citep{xu2026SibylSense} takes a complementary approach, dynamically updating the rubric generator through a tunable memory repository that retains discriminative criteria and replaces saturated ones, while the policy model is adversarially updated to shrink the discriminative gap and trigger exploration of new quality dimensions. \citet{liu2026arbor} similarly maintains a reusable rubric buffer across queries, consolidating criteria and retiring stale entries as the policy evolves to provide process-level gradients when outcome rewards become uninformative.

\begin{summarybox}{Summary}
This chapter organizes existing work along five dimensions: theoretical grounding, reward signal design, online RL training, offline training, and endogenous evolution, from which three core conclusions can be drawn.
\begin{itemize}
\item[\raisebox{-0.15ex}{\includegraphics[width=1.2em]{image/lightbulb.png}}] Scalar rewards fail not because they are poorly designed but because they are an insufficient formalism: the finite criteria failure theorem and reward over-optimization results establish that rubrics are not a fallback but a structurally superior interface between human values and machine optimization.
\item[\raisebox{-0.15ex}{\includegraphics[width=1.2em]{image/lightbulb.png}}] Rubric rewards operate across multiple granularities and training paradigms: rubricRubric  supervision provides targeted feedback where scalar rewards are blind, rubric-grounded RL extends reinforcement learning from verifiable to non-verifiable domains and from text to multimodal settings, and offline rubric intervention improves preference data quality while enabling black-box knowledge distillation.
\item[\raisebox{-0.15ex}{\includegraphics[width=1.2em]{image/lightbulb.png}}] The most consequential shift is rubric endogenization: when rubrics co-evolve with the model through intra-model generator-judge loops and inter-training adaptive evolution, they transition from externally imposed constraints into self-sustaining training mechanisms that require no external supervisory source.
\end{itemize}
\end{summarybox}
\section{How Reliable Are Rubrics?}\label{section6}
Rubrics promise to enhance the reliability of LLM evaluation and alignment by making implicit criteria explicit and structured. However, the assumption that rubrics are inherently reliable deserves scrutiny in its own right. This section demonstrates that rubric reliability is not monolithic but distributed across four mutually independent yet progressively deepening dimensions: quality failures in generation, systematic biases in execution, fundamental constraints at the theoretical level, and security vulnerabilities arising from rubrics' role as a high-level interface. When these failure modes converge, we must confront the existence of boundary scenarios where rubrics are fundamentally ill-suited, and explore alternative paths beyond the rubric paradigm.

\subsection{Generation Quality: Are Rubrics Well-Constructed?}\label{6.1Generation}
LLM-generated rubrics exhibit systematic cognitive misalignment that cannot be resolved by scaling inference compute, and whose harm may in fact exceed that of using no rubric at all. In current practice, rubrics are predominantly generated automatically by LLMs, a process that introduces systematic quality risks. \citet{zhang2026rubricbench} quantify this risk through 1,147 carefully curated hard-instance comparison pairs: replacing model-generated rubrics with human-annotated counterparts improves judgment accuracy by an average of 27\% across state-of-the-art models.

\textbf{Cognitive misalignment as the root cause.} The gap stems not from insufficient generative capacity but from value misalignment: model-generated rubrics gravitate toward surface features such as format and length while systematically neglecting core implicit constraints such as task feasibility and safety boundaries, a gap that scaling inference compute cannot close~\citep{zhang2026rubricbench}. GER-Eval~\citep{siro2026GER}further reveals that this misalignment carries a model-specificity property: LLM-generated rubrics remain internally consistent within a given model but fragment significantly across models, rendering the widespread practice of “generate rubrics once, apply to all models” methodologically flawed.

\textbf{From diagnosis to quantified harm.} RIFT~\citep{qi2026rift} provides the first systematic diagnostic framework, inducting eight failure modes under three high-level categories: reliability failures, content validity failures, and consequential validity failures, with automated diagnostic signals achieving up to 0.86 F1 against expert annotations. The severity of these failures is further quantified: naively generated rubrics reduce GPT-4o's accuracy on JudgeBench from 55.6\% to 42.9\%, which is 13 percentage points below the no-rubric baseline~\citep{shen2026RRD}. This establishes a critical threshold: below a certain quality level, rubrics are not merely unhelpful but actively harmful.

Taken together, these findings establish that rubric quality is not an inherent property but a prerequisite that must be actively ensured, and that quality control alone is insufficient to guarantee reliable evaluation.

\subsection{Execution Fidelity: Are Rubrics Faithfully Executed?}\label{6.2Execution}
While Section~\ref{4.2.2biassuppression} catalogs the types of biases that affect LLM judges and overviews mitigation strategies, a distinct question remains: what do these biases collectively imply for rubric reliability as a paradigm? This section reframes the same evidence from a reliability perspective, arguing that execution-stage biases constitute a systemic threat through two independent mechanisms whose interaction effects are unique to the rubric setting.

\textbf{Two independent sources of fidelity failure.} Execution biases arise from two mutually independent sources: the intrinsic behavioral characteristics of judge models, and the structural design of scoring prompts. The heterogeneity of these sources carries a structural implication that goes beyond taxonomy: because the two classes of bias are rooted in fundamentally different mechanisms, no single mitigation strategy can address both simultaneously. Biases rooted in judge behavior, such as self-preference bias~\citep{pombal2026selfpreferencebias} and contextual criterion drift~\citep{feng2025sage}, originate in how models represent and process their own outputs relative to others; they persist across prompt redesigns because the prompt is not their source. Conversely, biases rooted in prompt structure, such as rubric ordering bias, score ID bias, and reference score bias~\citep{li2026evaluatingscoringbias}, as well as position bias in score selection~\citep{xu2026ipointwise}, are insensitive to judge model substitution because they are encoded in the evaluation interface itself rather than in the model. Empirically, even state-of-the-art LLM judges remain error-prone in rubric verification~\citep{peng2026llmasajudge}. Therefore, A rubric-based evaluation system must contend with two orthogonal failure channels simultaneously, and improving along one dimension leaves the other entirely unaddressed.

\textbf{Rubric structure as an amplifier of distortion.} More critically, rubric's multi-dimensional structure does not merely co-exist with these biases but actively amplifies them. \citet{li2025curseknowledge} identify criteria gap bias and criteria entanglement bias as structurally inevitable consequences of explicit dimension enumeration: by defining what judges attend to, rubrics simultaneously license them to ignore everything else, and the more dimensions a rubric specifies, the more interference pathways are created between them. GEAR~\citep{lv2026gear} reveals that this inter-criteria interference extends beyond scoring to aggregation: flat weighted summation assumes each criterion contributes independently, but real rubrics contain prerequisite and activation dependencies that, when ignored, cause false credit propagation and amplify local judge errors into optimization errors. The empirical manifestation of these structural tensions is directional: rubrics improve evaluation accuracy on standard samples but reduce it on adversarial ones, a reversal that would not occur in holistic evaluation. At a more fundamental level, \citet{weng2026BeyondAccuracy} demonstrate that structural instability extends to rubric interpretation itself, finding that judge verdicts are systematically sensitive to rubric wording variations and that existing judges broadly fail semantic invariance, threshold invariance, and ambiguity-aware calibration.

\subsection{Theoretical Constraints: Does the Rubric Paradigm Have Fundamental Limits?}\label{6.3Theoretical}
The preceding two sections addressed failures that are in principle amenable to engineering solutions: better generation methods can improve rubric quality, and better training or prompt design can reduce execution bias. This section confronts a deeper class of constraints, ones that are not implementation deficiencies but intrinsic to the theoretical structure of the rubric paradigm itself. 

\textbf{The Cross-Task Failure Theorem.} CARMO~\citep{gupta2025carmo} establishes through formal proof that for any finite fixed set of criteria, there always exists a true reward function under which the rubric yields zero predictive correlation. The intuition is fundamental: the space of possible true reward functions is unbounded, while any finite rubric spans only a fixed-dimensional subspace within it. As task diversity grows, the probability that a fixed rubric remains aligned with the true reward function approaches zero regardless of how carefully it is designed. This is not an engineering deficiency that better rubric design can overcome, but a mathematical property of the fixed-criteria paradigm itself. It also provides theoretical grounding for why dynamically generated criteria are a principled necessity rather than a practical convenience: only by allowing criteria to vary with the task can alignment between rubric and true reward function be maintained across the unbounded space of possible tasks. POW3R~\citep{tyagi2026pow3r} and Focal Reward~\citep{huang2026focalreward} instantiate this in training dynamics: the former shows that human-assigned weights diverge from actual informativeness as training progresses; the latter identifies dimension polarization as an independent failure mode where easily optimized dimensions mask persistent deficits in harder ones.

\textbf{The High-Reward Discrimination Bottleneck.} Even within a fixed task, a structurally distinct failure emerges at the high-reward tail. Rubrics assess whether criteria are met; this binary orientation provides adequate signal for distinguishing poor responses from acceptable ones, but loses resolution at the boundary between good and excellent. Under Pareto-optimal conditions in reinforcement fine-tuning, errors concentrated at the high-reward tail are disproportionately costly~\citep{zhang2026chasingtail}: a rubric that cannot reliably rank top responses forces the model to optimize noise rather than genuine quality differences. Tournament-GRPO~\citep{yang2026tournamentgrpo} corroborates this bottleneck through three structural failures of absolute rubric scoring: scale inconsistency, score compression among top candidates, and rapid saturation. The deeper implication is that rubric-based reward signals are structurally front-loaded, providing dense signal in the low-to-mid reward range but becoming progressively less informative as response quality approaches the ceiling of what rubric criteria can distinguish.

\textbf{Dimensional Misalignment with True Objectives.} Even when rubric dimensions are internally consistent and faithfully executed, they may remain structurally misaligned with the outcomes they purport to predict. This failure is not a consequence of poor rubric design in any conventional sense. It arises because rubric dimensions are typically selected based on domain expert intuition about what constitutes quality, while human intuition about which dimensions actually predict downstream outcomes is systematically unreliable. \citet{chen2026criterion} using business conversion rates as an external criterion, provides direct empirical evidence: some rubric dimensions correlate strongly with true outcomes while others show near-zero correlation. Critically, the presence of uninformative dimensions does not merely add noise but actively dilutes the predictive signal of informative ones through aggregation. This failure mode is particularly insidious because it is invisible to standard evaluation metrics: a rubric can achieve high inter-annotator agreement and strong judge consistency while remaining entirely disconnected from the outcomes it is meant to serve. JudgmentBench~\citep{yang2026judgmentbench} reveals the same misalignment in legal tasks: rubric scoring recovers quality rankings with Spearman correlation of only 0.150, compared to 0.908 for pairwise comparison, suggesting that rubric dimensions can be structurally disconnected from the quality signals that expert judgment actually tracks.

\textbf{The Model Capability Prerequisite.} All three constraints discussed above presuppose that models can interpret and reason over rubric criteria, yet this presupposition is not always warranted. \citet{wei2025conceptbased} demonstrates that rubric effectiveness requires a minimum level of conceptual reasoning capacity: rubrics prove effective for LLMs capable of criterion-driven reasoning but entirely ineffective for pre-trained language models, whose apparent responsiveness to rubric instructions reflects surface-level text conditioning rather than genuine criterion comprehension. \citet{jayarao2026explicitreasoning} further find that rubric guidance yields only marginal improvement at more than eight times the computational cost, while thinking models outperform non-thinking counterparts by approximately 10 percentage points in accuracy. Together, these findings establish that the capability prerequisite functions as the logical precondition for the entire rubric paradigm. Without sufficient reasoning capacity, rubric design loses its relevance, and advances in model reasoning capacity become the more fundamental requirement.

\subsection{Security Threats: Can Rubrics Be Weaponized?}\label{6.4Security}
The failure modes discussed in the preceding sections all arise under normal operating conditions. This section identifies a qualitatively different class of risk: rubrics' role as a high-level decision interface makes them an active attack surface, one whose exploitation is both difficult to detect and capable of producing irreversible downstream consequences.

\textbf{Preference Drift via Subtle Rubric Edits.} RIPD~\citep{ding2026ripd} demonstrates that subtle edits to rubrics, edits that appear natural and preserve the original criteria, can induce systematic, directional preference drift on target domains. The defining feature of this attack is its covertness: the induced drift is nearly invisible on aggregated benchmark metrics, with the most effective attacks reducing target-domain accuracy by up to 27.9\% while maintaining benchmark performance. This reveals a fundamental blind spot in current validation practices: strong benchmark performance no longer implies that the evaluation system is trustworthy, because the attack operates precisely in the gap between what benchmarks measure and what actually matters. \citet{lim2026reliableexpress} corroborate this vulnerability from a different angle, showing that safety judges exhibit systematic sensitivity to rubric phrasing variations, broadly failing semantic invariance and threshold invariance even without adversarial intent.

\textbf{Irreversibility Through the Training Pipeline.} The more consequential threat is not the attack itself but its persistence. When contaminated rubrics are used to generate preference labels for downstream post-training, the induced bias propagates through the alignment pipeline and becomes internalized in trained model parameters, producing behavioral drift that cannot be reversed by subsequently replacing the rubric~\citep{ding2026ripd}. This elevates rubric manipulation from a localized evaluation problem to a systemic alignment risk, transforming the rubric from an evaluation tool into a permanent bias injection mechanism. This elevates rubric manipulation from a localized evaluation problem to a systemic alignment risk, as by the time contamination is detected, the damage to model behavior may already be permanent. CHERRL~\citep{wang2026CHERRL} makes this dynamic observable under controlled conditions, injecting known biases into LLM judges to precisely localize hacking onset and reveal systematic differences in how different bias types are discovered and exploited by the policy.

\subsection{Boundaries and Alternatives: When Should We Look Beyond Rubrics?}\label{6.5Boundaries}
Rubric failures are not uniformly addressable. While some stem from implementation deficiencies amenable to engineering solutions, others reflect boundaries intrinsic to the rubric paradigm where no amount of refinement renders rubrics adequate. This section identifies one such boundary and maps the alternatives it admits, ranging from complete paradigm inversion to partial structural preservation.

\textbf{Complete abandonment: inverting the evaluation paradigm.} When tasks admit multiple valid outputs and lack a single correct answer, rubric generation has no anchor, as it is impossible to define what criteria should be satisfied without a unique reference point. JudgmentBench~\citep{yang2026judgmentbench} illustrates this boundary concretely, finding that pairwise comparative judgment substantially outperforms rubric-based scoring in expert legal tasks where quality resists decomposition into independently verifiable criteria. \citet{ikezogwo2026rubricsfail} addresses this through Implicit Error Counting (IEC), inverting the evaluation paradigm by shifting the reward signal from checking what the model got right to enumerating what the model got wrong. The key insight is that even when correct answers are not unique, error patterns tend to be more constrained and enumerable, making negative criteria more tractable than positive ones.

\textbf{Partial preservation: hybrid frameworks as an intermediate path.} Not all task-structural boundary cases demand complete rubric abandonment. When correct answers are not unique but response quality still varies along identifiable dimensions, rubric structure retains value as an organizational scaffold even when its execution logic proves inadequate. JADE~\citep{lin2026jade} demonstrates that a hybrid approach can be viable: the first layer retains a predefined set of evaluation skills that inherit rubrics' stability and reproducibility, while the second layer replaces rubrics' item-by-item verification with claim-level dynamic evaluation through an evidence dependency gating mechanism. The task-structural boundary therefore does not demand a binary choice between rubrics and their alternatives. Rubric structure can remain useful as an organizational scaffold even in scenarios where response quality cannot be reduced to criterion satisfaction, as long as its execution logic is replaced by a more flexible evaluation mechanism.

\begin{summarybox}{Summary}
This chapter examines rubric reliability across four progressively deepening dimensions: generation quality, execution fidelity, theoretical constraints, and security vulnerabilities, from which three core conclusions can be drawn.
\begin{itemize}
\item[\raisebox{-0.15ex}{\includegraphics[width=1.2em]{image/lightbulb.png}}] Rubric reliability is not a single property but a stratified one spanning multiple independent dimensions, each of which must be adequately addressed for evaluation to be trustworthy as a whole.
\item[\raisebox{-0.15ex}{\includegraphics[width=1.2em]{image/lightbulb.png}}] The theoretical limits of rubrics are not engineering deficiencies but intrinsic features of the paradigm, as fixed rubrics are mathematically guaranteed to fail across tasks and lose discriminative capacity precisely where training signal matters most.
\item[\raisebox{-0.15ex}{\includegraphics[width=1.2em]{image/lightbulb.png}}] Rubrics face both security vulnerabilities that can propagate irreversibly through training pipelines and structural boundaries where no amount of refinement renders them suitable. How to detect these limits in practice, defend against exploitation, and select appropriate alternatives remains an open challenge for the field.
\end{itemize}
\end{summarybox}

\section{Where Are Rubrics Applied?}\label{section7}
The preceding chapters have examined how rubrics are constructed, optimized, and deployed as reward signals, this chapter shifts focus to where these methods have been put to use. §~\ref{7.1benchmark} overviews representative rubric-based benchmarks, identifying systematic patterns in construction choices and scoring designs; §~\ref{7.2application} examines downstream applications across six domains, showing how rubric-based approaches have been adapted to the specific constraints of healthcare, law, education, finance, society, and scientific research.

\subsection{Benchmark}\label{7.1benchmark}
The proliferation of rubric-based benchmarks over recent years reflects a broader dissatisfaction with holistic evaluation metrics. As rubric-based evaluation has matured, a growing number of benchmarks have adopted structured criteria as their primary assessment mechanism. Table~\ref{tab:rubric_benchmarks} overviews a representative selection of benchmarks across five categories, characterizing each along construction source, scoring granularity, and content grounding following the taxonomy established in §~\ref{section2}.

General benchmarks establish the methodological baseline for rubric-based evaluation, targeting general-purpose instruction-following and natural language generation tasks rather than any specific professional domain. SedarEval~\citep{fan2025sedareval} introduces adaptive dual-track rubrics with credit and deduction items, demonstrating that fine-grained per-item scoring provides more actionable diagnostic signal than holistic judgments. RubricBench~\citep{zhang2026rubricbench} and RubricEval~\citep{pan2026rubriceval} then turn the lens on the rubrics themselves. RubricBench directly quantifies the gap between human-written and LLM-generated rubrics, finding that replacing model-generated criteria with expert ones boosts judge accuracy by approximately 27\%; RubricEval provides the first meta-evaluation benchmark at rubric-criterion granularity, asking not whether the judge's final verdict aligns with humans, but whether each individual criterion is applied correctly. Together, these two benchmarks establish a foundational problem that every subsequent category must confront: \textbf{LLM-generated rubrics systematically misalign with human judgment}, concentrating on surface features like format and length while ignoring task feasibility and implicit constraints. LLMEval-Logic~\citep{zhang2026llmevallogic} pushes evaluation further by pairing rubric atoms with a Z3 theorem prover, bridging natural language reasoning quality and machine-verifiable formal verification in logical reasoning tasks.

Professional benchmarks evaluate LLM performance on high-stakes expert-level tasks drawn from real-world workflows in fields such as medicine, law, and finance, where evaluation errors carry tangible consequences. When evaluation stakes are high enough, the community's answer to the LLM rubric quality problem is simply to remove LLMs from the construction process altogether. HealthBench~\citep{arora2025healthbench} enlists 262 physicians, PRBench~\citep{akyrek2025prbench} draws on 182 credentialed legal and financial professionals across 114 countries, and ProfBench~\citep{wang2025profbench} explicitly prohibits its 38 expert contributors from using any LLM assistance. This Expert-only convergence is not coincidental: ProfBench's controlled comparison directly demonstrates that rubrics generated by the same model used for scoring produce systematic self-enhancement bias, inflating quality estimates in ways that human-constructed rubrics do not. Recent benchmarks extend this expert-construction paradigm to new domains and task types: LexRubric~\citep{chen2026lexrubric} covers Chinese open-ended legal tasks with 12,337 atomic expert-written criteria across 14 legal scenarios; BIGFINANCEBENCH~\citep{wang2026bigfinancebench} grounds financial research agent evaluation in analyst workflows, decomposing 928 tasks into 15,656 auditable rubric criteria; PanCanBench~\citep{zhao2026pancanbench} collects real patient queries from a cancer helpline and pairs them with Boolean rubric criteria through a physician-led HIL pipeline; and LP-Eval~\citep{xu2026lpeval} takes a smaller-scale approach, co-designing a three-step rubric with legal experts to assess LLM-generated legal propositions from EU court judgments. However, the cost of this quality guarantee is severe. HealthBench required 262 physicians and ExpertLongBench~\citep{ruan2025expertbench} required over 10 hours of expert investment per task, construction efforts that are by definition non-replicable at scale. XpertBench~\citep{liu2026xpertbench} offers a partial concession through HIL construction, but its peak success rate of only 66\% across seven domains suggests that even human-validated rubrics cannot fully bridge the gap between current models and genuine expert-level performance. The Professional category thus crystallizes a fundamental trade-off: \textbf{Expert construction is the quality ceiling, but it is also a scalability wall}.
\begin{table*}[!t]
\centering
\small
\setlength{\tabcolsep}{4.5pt}
\renewcommand{\arraystretch}{1.1}
\caption{Overview of rubric-based benchmarks.
\textbf{Rubric Construction} follows the source taxonomy in \S3:
\textit{Expert} = domain experts without LLM assistance;
\textit{Auto} = fully automated LLM generation;
\textit{HIL} = human-in-the-loop.
\textbf{Scoring Design} follows the structural and content taxonomy in \S2:
\textit{Granularity}: Holistic / Analytic / Atomic;
\textit{Content}: Task-Grounded (\textit{Task}) / Behavior-Grounded (\textit{Behav.}) /
Knowledge-Grounded (\textit{Know.}).}
\label{tab:rubric_benchmarks}
\begin{tabular}{l l r ccc ll}
\toprule
\multirow{2}{*}{\textbf{Category}}
  & \multirow{2}{*}{\textbf{Benchmark}}
  & \multirow{2}{*}{\textbf{Data Size}}
  & \multicolumn{3}{c}{\textbf{Rubric Construction}}
  & \multicolumn{2}{c}{\textbf{Scoring Design}} \\
\cmidrule(lr){4-6}\cmidrule(lr){7-8}
  & & & \textbf{Expert} & \textbf{Auto} & \textbf{HIL}
        & \textbf{Granularity} & \textbf{Content} \\
\midrule

\multirow{5}{*}{General}
  & SedarEval~(\citeyear{fan2025sedareval})        & 1,000      & \xmark & \xmark & \cmark & Analytic & Task    \\
  & RubricBench~(\citeyear{zhang2026rubricbench})    & 1,147     & \cmark & \xmark & \xmark & Analytic & Task    \\
  & RubricEval~(\citeyear{pan2026rubriceval})      & 3,486    & \cmark & \xmark & \xmark & Atomic   & Task    \\
  & BenchBench~(\citeyear{zheng2026benchbench})    &  15,000           & \xmark & \cmark & \xmark & Holistic & Task    \\
  & LLMEval-Logic~(\citeyear{zhang2026llmevallogic})           & 2,338      & \cmark & \xmark & \xmark & Atomic   & Task   \\

\midrule
\multirow{10}{*}{Professional}
  & HealthBench~(\citeyear{arora2025healthbench})         & 48,562  & \cmark & \xmark & \xmark & Atomic   & Know.   \\
  & PRBench~(\citeyear{akyrek2025prbench})                 & 19,356  & \cmark & \xmark & \xmark & Analytic & Know.   \\
  & ProfBench~(\citeyear{wang2025profbench})             & 7,347     & \cmark & \xmark & \xmark & Analytic & Know.   \\
  & ExpertLongBench~(\citeyear{ruan2025expertbench}) & 1,050 & \cmark & \xmark & \xmark & Analytic & Know.   \\
  & XpertBench~(\citeyear{liu2026xpertbench})           & 1,346      & \xmark & \xmark & \cmark & Analytic & Know.   \\
  & PLAWBENCH~(\citeyear{shi2026plawbench})           & 12,500      & \cmark & \xmark & \xmark & Analytic & Know.   \\
  & LP-Eval~(\citeyear{xu2026lpeval})           & -      & \xmark & \xmark & \cmark & Analytic & Know.   \\
  & LexRubric~(\citeyear{chen2026lexrubric})           & 12,337      & \cmark & \xmark & \xmark & Atomic & Know.   \\
  & BIGFINANCEBENCH~(\citeyear{wang2026bigfinancebench})           & 15,656      & \cmark & \xmark & \xmark & Analytic & Know.   \\
  & PanCanBench~(\citeyear{zhao2026pancanbench})           & 3,130      & \xmark & \xmark & \cmark & Atomic   & Know.   \\

\midrule
\multirow{3}{*}{\shortstack{Deep\\Research}}
  & ResearchRubrics~(\citeyear{sharma2025researchrubrics}) & 2,593   & \cmark & \xmark & \xmark & Analytic & Task    \\
  & DRACO~(\citeyear{zhong2026draco})                     & 3,934        & \xmark & \xmark & \cmark & Analytic & Task    \\
  & ReportLogic~(\citeyear{zhao2026reportlogic})         & -         & \xmark & \xmark & \cmark & Analytic & Behav.  \\
\midrule
\multirow{6}{*}{Multi-modal}
  & MTalk-Bench~(\citeyear{du2025mtalkbench})          & 568 & \xmark & \xmark & \cmark & Analytic & Behav.  \\
  & UEval~(\citeyear{li2026ueval})                     & 10,417  & \xmark & \xmark & \cmark & Analytic & Task    \\
  & TechImage-Bench~(\citeyear{ni2026techimagebench})       & 44,131     & \xmark & \cmark & \xmark & Atomic   & Know.   \\
  & AesRM~(\citeyear{han2026aesrm})                     & 2,500 & \cmark & \xmark & \xmark & Analytic & Behav.  \\
  & MMAE~(\citeyear{ma2026mmae}) & 17,741 & \xmark & \xmark & \cmark & Atomic & Task \\
  & PerceptionRubrics~(\citeyear{wei2026perception}) & 12,000 & \xmark & \cmark & \xmark & Atomic & Task \\
\midrule
\multirow{3}{*}{Academic}
  & PaperBench~(\citeyear{starace2025paperbench})           & 8,316       & \cmark & \xmark & \xmark & Analytic & Know.   \\
  & PresentBench~(\citeyear{chen2026presentbench})       & 12,900        & \cmark & \xmark & \xmark & Atomic   & Task    \\
  & TabXEval~(\citeyear{pancholi2026tabxeval})               & 255      & \xmark & \xmark & \cmark & Analytic & Task    \\
  & PPT-EVAL~(\citeyear{gandhi2026ppteval})               & -     & \xmark & \xmark & \cmark & Atomic & Task    \\

\bottomrule
\end{tabular}
\end{table*}

Deep Research benchmarks assess the ability of models and agents to produce long-form research reports in response to complex, open-ended queries, a task that demands multi-step retrieval, synthesis, and structured argumentation. Long-form report evaluation cannot rely on Expert construction for every query, yet automated generation risks the quality degradation documented by General benchmarks. The three Deep Research entries each navigate this trade-off differently: ResearchRubrics~\citep{sharma2025researchrubrics} resolves it by investing 2,800+ hours of human labor into 2,500+ expert-written criteria, closer to the Professional approach. DRACO~\citep{zhong2026draco} and ReportLogic~\citep{zhao2026reportlogic} adopt human-in-the-loop construction, accepting some quality risk in exchange for coverage of authentic user queries sampled from real request logs. However, what unites all three is the discovery that \textbf{implicit criteria represent the hardest evaluation frontier}. ResearchRubrics finds that implicit standards, defined as dimensions users expect but never explicitly state, account for 39.4\% of all criteria, and that even top-tier systems comply with fewer than 68\% of them. ReportLogic further shows that context-aware rubrics, which instantiate each dimension as concrete task-specific checks, significantly outperform generic rubric definitions, suggesting that rubric effectiveness is fundamentally tied to the degree of task specificity encoded in the criteria.

Multi-modal benchmarks extend rubric-based evaluation beyond text to images, video, and speech, targeting tasks such as technical image generation, multimodal reasoning, video aesthetic assessment, and speech dialogue understanding. When evaluation targets shift across modalities, the gap between what rubric designers can specify in advance and what actually determines quality becomes acute. UEval~\citep{li2026ueval} covers eight real-world task types with over 10,000 human-verified rubric criteria, finding that GPT-5-Thinking scores only 66.4 out of 100 and that reasoning models consistently outperform non-reasoning ones, suggesting that rubric compliance in complex multimodal tasks requires genuine inferential capacity rather than pattern matching. TechImage-Bench~\citep{ni2026techimagebench} takes atomic decomposition to its logical extreme, automatically deriving hierarchical rubrics from surrounding text and expanding them into 44,131 binary check items; the best model achieves under 80\% accuracy, revealing that fine-grained scientific precision remains far beyond current capabilities. MMAE~\citep{ma2026mmae} extends the same atomic decomposition to audio editing across seven modalities, with models achieving near-zero exact match rates on complex tasks. MTalk-Bench~\citep{du2025mtalkbench} and AesRM~\citep{han2026aesrm} respond to these limits by adopting Behavior-Grounded rubrics, which are derived not from task specifications but from observed model behaviors and aesthetic judgments, acknowledging that \textbf{some evaluation dimensions resist advance specification and must instead be grounded in empirically observable behavior}.

Academic benchmarks apply rubric-based evaluation to the outputs of the research process itself, including paper reproduction, presentation generation, and table quality assessment, rather than to downstream user interactions. PaperBench~\citep{starace2025paperbench} decomposes ML paper reproduction into 8,316 independently scorable items using hierarchical weighted rubrics co-developed with original authors, replacing the binary success-or-failure verdict of traditional replication with partial-credit evaluation that captures incremental progress. Its adoption into the OpenAI Preparedness Framework, Anthropic Responsible Scaling Policy, and Google DeepMind Frontier Safety Framework signals that \textbf{rubric-based evaluation has become infrastructure for AI governance, not just a research methodology}. Beyond text generation, rubric-based evaluation has also been extended to presentation assessment~\citep{chen2026presentbench,gandhi2026ppteval} and table quality evaluation~\citep{pancholi2026tabxeval}, demonstrating that structured criteria generalize well across diverse forms of academic content.

Three cross-cutting observations emerge from Table~\ref{tab:rubric_benchmarks}. First, the Professional category is the only one where every benchmark adopts Expert construction, reflecting the irreducible role of domain knowledge in high-stakes evaluation. Second, Holistic granularity appears only in BenchBench, where it serves as one component of a dynamic pipeline rather than a standalone scoring mechanism, suggesting that the community has broadly converged on Analytic and Atomic designs as more reliable and actionable. Third, Behavior-Grounded content remains systematically underrepresented, suggesting that constructing rubrics grounded in empirically observable model behavior rather than predefined task specifications remains methodologically challenging and is yet to be adopted at scale.

\subsection{Downstream Applications}\label{7.2application}
As rubric-based evaluation and training have matured, their application has expanded well beyond general-purpose NLP tasks into specialized domains where the stakes of evaluation errors are considerably higher. This section examines five representative downstream application areas: healthcare, law, education, finance and industry, and academic research. Across all five, a common challenge recurs: translating the implicit judgment standards of domain experts into explicit, programmatically verifiable criteria without sacrificing the depth of knowledge that makes those standards meaningful.

\textbf{Healthcare.} The medical domain presents the starkest version of the tension between rubric quality and scalability. HealthBench's 262 physicians and 48,562 hand-written criteria set the quality ceiling~\citep{arora2025healthbench}, but at a construction cost that is by definition non-replicable. Health-SCORE~\citep{yang2026healthscore} addresses this through distillation, clustering expert-written criteria into a reusable general rubric library and dynamically selecting the most relevant subset per query. \citet{shah2026casespecific} further show that after iterative refinement, LLM-generated rubrics can match inter-physician ranking consistency at one-thousandth of the annotation cost, suggesting that the quality-scalability tension requires layered solutions rather than a choice between extremes. \citet{ahmadi2026improvingheart} further extend rubric-based supervision to cardiology question answering, applying variance-aware rubric rewards to post-train small edge-deployed LLMs for specialist medical dialogue.

\textbf{Law.} Legal evaluation introduces a requirement absent from other domains: rubrics must support the auditability of reasoning processes, not just the correctness of outputs. PLawBench~\citep{shi2026plawbench} and PRBench~\citep{akyrek2025prbench} consistently find that models perform relatively well on instruction-following but systematically underperform on process transparency and domain due diligence, which are precisely the dimensions most central to legal practice. LEGIT~\citep{lee2026LEGIT} addresses this by converting court judgments into hierarchical legal issue trees where each node constitutes a verifiable rubric criterion with legal authority backing, enabling layered diagnosis from final conclusion down to individual argument coverage.

\textbf{Education.} The educational context reframes the purpose of rubric-based evaluation: the goal is not to rank outputs but to promote learner improvement, transforming rubrics from passive scoring instruments into active learning scaffolds. iRULER~\citep{bai2026iruler} embodies this shift most clearly, providing Why, Why Not, and How To feedback per criterion alongside a rubric-of-rubrics mechanism that applies evaluative logic recursively to the rubric itself. REC-CBM~\citep{zhao2026reccbm} addresses a complementary challenge of trustworthiness, upgrading rubric dimensions from scoring prompts to mechanistic constraint nodes in the reasoning path through rubric-aware concept encoders and ordinal calibration, enabling educators to directly inspect and intervene in per-dimension scoring. When deployed in real educational settings, \citet{yu2026aigrading} demonstrate feasibility in a real university classroom covering over 1,000 students, while~\citet{favero2026holistic} and~\citet{indel2026trainingdata} extend rubric-based assessment to standardized testing contexts, the latter constructing proxy datasets that replicate protected evaluation corpora under confidentiality constraints.

\textbf{Finance.} Financial evaluation demands rubrics that can handle both the technical precision of quantitative reasoning and the interpretive complexity of open-ended professional judgment. FIRE~\citep{zhang2026fire} addresses this through a dual-track design, pairing 1,000 closed-form questions with direct automated evaluation while providing 2,000 open-ended questions with query-specific rubrics and dedicated scoring models, enabling scalable assessment without sacrificing task specificity. IPO Finance Agent~\citep{benhenda2026ipofinance} extends evaluation to IPO due diligence scenarios, automatically generating benchmark rubrics through an evaluator-optimizer pipeline that iteratively refines candidate criteria before final expert review. PRBench's financial subset contributed by CFA-qualified professionals across 114 countries and 47 jurisdictions~\citep{akyrek2025prbench} further reveals a consistent pattern that cuts across these benchmarks: models perform relatively well on instruction-following but fall significantly short on due diligence and domain-specific reasoning, with top scores reaching only 0.39, suggesting that financial rubric compliance requires not just instruction adherence but genuine domain expertise that current models have yet to internalize.

\textbf{Society.} Rubric-based evaluation has extended into high-throughput social decision-making contexts where structured criteria can replace subjective judgment at scale. In recruitment, \citet{yuksel2026agentic} and~\citet{sun2026comai} adopt rubric-driven multi-agent frameworks to evaluate candidates across dimensions such as technical competency and communication, replacing keyword-matching tools with criteria-grounded assessments that produce transparent and auditable ranking rationales. In news trustworthiness assessment, \citet{zhang2026resource} decompose credibility evaluation into verifiable criteria covering source reliability, factual grounding, and reasoning consistency, enabling scalable automated assessment that would otherwise require expert human review. SCRuB~\citep{watsondaniels2026scrub} further extends rubric-based evaluation to social concept reasoning, where no binary correct answer exists and quality can only be measured by the depth and critical rigor of the response, using a five-dimensional critical thinking rubric and a panel of disciplinary perspectives to measure reasoning depth and critical rigor in place of conventional accuracy metrics.

\textbf{Academic Research.} The scientific research domain applies rubric-based evaluation to the processes that produce and validate knowledge itself. ReviewGrounder~\citep{li2026reviewgrounder} finds that a smaller specialized model with rubric-guided literature grounding outperforms GPT-4.1 across all eight peer review dimensions, demonstrating that structured criteria amplify domain-specific evidence more than raw model scale. MERIT~\citep{yang2026merit} extends rubric-based reasoning to reviewer assignment, generating paper-specific expertise rubrics that decompose the subjective fit judgment into independently verifiable knowledge dimensions, enabling RL-trained evaluators to outperform larger general-purpose LLMs on reviewer matching. DataRubrics~\citep{winata2025datasheet} further reveals that even QA-verified human annotations carry a 26\% error rate on dataset quality assessment, while LLM-based rubric evaluation performs comparably on fine-grained criteria, challenging the assumption that human review is inherently more reliable. Beyond evaluation, \citet{starace2025paperbench} replaces binary replication verdicts with 8,316 partially scorable items, its adoption into three major AI safety frameworks signaling that rubric-based evaluation has become infrastructure for AI governance. This logic extends to data-scarce industrial science: WaferSAGE~\citep{xu2026wafersage} constructs structured rubrics covering defect type, spatial distribution, and root cause analysis to drive a synthetic data pipeline, bringing a 4B-parameter model to near Gemini-3-Flash performance under strict local deployment constraints. \citet{goel2025aicoscientists} push rubric-based approaches further into scientific planning itself, automatically extracting research-objective-specific rubrics as reward signals in a generator-verifier loop, with cross-domain transfer from medical to ML targets demonstrating that the learned planning capabilities generalize beyond the training domain.

Across all these domains, the specific form of the core challenge differs, ranging from scalability in healthcare, auditability in law, and interactivity in education, to transferability in finance, normative consensus in society, and reflexivity in scientific research. Yet the underlying tension remains constant, namely translating the implicit judgment standards of domain experts into explicit, programmatically verifiable criteria without sacrificing the depth of knowledge that makes those standards meaningful.

\begin{summarybox}{Summary}
This chapter examines where rubric-based evaluation and training have been put to use, overviewing representative benchmarks across five categories and downstream applications across six domains, from which three core conclusions can be drawn.
\begin{itemize}
\item[\raisebox{-0.15ex}{\includegraphics[width=1.2em]{image/lightbulb.png}}] The benchmark landscape reveals a consistent convergence away from holistic scoring toward Analytic and Atomic designs, while grounding evaluation criteria in empirically observable model behavior remains an underdeveloped direction.
\item[\raisebox{-0.15ex}{\includegraphics[width=1.2em]{image/lightbulb.png}}] Across all six application domains, the core challenge takes different forms, ranging from scalability in healthcare and auditability in law to normative consensus in society and reflexivity in scientific research, yet the underlying tension between explicit structure and implicit expertise remains constant.
\item[\raisebox{-0.15ex}{\includegraphics[width=1.2em]{image/lightbulb.png}}] The breadth of these applications demonstrates that rubrics are no longer confined to evaluation benchmarks, but are increasingly emerging as a general interface through which human intentions, domain knowledge, and alignment objectives can be translated into actionable machine behavior.
\end{itemize}
\end{summarybox}

\section{Where Does Rubric Research Head?}\label{section8}
Tracing the full arc of rubric research, from its early role as an evaluation auxiliary to its current deep integration into model training and self-evolution pipelines, the boundaries of the rubric paradigm continue to expand rapidly. This chapter discusses five interconnected open questions that we believe will define the future of rubric research across the evolving LLM landscape.

\subsection{Reliability of Rubric Generation}
Rubric generation has long been treated as a preprocessing step rather than a core research object, operating under the implicit assumption that any structured criterion is inherently more reliable than holistic scoring. This assumption has proven increasingly fragile as rubrics are deployed at scale. Low-quality rubrics can go further than simply failing to help, actively pushing reward models toward incorrect preferences and causing more harm than using no rubric at all~\citep{kawabata2026c2, shen2026RRD, qi2026rift}. The gap between model-generated and human-authored rubrics remains substantial even for frontier models~\citep{zhang2026rubricbench}, and this gap reflects deeper misalignments in how models prioritize evaluation dimensions.

The root cause of this problem is structural. Rubric generation relies on the model's own value judgments as a quality anchor, but that anchor is precisely what needs to be calibrated. This creates a self-referential loop that cannot be resolved from within the system. Breaking it requires quality reference points that are independent of the generating model, which is in essence a meta-evaluation problem. A reliable rubric quality metric would need to capture more than whether a rubric discriminates correctly on observed preference pairs. It should also assess robustness under distribution shift and resistance to adversarial manipulation. Constructing such a metric without large-scale human annotation remains one of the most pressing open challenges for scalable rubric deployment, with direct implications for the reliability of the broader rubric-driven ecosystem. This circularity becomes especially pronounced in self-evolution settings, where rubric generation, execution, and training all share the same signal source, making external anchors not optional additions but necessary conditions for reliable rubric-driven systems \citep{pombal2026selfpreferencebias, siro2026GER}

\subsection{Beyond Static Rubric Design}
Recent research work identify that static rubrics face a theoretical ceiling. For any finite fixed set of criteria, there always exists a true reward function for which those criteria yield zero predictive correlation~\citep{gupta2025carmo}. In practice, this manifests as rubric saturation: models learn to satisfy all existing criteria, the reward signal degrades into noise, and reward hacking becomes possible~\citep{xu2026SibylSense, shao2025drtulu}. Dynamic generation addresses this directly, but it introduces a different difficulty. When rubrics change continuously, cross-model and cross-checkpoint comparisons become hard to interpret. Performance improvements may reflect changes in the evaluation standard rather than genuine capability gains. This problem becomes especially acute in long-horizon training runs, where rubric evolution and policy evolution are interleaved.

A layered design offers a promising direction. Abstract-level core dimensions such as factuality, safety, and instruction following possess cross-temporal stability and are well-suited as persistent criteria, whereas instance-level fine-grained checklist items depend heavily on the current ability boundary of the model and may benefit from dynamic adjustment as training progresses. This separation finds partial support in architectures that maintain globally shared criteria alongside task-specific dimensions~\citep{kong2026omni}, but principled methods for governing the boundary between stable and dynamic layers, and for ensuring that dynamic updates improve discrimination remain largely underdeveloped. Formalizing the conditions under which rubric updates constitute genuine evaluation improvement rather than overfitting to the current policy represents an important theoretical problem for the field.

\subsection{Multimodal Extension and Cross-Modal Unification}
Initial evidence suggests that certain evaluation principles transfer across modalities, and that rubric-grounded judgment learned on one modality can improve reward accuracy on unseen modalities~\citep{kong2026omni, jia2026autorubricr1v, yu2026rdpo}. These results motivate the broader goal of a unified rubric framework that supports coherent evaluation across text, image, video, and audio. Extending rubric methodology to multimodal settings is, however, fundamentally more challenging than adapting text-based rubrics to new input formats~\citep{long2026a2rd}. The perceptual mechanisms, quality dimensions, and human judgment standards across modalities are inherently heterogeneous. Evaluating temporal coherence in video involves tracking causal relationships across frames that have no direct analog in static image or text evaluation. Assessing prosodic appropriateness in speech requires sensitivity to paralinguistic features largely absent from text. These differences reflect distinct cognitive faculties rather than surface variation in representation format.

The central open question is how to draw a principled boundary between evaluation criteria that are genuinely modality-agnostic and those that are intrinsically modality-specific. The former constitute transferable alignment signals suitable for a shared rubric layer, while the latter require dedicated design and cannot be meaningfully unified without distortion. This boundary remains empirically underexplored, and current multimodal rubric designs largely reflect engineering intuition rather than validated decompositions. A more rigorous approach would involve systematic studies of how human quality judgments correlate and diverge across modalities on matched content. Establishing this foundation may benefit from engagement with cognitive science and perceptual psychology, as the question of what constitutes quality across sensory modalities is ultimately about human perception and cannot be fully answered from training data alone.

\subsection{From External Criteria to Internalized Mechanisms}
Recent work has shown that rubric generation ability and response generation ability can be co-trained within a single model, with each capability reinforcing the other through a shared parameter space \citep{li2026evolm, xu2026rubricARM, ye2025selfrewardingrubric}. These results suggest that rubrics need not exist solely as externally injected standards but can function as structured carriers of the model's self-assessment capability throughout training, and that joint optimization of generation and evaluation produces more data-efficient alignment than treating the two as separate processes.

This direction raises a deeper question about the nature of the rubric paradigm itself. In its current dominant form, rubrics are delivered via prompts or reward signals and remain outside the model as separate artifacts. This implies a conditional form of alignment: the model behaves according to rubric standards when rubrics are present, but there is no guarantee that this behavior reflects internalized values that persist without explicit guidance. A model that has genuinely internalized rubric-like evaluation standards would apply them spontaneously, catch its own errors before they surface in outputs, and update its internal criteria in response to new information about quality in a given domain. Such a model would be more robust to distribution shift than one that depends on external rubric injection.

The pathway toward this transition remains open. Implicit internalization through large-scale rubric-conditioned training assumes that sufficient exposure to rubric-guided feedback gradually encodes evaluation principles into model weights, analogously to how factual knowledge is absorbed through pretraining. Explicit metacognitive architecture design proposes dedicated self-evaluation components integrated into the forward pass, allowing the model to assess intermediate generation states before committing to a final output. Each direction carries distinct theoretical justifications and engineering challenges, and it remains unclear which will produce more reliable internalization of human-aligned standards. As LLM self-evolution capabilities continue to grow, the role of rubrics is expected to shift progressively from an external evaluation tool toward a constituent element of the model's representational structure. In this sense, the paradigm shift from holistic evaluation to structured criteria described throughout this article remains a work in progress, and its endpoint may be a state in which rubrics are no longer visible as separate artifacts but have become part of how the model understands and monitors its own outputs.
\section{Conclusion}
This work has traced the trajectory of rubrics across the full lifecycle of LLM development, revealing a consistent pattern: each leap in model capability exposes an alignment gap that existing mechanisms cannot bridge, and rubrics have repeatedly proven to be the structural response that closes it. From evaluation instruments that decompose holistic judgments into verifiable criteria, to training signals that extend reinforcement learning beyond verifiable domains, to endogenous mechanisms that co-evolve with the models they govern, rubrics have progressively deepened their integration into the alignment pipeline. Their theoretical limits and practical failure modes, however, remain open challenges that the field has only begun to systematically confront. Ultimately, the trajectory points toward a future in which rubrics are no longer visible as separate artifacts imposed from outside, but have become part of how models understand and monitor their own outputs, serving as a durable foundation for aligning increasingly autonomous AI systems.

\bibliographystyle{abbrvnat}
\bibliography{main}

@article{singh2025gpt5,
  title={Openai gpt-5 system card},
  author={Singh, Aaditya and Fry, Adam and Perelman, Adam and Tart, Adam and Ganesh, Adi and El-Kishky, Ahmed and McLaughlin, Aidan and Low, Aiden and Ostrow, AJ and Ananthram, Akhila and others},
  journal={arXiv preprint arXiv:2601.03267},
  year={2025}
}

@misc{guo2024multiagent,
      title={Large Language Model based Multi-Agents: A Survey of Progress and Challenges}, 
      author={Taicheng Guo and Xiuying Chen and Yaqi Wang and Ruidi Chang and Shichao Pei and Nitesh V. Chawla and Olaf Wiest and Xiangliang Zhang},
      year={2024},
      eprint={2402.01680},
      archivePrefix={arXiv},
      primaryClass={cs.CL},
      url={https://arxiv.org/abs/2402.01680}, 
}

@misc{wu2025agenticreasoning,
      title={Agentic Reasoning: A Streamlined Framework for Enhancing LLM Reasoning with Agentic Tools}, 
      author={Junde Wu and Jiayuan Zhu and Yuyuan Liu and Min Xu and Yueming Jin},
      year={2025},
      eprint={2502.04644},
      archivePrefix={arXiv},
      primaryClass={cs.AI},
      url={https://arxiv.org/abs/2502.04644}, 
}

@misc{kinniment2024evaluating,
      title={Evaluating Language-Model Agents on Realistic Autonomous Tasks}, 
      author={Megan Kinniment and Lucas Jun Koba Sato and Haoxing Du and Brian Goodrich and Max Hasin and Lawrence Chan and Luke Harold Miles and Tao R. Lin and Hjalmar Wijk and Joel Burget and Aaron Ho and Elizabeth Barnes and Paul Christiano},
      year={2024},
      eprint={2312.11671},
      archivePrefix={arXiv},
      primaryClass={cs.CL},
      url={https://arxiv.org/abs/2312.11671}, 
}

@misc{yehudai2026surveyevaluation,
      title={Survey on Evaluation of LLM-based Agents}, 
      author={Asaf Yehudai and Lilach Eden and Alan Li and Guy Uziel and Yilun Zhao and Roy Bar-Haim and Arman Cohan and Michal Shmueli-Scheuer},
      year={2026},
      eprint={2503.16416},
      archivePrefix={arXiv},
      primaryClass={cs.AI},
      url={https://arxiv.org/abs/2503.16416}, 
}

@misc{kim2024prometheus,
      title={Prometheus: Inducing Fine-grained Evaluation Capability in Language Models}, 
      author={Seungone Kim and Jamin Shin and Yejin Cho and Joel Jang and Shayne Longpre and Hwaran Lee and Sangdoo Yun and Seongjin Shin and Sungdong Kim and James Thorne and Minjoon Seo},
      year={2024},
      eprint={2310.08491},
      archivePrefix={arXiv},
      primaryClass={cs.CL},
      url={https://arxiv.org/abs/2310.08491}, 
}

@misc{lambert2024rewardbench,
      title={RewardBench: Evaluating Reward Models for Language Modeling}, 
      author={Nathan Lambert and Valentina Pyatkin and Jacob Morrison and LJ Miranda and Bill Yuchen Lin and Khyathi Chandu and Nouha Dziri and Sachin Kumar and Tom Zick and Yejin Choi and Noah A. Smith and Hannaneh Hajishirzi},
      year={2024},
      eprint={2403.13787},
      archivePrefix={arXiv},
      primaryClass={cs.LG},
      url={https://arxiv.org/abs/2403.13787}, 
}

@misc{bai2022traininghelpful,
      title={Training a Helpful and Harmless Assistant with Reinforcement Learning from Human Feedback}, 
      author={Yuntao Bai and Andy Jones and Kamal Ndousse and Amanda Askell and Anna Chen and Nova DasSarma and Dawn Drain and Stanislav Fort and Deep Ganguli and Tom Henighan and Nicholas Joseph and Saurav Kadavath and Jackson Kernion and Tom Conerly and Sheer El-Showk and Nelson Elhage and Zac Hatfield-Dodds and Danny Hernandez and Tristan Hume and Scott Johnston and Shauna Kravec and Liane Lovitt and Neel Nanda and Catherine Olsson and Dario Amodei and Tom Brown and Jack Clark and Sam McCandlish and Chris Olah and Ben Mann and Jared Kaplan},
      year={2022},
      eprint={2204.05862},
      archivePrefix={arXiv},
      primaryClass={cs.CL},
      url={https://arxiv.org/abs/2204.05862}, 
}

@misc{mahmoud2026rewardhacking,
      title={Reward Hacking in Rubric-Based Reinforcement Learning}, 
      author={Anas Mahmoud and MohammadHossein Rezaei and Zihao Wang and Anisha Gunjal and Bing Liu and Yunzhong He},
      year={2026},
      eprint={2605.12474},
      archivePrefix={arXiv},
      primaryClass={cs.AI},
      url={https://arxiv.org/abs/2605.12474}, 
}

@book{brookhart2013create,
  title={How to create and use rubrics for formative assessment and grading},
  author={Brookhart, Susan M},
  year={2013},
  publisher={Ascd}
}

@misc{lightman2024let,
      title={Let's Verify Step by Step}, 
      author={Hunter Lightman and Vineet Kosaraju and Yura Burda and Harri Edwards and Bowen Baker and Teddy Lee and Jan Leike and John Schulman and Ilya Sutskever and Karl Cobbe},
      year={2023},
      eprint={2305.20050},
      archivePrefix={arXiv},
      primaryClass={cs.LG},
      url={https://arxiv.org/abs/2305.20050}, 
}

@misc{yuan2025Miracle,
      title={Curing Miracle Steps in LLM Mathematical Reasoning with Rubric Rewards}, 
      author={Youliang Yuan and Qiuyang Mang and Jingbang Chen and Hong Wan and Xiaoyuan Liu and Junjielong Xu and Jen-tse Huang and Wenxuan Wang and Wenxiang Jiao and Pinjia He},
      year={2026},
      eprint={2510.07774},
      archivePrefix={arXiv},
      primaryClass={cs.CL},
      url={https://arxiv.org/abs/2510.07774}, 
}

@misc{liu2023geval,
      title={G-Eval: NLG Evaluation using GPT-4 with Better Human Alignment}, 
      author={Yang Liu and Dan Iter and Yichong Xu and Shuohang Wang and Ruochen Xu and Chenguang Zhu},
      year={2023},
      eprint={2303.16634},
      archivePrefix={arXiv},
      primaryClass={cs.CL},
      url={https://arxiv.org/abs/2303.16634}, 
}

@misc{arora2025healthbench,
      title={HealthBench: Evaluating Large Language Models Towards Improved Human Health}, 
      author={Rahul K. Arora and Jason Wei and Rebecca Soskin Hicks and Preston Bowman and Joaquin Quiñonero-Candela and Foivos Tsimpourlas and Michael Sharman and Meghan Shah and Andrea Vallone and Alex Beutel and Johannes Heidecke and Karan Singhal},
      year={2025},
      eprint={2505.08775},
      archivePrefix={arXiv},
      primaryClass={cs.CL},
      url={https://arxiv.org/abs/2505.08775}, 
}

@article{jia2026autorubricr1v,
  title={AutoRubric-R1V: Rubric-Based Generative Rewards for Faithful Multimodal Reasoning},
  author={Jia, Mengzhao and Zhang, Zhihan and Cases, Ignacio and Liu, Zheyuan and Jiang, Meng and Qi, Peng},
  journal={arXiv preprint arXiv:2510.14738},
  year={2025}
}

@misc{chen2026rmr1,
      title={RM-R1: Reward Modeling as Reasoning}, 
      author={Xiusi Chen and Gaotang Li and Ziqi Wang and Bowen Jin and Cheng Qian and Yu Wang and Hongru Wang and Yu Zhang and Denghui Zhang and Tong Zhang and Hanghang Tong and Heng Ji},
      year={2026},
      eprint={2505.02387},
      archivePrefix={arXiv},
      primaryClass={cs.CL},
      url={https://arxiv.org/abs/2505.02387}, 
}

@inproceedings{brookhart2018appropriate,
  title={Appropriate criteria: Key to effective rubrics},
  author={Brookhart, Susan M},
  booktitle={Frontiers in education},
  volume={3},
  pages={22},
  year={2018},
  organization={Frontiers Media SA}
}

@article{hunter1996use,
  title={The use of holistic versus analytic scoring for large-scale assessment of writing},
  author={Hunter, Darryl M and Jones, Richard M and Randhawa, Bikkar S},
  journal={Canadian Journal of Program Evaluation},
  volume={11},
  number={2},
  pages={61--86},
  year={1996},
  publisher={University of Toronto Press Toronto, ON}
}

@misc{zheng2023mtbench,
      title={Judging LLM-as-a-Judge with MT-Bench and Chatbot Arena}, 
      author={Lianmin Zheng and Wei-Lin Chiang and Ying Sheng and Siyuan Zhuang and Zhanghao Wu and Yonghao Zhuang and Zi Lin and Zhuohan Li and Dacheng Li and Eric P. Xing and Hao Zhang and Joseph E. Gonzalez and Ion Stoica},
      year={2023},
      eprint={2306.05685},
      archivePrefix={arXiv},
      primaryClass={cs.CL},
      url={https://arxiv.org/abs/2306.05685}, 
}

@misc{fan2025sedareval,
      title={SedarEval: Automated Evaluation using Self-Adaptive Rubrics}, 
      author={Zhiyuan Fan and Weinong Wang and Xing Wu and Debing Zhang},
      year={2025},
      eprint={2501.15595},
      archivePrefix={arXiv},
      primaryClass={cs.CV},
      url={https://arxiv.org/abs/2501.15595}, 
}

@misc{viswanathan2025rlcf,
      title={Checklists Are Better Than Reward Models For Aligning Language Models}, 
      author={Vijay Viswanathan and Yanchao Sun and Shuang Ma and Xiang Kong and Meng Cao and Graham Neubig and Tongshuang Wu},
      year={2025},
      eprint={2507.18624},
      archivePrefix={arXiv},
      primaryClass={cs.CL},
      url={https://arxiv.org/abs/2507.18624}, 
}

@misc{wang2024mathshepherd,
      title={Math-Shepherd: Verify and Reinforce LLMs Step-by-step without Human Annotations}, 
      author={Peiyi Wang and Lei Li and Zhihong Shao and R. X. Xu and Damai Dai and Yifei Li and Deli Chen and Y. Wu and Zhifang Sui},
      year={2024},
      eprint={2312.08935},
      archivePrefix={arXiv},
      primaryClass={cs.AI},
      url={https://arxiv.org/abs/2312.08935}, 
}

@inproceedings{Hashemi_2024llmrubric,
   title={LLM-Rubric: A Multidimensional, Calibrated Approach to Automated Evaluation of Natural Language Texts},
   url={http://dx.doi.org/10.18653/v1/2024.acl-long.745},
   DOI={10.18653/v1/2024.acl-long.745},
   booktitle={Proceedings of the 62nd Annual Meeting of the Association for Computational Linguistics (Volume 1: Long Papers)},
   publisher={Association for Computational Linguistics},
   author={Hashemi, Helia and Eisner, Jason and Rosset, Corby and Van Durme, Benjamin and Kedzie, Chris},
   year={2024},
   pages={13806–13834} }

@misc{wang2026criticrubric,
      title={A Rubric-Supervised Critic from Sparse Real-World Outcomes}, 
      author={Xingyao Wang and Valerie Chen and Heng Ji and Graham Neubig},
      year={2026},
      eprint={2603.03800},
      archivePrefix={arXiv},
      primaryClass={cs.AI},
      url={https://arxiv.org/abs/2603.03800}, 
}

@misc{ding2026adarubric,
      title={AdaRubric: Task-Adaptive Rubrics for Reliable LLM Agent Evaluation and Reward Learning}, 
      author={Liang Ding},
      year={2026},
      eprint={2603.21362},
      archivePrefix={arXiv},
      primaryClass={cs.AI},
      url={https://arxiv.org/abs/2603.21362}, 
}

@misc{mu2024RBR,
      title={Rule Based Rewards for Language Model Safety}, 
      author={Tong Mu and Alec Helyar and Johannes Heidecke and Joshua Achiam and Andrea Vallone and Ian Kivlichan and Molly Lin and Alex Beutel and John Schulman and Lilian Weng},
      year={2024},
      eprint={2411.01111},
      archivePrefix={arXiv},
      primaryClass={cs.AI},
      url={https://arxiv.org/abs/2411.01111}, 
}

@misc{sheng2026RLCER,
      title={Reinforcing Chain-of-Thought Reasoning with Self-Evolving Rubrics}, 
      author={Leheng Sheng and Wenchang Ma and Ruixin Hong and Xiang Wang and An Zhang and Tat-Seng Chua},
      year={2026},
      eprint={2602.10885},
      archivePrefix={arXiv},
      primaryClass={cs.AI},
      url={https://arxiv.org/abs/2602.10885}, 
}

@misc{ma2025search-Gen-V,
      title={An Efficient Rubric-based Generative Verifier for Search-Augmented LLMs}, 
      author={Linyue Ma and Yilong Xu and Xiang Long and Zhi Zheng},
      year={2025},
      eprint={2510.14660},
      archivePrefix={arXiv},
      primaryClass={cs.CL},
      url={https://arxiv.org/abs/2510.14660}, 
}

@misc{zhang2026CaRR,
      title={Chaining the Evidence: Robust Reinforcement Learning for Deep Search Agents with Citation-Aware Rubric Rewards}, 
      author={Jiajie Zhang and Xin Lv and Ling Feng and Lei Hou and Juanzi Li},
      year={2026},
      eprint={2601.06021},
      archivePrefix={arXiv},
      primaryClass={cs.CL},
      url={https://arxiv.org/abs/2601.06021}, 
}

@misc{wan2026DeepVerifier,
      title={Inference-Time Scaling of Verification: Self-Evolving Deep Research Agents via Test-Time Rubric-Guided Verification}, 
      author={Yuxuan Wan and Tianqing Fang and Zaitang Li and Yintong Huo and Wenxuan Wang and Haitao Mi and Dong Yu and Michael R. Lyu},
      year={2026},
      eprint={2601.15808},
      archivePrefix={arXiv},
      primaryClass={cs.AI},
      url={https://arxiv.org/abs/2601.15808}, 
}

@misc{xie2026autorubric,
      title={Auto-Rubric: Learning From Implicit Weights to Explicit Rubrics for Reward Modeling}, 
      author={Lipeng Xie and Sen Huang and Zhuo Zhang and Anni Zou and Yunpeng Zhai and Dingchao Ren and Kezun Zhang and Haoyuan Hu and Boyin Liu and Haoran Chen and Zhaoyang Liu and Bolin Ding},
      year={2026},
      eprint={2510.17314},
      archivePrefix={arXiv},
      primaryClass={cs.LG},
      url={https://arxiv.org/abs/2510.17314}, 
}

@misc{rezaei2025onlinerubric,
      title={Online Rubrics Elicitation from Pairwise Comparisons}, 
      author={MohammadHossein Rezaei and Robert Vacareanu and Zihao Wang and Clinton Wang and Bing Liu and Yunzhong He and Afra Feyza Akyürek},
      year={2025},
      eprint={2510.07284},
      archivePrefix={arXiv},
      primaryClass={cs.CL},
      url={https://arxiv.org/abs/2510.07284}, 
}

@misc{wadhwa2025evalagent,
      title={EvalAgent: Discovering Implicit Evaluation Criteria from the Web}, 
      author={Manya Wadhwa and Zayne Sprague and Chaitanya Malaviya and Philippe Laban and Junyi Jessy Li and Greg Durrett},
      year={2025},
      eprint={2504.15219},
      archivePrefix={arXiv},
      primaryClass={cs.CL},
      url={https://arxiv.org/abs/2504.15219}, 
}

@misc{sharma2025researchrubrics,
      title={ResearchRubrics: A Benchmark of Prompts and Rubrics For Evaluating Deep Research Agents}, 
      author={Manasi Sharma and Chen Bo Calvin Zhang and Chaithanya Bandi and Clinton Wang and Ankit Aich and Huy Nghiem and Tahseen Rabbani and Ye Htet and Brian Jang and Sumana Basu and Aishwarya Balwani and Denis Peskoff and Marcos Ayestaran and Sean M. Hendryx and Brad Kenstler and Bing Liu},
      year={2025},
      eprint={2511.07685},
      archivePrefix={arXiv},
      primaryClass={cs.AI},
      url={https://arxiv.org/abs/2511.07685}, 
}

@misc{fan2026agent-RRM,
      title={Exploring Reasoning Reward Model for Agents}, 
      author={Kaixuan Fan and Kaituo Feng and Manyuan Zhang and Tianshuo Peng and Zhixun Li and Yilei Jiang and Shuang Chen and Peng Pei and Xunliang Cai and Xiangyu Yue},
      year={2026},
      eprint={2601.22154},
      archivePrefix={arXiv},
      primaryClass={cs.AI},
      url={https://arxiv.org/abs/2601.22154}, 
}

@misc{sanders2026generatingdatadriven,
      title={Generating Data-Driven Reasoning Rubrics for Domain-Adaptive Reward Modeling}, 
      author={Kate Sanders and Nathaniel Weir and Sapana Chaudhary and Kaj Bostrom and Huzefa Rangwala},
      year={2026},
      eprint={2602.06795},
      archivePrefix={arXiv},
      primaryClass={cs.CL},
      url={https://arxiv.org/abs/2602.06795}, 
}

@misc{lee2026LEGIT,
      title={Evaluating Legal Reasoning Traces with Legal Issue Tree Rubrics}, 
      author={Jinu Lee and Kyoung-Woon On and Simeng Han and Arman Cohan and Julia Hockenmaier},
      year={2026},
      eprint={2512.01020},
      archivePrefix={arXiv},
      primaryClass={cs.AI},
      url={https://arxiv.org/abs/2512.01020}, 
}

@misc{akyrek2025prbench,
      title={PRBench: Large-Scale Expert Rubrics for Evaluating High-Stakes Professional Reasoning}, 
      author={Afra Feyza Akyürek and Advait Gosai and Chen Bo Calvin Zhang and Vipul Gupta and Jaehwan Jeong and Anisha Gunjal and Tahseen Rabbani and Maria Mazzone and David Randolph and Mohammad Mahmoudi Meymand and Gurshaan Chattha and Paula Rodriguez and Diego Mares and Pavit Singh and Michael Liu and Subodh Chawla and Pete Cline and Lucy Ogaz and Ernesto Hernandez and Zihao Wang and Pavi Bhatter and Marcos Ayestaran and Bing Liu and Yunzhong He},
      year={2025},
      eprint={2511.11562},
      archivePrefix={arXiv},
      primaryClass={cs.CL},
      url={https://arxiv.org/abs/2511.11562}, 
}

@misc{bai2022constitutionalai,
      title={Constitutional AI: Harmlessness from AI Feedback}, 
      author={Yuntao Bai and Saurav Kadavath and Sandipan Kundu and Amanda Askell and Jackson Kernion and Andy Jones and Anna Chen and Anna Goldie and Azalia Mirhoseini and Cameron McKinnon and Carol Chen and Catherine Olsson and Christopher Olah and Danny Hernandez and Dawn Drain and Deep Ganguli and Dustin Li and Eli Tran-Johnson and Ethan Perez and Jamie Kerr and Jared Mueller and Jeffrey Ladish and Joshua Landau and Kamal Ndousse and Kamile Lukosuite and Liane Lovitt and Michael Sellitto and Nelson Elhage and Nicholas Schiefer and Noemi Mercado and Nova DasSarma and Robert Lasenby and Robin Larson and Sam Ringer and Scott Johnston and Shauna Kravec and Sheer El Showk and Stanislav Fort and Tamera Lanham and Timothy Telleen-Lawton and Tom Conerly and Tom Henighan and Tristan Hume and Samuel R. Bowman and Zac Hatfield-Dodds and Ben Mann and Dario Amodei and Nicholas Joseph and Sam McCandlish and Tom Brown and Jared Kaplan},
      year={2022},
      eprint={2212.08073},
      archivePrefix={arXiv},
      primaryClass={cs.CL},
      url={https://arxiv.org/abs/2212.08073}, 
}

@misc{ouyang2022rlhf,
      title={Training language models to follow instructions with human feedback}, 
      author={Long Ouyang and Jeff Wu and Xu Jiang and Diogo Almeida and Carroll L. Wainwright and Pamela Mishkin and Chong Zhang and Sandhini Agarwal and Katarina Slama and Alex Ray and John Schulman and Jacob Hilton and Fraser Kelton and Luke Miller and Maddie Simens and Amanda Askell and Peter Welinder and Paul Christiano and Jan Leike and Ryan Lowe},
      year={2022},
      eprint={2203.02155},
      archivePrefix={arXiv},
      primaryClass={cs.CL},
      url={https://arxiv.org/abs/2203.02155}, 
}

@misc{wei2023cot,
      title={Chain-of-Thought Prompting Elicits Reasoning in Large Language Models}, 
      author={Jason Wei and Xuezhi Wang and Dale Schuurmans and Maarten Bosma and Brian Ichter and Fei Xia and Ed Chi and Quoc Le and Denny Zhou},
      year={2023},
      eprint={2201.11903},
      archivePrefix={arXiv},
      primaryClass={cs.CL},
      url={https://arxiv.org/abs/2201.11903}, 
}

@article{openai2024gpt4,
  title={Gpt-4 technical report},
  author={Achiam, Josh and Adler, Steven and Agarwal, Sandhini and Ahmad, Lama and Akkaya, Ilge and Aleman, Florencia Leoni and Almeida, Diogo and Altenschmidt, Janko and Altman, Sam and Anadkat, Shyamal and others},
  journal={arXiv preprint arXiv:2303.08774},
  year={2023}
}

@misc{shao2024rlvr,
      title={DeepSeekMath: Pushing the Limits of Mathematical Reasoning in Open Language Models}, 
      author={Zhihong Shao and Peiyi Wang and Qihao Zhu and Runxin Xu and Junxiao Song and Xiao Bi and Haowei Zhang and Mingchuan Zhang and Y. K. Li and Y. Wu and Daya Guo},
      year={2024},
      eprint={2402.03300},
      archivePrefix={arXiv},
      primaryClass={cs.CL},
      url={https://arxiv.org/abs/2402.03300}, 
}

@misc{gupta2025carmo,
      title={CARMO: Dynamic Criteria Generation for Context-Aware Reward Modelling}, 
      author={Taneesh Gupta and Shivam Shandilya and Xuchao Zhang and Rahul Madhavan and Supriyo Ghosh and Chetan Bansal and Huaxiu Yao and Saravan Rajmohan},
      year={2025},
      eprint={2410.21545},
      archivePrefix={arXiv},
      primaryClass={cs.CL},
      url={https://arxiv.org/abs/2410.21545}, 
}

@article{jaech2024openaio1,
  title={Openai o1 system card},
  author={Jaech, Aaron and Kalai, Adam and Lerer, Adam and Richardson, Adam and El-Kishky, Ahmed and Low, Aiden and Helyar, Alec and Madry, Aleksander and Beutel, Alex and Carney, Alex and others},
  journal={arXiv preprint arXiv:2412.16720},
  year={2024}
}

@misc{ma2024agentic,
      title={AgentBoard: An Analytical Evaluation Board of Multi-turn LLM Agents}, 
      author={Chang Ma and Junlei Zhang and Zhihao Zhu and Cheng Yang and Yujiu Yang and Yaohui Jin and Zhenzhong Lan and Lingpeng Kong and Junxian He},
      year={2024},
      eprint={2401.13178},
      archivePrefix={arXiv},
      primaryClass={cs.CL},
      url={https://arxiv.org/abs/2401.13178}, 
}

@article{guo2025deepseekr1,
  title={Deepseek-r1: Incentivizing reasoning capability in llms via reinforcement learning},
  author={Guo, Daya and Yang, Dejian and Zhang, Haowei and Song, Junxiao and Wang, Peiyi and Zhu, Qihao and Xu, Runxin and Zhang, Ruoyu and Ma, Shirong and Bi, Xiao and others},
  journal={arXiv preprint arXiv:2501.12948},
  year={2025}
}

@inproceedings{Dineen_2025QALIGN,
   title={QA‐LIGN: Aligning LLMs through Constitutionally Decomposed QA},
   url={http://dx.doi.org/10.18653/v1/2025.findings-emnlp.1123},
   DOI={10.18653/v1/2025.findings-emnlp.1123},
   booktitle={Findings of the Association for Computational Linguistics: EMNLP 2025},
   publisher={Association for Computational Linguistics},
   author={Dineen, Jacob and Rrv, Aswin and Liu, Qin and Xu, Zhikun and Ye, Xiao and Shen, Ming and Li, Zhaonan and Lu, Shijie and Baral, Chitta and Chen, Muhao and Zhou, Ben},
   year={2025},
   pages={20619–20642} }

@misc{galvansosa2025rubrikscube,
      title={Rubrik's Cube: Testing a New Rubric for Evaluating Explanations on the CUBE dataset}, 
      author={Diana Galvan-Sosa and Gabrielle Gaudeau and Pride Kavumba and Yunmeng Li and Hongyi gu and Zheng Yuan and Keisuke Sakaguchi and Paula Buttery},
      year={2025},
      eprint={2503.23899},
      archivePrefix={arXiv},
      primaryClass={cs.CL},
      url={https://arxiv.org/abs/2503.23899}, 
}

@misc{li2025leveraging,
      title={Leveraging LLMs as Meta-Judges: A Multi-Agent Framework for Evaluating LLM Judgments}, 
      author={Yuran Li and Jama Hussein Mohamud and Chongren Sun and Di Wu and Benoit Boulet},
      year={2025},
      eprint={2504.17087},
      archivePrefix={arXiv},
      primaryClass={cs.AI},
      url={https://arxiv.org/abs/2504.17087}, 
}

@misc{huang2025rubicon,
      title={Reinforcement Learning with Rubric Anchors}, 
      author={Zenan Huang and Yihong Zhuang and Guoshan Lu and Zeyu Qin and Haokai Xu and Tianyu Zhao and Ru Peng and Jiaqi Hu and Zhanming Shen and Xiaomeng Hu and Xijun Gu and Peiyi Tu and Jiaxin Liu and Wenyu Chen and Yuzhuo Fu and Zhiting Fan and Yanmei Gu and Yuanyuan Wang and Zhengkai Yang and Jianguo Li and Junbo Zhao},
      year={2025},
      eprint={2508.12790},
      archivePrefix={arXiv},
      primaryClass={cs.AI},
      url={https://arxiv.org/abs/2508.12790}, 
}

@misc{goel2025aicoscientists,
      title={Training AI Co-Scientists Using Rubric Rewards}, 
      author={Shashwat Goel and Rishi Hazra and Dulhan Jayalath and Timon Willi and Parag Jain and William F. Shen and Ilias Leontiadis and Francesco Barbieri and Yoram Bachrach and Jonas Geiping and Chenxi Whitehouse},
      year={2025},
      eprint={2512.23707},
      archivePrefix={arXiv},
      primaryClass={cs.LG},
      url={https://arxiv.org/abs/2512.23707}, 
}

@misc{kong2026omni,
      title={Omni-RRM: Advancing Omni Reward Modeling via Automatic Rubric-Grounded Preference Synthesis}, 
      author={Zicheng Kong and Dehua Ma and Zhenbo Xu and Alven Yang and Yiwei Ru and Haoran Wang and Zixuan Zhou and Fuqing Bie and Liuyu Xiang and Huijia Wu and Jian Zhao and Zhaofeng He},
      year={2026},
      eprint={2602.00846},
      archivePrefix={arXiv},
      primaryClass={cs.CL},
      url={https://arxiv.org/abs/2602.00846}, 
}

@misc{zhang2026rubricbench,
      title={RubricBench: Aligning Model-Generated Rubrics with Human Standards}, 
      author={Qiyuan Zhang and Junyi Zhou and Yufei Wang and Fuyuan Lyu and Yidong Ming and Can Xu and Qingfeng Sun and Kai Zheng and Peng Kang and Xue Liu and Chen Ma},
      year={2026},
      eprint={2603.01562},
      archivePrefix={arXiv},
      primaryClass={cs.AI},
      url={https://arxiv.org/abs/2603.01562}, 
}

@misc{xu2026rubricARM,
      title={Alternating Reinforcement Learning for Rubric-Based Reward Modeling in Non-Verifiable LLM Post-Training}, 
      author={Ran Xu and Tianci Liu and Zihan Dong and Tony Yu and Ilgee Hong and Carl Yang and Linjun Zhang and Tao Zhao and Haoyu Wang},
      year={2026},
      eprint={2602.01511},
      archivePrefix={arXiv},
      primaryClass={cs.CL},
      url={https://arxiv.org/abs/2602.01511}, 
}

@misc{qi2026rift,
      title={RIFT: A RubrIc Failure Mode Taxonomy and Automated Diagnostics}, 
      author={Zhengyang Qi and Charles Dickens and Derek Pham and Amanda Dsouza and Armin Parchami and Frederic Sala and Paroma Varma},
      year={2026},
      eprint={2604.01375},
      archivePrefix={arXiv},
      primaryClass={cs.AI},
      url={https://arxiv.org/abs/2604.01375}, 
}

@misc{pombal2026selfpreferencebias,
      title={Self-Preference Bias in Rubric-Based Evaluation of Large Language Models}, 
      author={José Pombal and Ricardo Rei and André F. T. Martins},
      year={2026},
      eprint={2604.06996},
      archivePrefix={arXiv},
      primaryClass={cs.CL},
      url={https://arxiv.org/abs/2604.06996}, 
}

@misc{shen2026RRD,
      title={Rethinking Rubric Generation for Improving LLM Judge and Reward Modeling for Open-ended Tasks}, 
      author={William F. Shen and Xinchi Qiu and Chenxi Whitehouse and Lisa Alazraki and Shashwat Goel and Francesco Barbieri and Timon Willi and Akhil Mathur and Ilias Leontiadis},
      year={2026},
      eprint={2602.05125},
      archivePrefix={arXiv},
      primaryClass={cs.LG},
      url={https://arxiv.org/abs/2602.05125}, 
}

@misc{kawabata2026c2,
      title={C2: Scalable Rubric-Augmented Reward Modeling from Binary Preferences}, 
      author={Akira Kawabata and Saku Sugawara},
      year={2026},
      eprint={2604.13618},
      archivePrefix={arXiv},
      primaryClass={cs.CL},
      url={https://arxiv.org/abs/2604.13618}, 
}

@misc{wang2025profbench,
      title={ProfBench: Multi-Domain Rubrics requiring Professional Knowledge to Answer and Judge}, 
      author={Zhilin Wang and Jaehun Jung and Ximing Lu and Shizhe Diao and Ellie Evans and Jiaqi Zeng and Pavlo Molchanov and Yejin Choi and Jan Kautz and Yi Dong},
      year={2025},
      eprint={2510.18941},
      archivePrefix={arXiv},
      primaryClass={cs.CL},
      url={https://arxiv.org/abs/2510.18941}, 
}

@misc{liu2026xpertbench,
      title={Xpertbench: Expert Level Tasks with Rubrics-Based Evaluation}, 
      author={Xue Liu and Xin Ma and Yuxin Ma and Yongchang Peng and Duo Wang and Zhoufutu Wen and Ge Zhang and Kaiyuan Zhang and Xinyu Chen and Yida Ding and Tianci He and Jiani Hou and Liang Hu and Ziyun Huang and Yongzhe Hui and Jianpeng Jiao and Chennan Ju and Yingru Kong and Yiran Li and Jiashuo Liu and Mengyun Liu and Luyao Ma and Fei Ni and Yiqing Ni and Pengbo Niu and Yueyan Qiu and Yanle Ren and Xinyu Shen and Zilin Shi and Zaiyuan Wang and Wenjie Yue and Chun Zhang and Shiyu Zhang and Xinyi Zhang and Kaiwen Zhao and Zhenwei Zhu and Shanshan Wu and Qi Zhao and Wenhao Huang},
      year={2026},
      eprint={2604.02368},
      archivePrefix={arXiv},
      primaryClass={cs.AI},
      url={https://arxiv.org/abs/2604.02368}, 
}

@misc{ruan2025expertbench,
      title={ExpertLongBench: Benchmarking Language Models on Expert-Level Long-Form Generation Tasks with Structured Checklists}, 
      author={Jie Ruan and Inderjeet Nair and Shuyang Cao and Amy Liu and Sheza Munir and Micah Pollens-Dempsey and Tiffany Chiang and Lucy Kates and Nicholas David and Sihan Chen and Ruxin Yang and Yuqian Yang and Jasmine Gump and Tessa Bialek and Vivek Sankaran and Margo Schlanger and Lu Wang},
      year={2025},
      eprint={2506.01241},
      archivePrefix={arXiv},
      primaryClass={cs.CL},
      url={https://arxiv.org/abs/2506.01241}, 
}

@misc{han2026aesrm,
      title={AesRM: Improving Video Aesthetics with Expert-Level Feedback}, 
      author={Yujin Han and Yujie Wei and Yefei He and Xinyu Liu and Tianle Li and Zichao Yu and Andi Han and Shiwei Zhang and Tingyu Weng and Difan Zou},
      year={2026},
      eprint={2604.28078},
      archivePrefix={arXiv},
      primaryClass={cs.CV},
      url={https://arxiv.org/abs/2604.28078}, 
}

@misc{cook2024tick,
      title={TICKing All the Boxes: Generated Checklists Improve LLM Evaluation and Generation}, 
      author={Jonathan Cook and Tim Rocktäschel and Jakob Foerster and Dennis Aumiller and Alex Wang},
      year={2024},
      eprint={2410.03608},
      archivePrefix={arXiv},
      primaryClass={cs.AI},
      url={https://arxiv.org/abs/2410.03608}, 
}

@misc{li2026rubrichub,
      title={RubricHub: A Comprehensive and Highly Discriminative Rubric Dataset via Automated Coarse-to-Fine Generation}, 
      author={Sunzhu Li and Jiale Zhao and Miteto Wei and Huimin Ren and Yang Zhou and Jingwen Yang and Shunyu Liu and Kaike Zhang and Wei Chen},
      year={2026},
      eprint={2601.08430},
      archivePrefix={arXiv},
      primaryClass={cs.AI},
      url={https://arxiv.org/abs/2601.08430}, 
}

@misc{gao2026qworld,
      title={Qworld: Question-Specific Evaluation Criteria for LLMs}, 
      author={Shanghua Gao and Yuchang Su and Pengwei Sui and Curtis Ginder and Marinka Zitnik},
      year={2026},
      eprint={2603.23522},
      archivePrefix={arXiv},
      primaryClass={cs.CL},
      url={https://arxiv.org/abs/2603.23522}, 
}

@misc{liu2026openrubric,
      title={OpenRubrics: Towards Scalable Synthetic Rubric Generation for Reward Modeling and LLM Alignment}, 
      author={Tianci Liu and Ran Xu and Tony Yu and Ilgee Hong and Carl Yang and Tuo Zhao and Haoyu Wang},
      year={2026},
      eprint={2510.07743},
      archivePrefix={arXiv},
      primaryClass={cs.CL},
      url={https://arxiv.org/abs/2510.07743}, 
}

@misc{liu2026cdrrm,
      title={CDRRM: Contrast-Driven Rubric Generation for Reliable and Interpretable Reward Modeling}, 
      author={Dengcan Liu and Fengkai Yang and Xiaohan Wang and Shurui Yan and Jiajun Chai and Jiahao Li and Yikun Ban and Zhendong Mao and Wei Lin and Guojun Yin},
      year={2026},
      eprint={2603.08035},
      archivePrefix={arXiv},
      primaryClass={cs.AI},
      url={https://arxiv.org/abs/2603.08035}, 
}

@misc{gunjal2025rar,
      title={Rubrics as Rewards: Reinforcement Learning Beyond Verifiable Domains}, 
      author={Anisha Gunjal and Anthony Wang and Elaine Lau and Vaskar Nath and Yunzhong He and Bing Liu and Sean Hendryx},
      year={2025},
      eprint={2507.17746},
      archivePrefix={arXiv},
      primaryClass={cs.LG},
      url={https://arxiv.org/abs/2507.17746}, 
}

@misc{yifei2025researchqa,
      title={ResearchQA: Evaluating Scholarly Question Answering at Scale Across 75 Fields with Survey-Mined Questions and Rubrics}, 
      author={Li S. Yifei and Allen Chang and Chaitanya Malaviya and Mark Yatskar},
      year={2025},
      eprint={2509.00496},
      archivePrefix={arXiv},
      primaryClass={cs.CL},
      url={https://arxiv.org/abs/2509.00496}, 
}

@inproceedings{
wei2026qurl,
title={Qu{RL}: Rubrics As Judge For Open-Ended Question Answering},
author={Xiyu Wei and Qingwei Zong and Xiaoguang Li and Eugene J. Yu and Sujian Li},
booktitle={The Fourteenth International Conference on Learning Representations},
year={2026},
url={https://openreview.net/forum?id=DrhWTuhtYq}
}

@misc{ni2026techimagebench,
      title={TechImage-Bench: Rubric-Based Evaluation for Technical Image Generation}, 
      author={Minheng Ni and Zhengyuan Yang and Yaowen Zhang and Linjie Li and Chung-Ching Lin and Kevin Lin and Zhendong Wang and Xiaofei Wang and Shujie Liu and Lei Zhang and Wangmeng Zuo and Lijuan Wang},
      year={2026},
      eprint={2512.12220},
      archivePrefix={arXiv},
      primaryClass={cs.CV},
      url={https://arxiv.org/abs/2512.12220}, 
}

@misc{dhole2026rubricrag,
      title={RubricRAG: Towards Interpretable and Reliable LLM Evaluation via Domain Knowledge Retrieval for Rubric Generation}, 
      author={Kaustubh D. Dhole and Eugene Agichtein},
      year={2026},
      eprint={2603.20882},
      archivePrefix={arXiv},
      primaryClass={cs.IR},
      url={https://arxiv.org/abs/2603.20882}, 
}

@misc{jayalath2026cat,
      title={Compute as Teacher: Turning Inference Compute Into Reference-Free Supervision}, 
      author={Dulhan Jayalath and Shashwat Goel and Thomas Foster and Parag Jain and Suchin Gururangan and Cheng Zhang and Anirudh Goyal and Alan Schelten},
      year={2026},
      eprint={2509.14234},
      archivePrefix={arXiv},
      primaryClass={cs.LG},
      url={https://arxiv.org/abs/2509.14234}, 
}

@misc{feng2025sage,
      title={Are We on the Right Way to Assessing LLM-as-a-Judge?}, 
      author={Yuanning Feng and Sinan Wang and Zhengxiang Cheng and Yao Wan and Dongping Chen},
      year={2025},
      eprint={2512.16041},
      archivePrefix={arXiv},
      primaryClass={cs.CL},
      url={https://arxiv.org/abs/2512.16041}, 
}

@misc{lv2026learningquery,
      title={Learning Query-Specific Rubrics from Human Preferences for DeepResearch Report Generation}, 
      author={Changze Lv and Jie Zhou and Wentao Zhao and Jingwen Xu and Zisu Huang and Muzhao Tian and Shihan Dou and Tao Gui and Le Tian and Xiao Zhou and Xiaoqing Zheng and Xuanjing Huang and Jie Zhou},
      year={2026},
      eprint={2602.03619},
      archivePrefix={arXiv},
      primaryClass={cs.CL},
      url={https://arxiv.org/abs/2602.03619}, 
}

@misc{shao2025drtulu,
      title={DR Tulu: Reinforcement Learning with Evolving Rubrics for Deep Research}, 
      author={Rulin Shao and Akari Asai and Shannon Zejiang Shen and Hamish Ivison and Varsha Kishore and Jingming Zhuo and Xinran Zhao and Molly Park and Samuel G. Finlayson and David Sontag and Tyler Murray and Sewon Min and Pradeep Dasigi and Luca Soldaini and Faeze Brahman and Wen-tau Yih and Tongshuang Wu and Luke Zettlemoyer and Yoon Kim and Hannaneh Hajishirzi and Pang Wei Koh},
      year={2025},
      eprint={2511.19399},
      archivePrefix={arXiv},
      primaryClass={cs.CL},
      url={https://arxiv.org/abs/2511.19399}, 
}

@misc{huang2026PoP,
      title={Bootstrapping Post-training Signals for Open-ended Tasks via Rubric-based Self-play on Pre-training Text}, 
      author={Chengyu Huang and Sheng-Yen Chou and Zhengxin Zhang and Claire Cardie},
      year={2026},
      eprint={2604.20051},
      archivePrefix={arXiv},
      primaryClass={cs.CL},
      url={https://arxiv.org/abs/2604.20051}, 
}

@misc{shi2025humanintheloop,
      title={Towards a Human-in-the-Loop Framework for Reliable Patch Evaluation Using an LLM-as-a-Judge}, 
      author={Sherry Shi and Renyao Wei and Michele Tufano and José Cambronero and Runxiang Cheng and Franjo Ivančić and Pat Rondon},
      year={2025},
      eprint={2511.10865},
      archivePrefix={arXiv},
      primaryClass={cs.SE},
      url={https://arxiv.org/abs/2511.10865}, 
}

@misc{masters2025arcane,
      title={ARCANE: A Multi-Agent Framework for Interpretable and Configurable Alignment}, 
      author={Charlie Masters and Marta Grześkiewicz and Stefano V. Albrecht},
      year={2025},
      eprint={2512.06196},
      archivePrefix={arXiv},
      primaryClass={cs.AI},
      url={https://arxiv.org/abs/2512.06196}, 
}

@misc{shah2026casespecific,
      title={Case-Specific Rubrics for Clinical AI Evaluation: Methodology, Validation, and LLM-Clinician Agreement Across 823 Encounters}, 
      author={Aaryan Shah and Andrew Hines and Alexia Downs and Denis Bajet and Paulius Mui and Fabiano Araujo and Laura Offutt and Aida Rutledge and Elizabeth Jimenez},
      year={2026},
      eprint={2604.24710},
      archivePrefix={arXiv},
      primaryClass={cs.AI},
      url={https://arxiv.org/abs/2604.24710}, 
}

@misc{li2026reviewgrounder,
      title={ReviewGrounder: Improving Review Substantiveness with Rubric-Guided, Tool-Integrated Agents}, 
      author={Zhuofeng Li and Yi Lu and Dongfu Jiang and Haoxiang Zhang and Yuyang Bai and Chuan Li and Yu Wang and Shuiwang Ji and Jianwen Xie and Yu Zhang},
      year={2026},
      eprint={2604.14261},
      archivePrefix={arXiv},
      primaryClass={cs.CL},
      url={https://arxiv.org/abs/2604.14261}, 
}

@misc{fan2026Optimsyn,
      title={Optimsyn: Influence-Guided Rubrics Optimization for Synthetic Data Generation}, 
      author={Zhiting Fan and Ruizhe Chen and Tianxiang Hu and Ru Peng and Zenan Huang and Haokai Xu and Yixin Chen and Jian Wu and Junbo Zhao and Zuozhu Liu},
      year={2026},
      eprint={2604.00536},
      archivePrefix={arXiv},
      primaryClass={cs.CL},
      url={https://arxiv.org/abs/2604.00536}, 
}

@misc{chu2026CARO,
      title={Confusion-Aware Rubric Optimization for LLM-based Automated Grading}, 
      author={Yucheng Chu and Hang Li and Kaiqi Yang and Yasemin Copur-Gencturk and Joseph Krajcik and Namsoo Shin and Jiliang Tang},
      year={2026},
      eprint={2603.00451},
      archivePrefix={arXiv},
      primaryClass={cs.AI},
      url={https://arxiv.org/abs/2603.00451}, 
}

@misc{xu2026SibylSense,
      title={SibylSense: Adaptive Rubric Learning via Memory Tuning and Adversarial Probing}, 
      author={Yifei Xu and Guilherme Potje and Shivam Shandilya and Tiancheng Yuan and Leonardo de Oliveira Nunes and Rakshanda Agarwal and Saeid Asgari and Adam Atkinson and Emre Kıcıman and Songwu Lu and Ranveer Chandra and Tusher Chakraborty},
      year={2026},
      eprint={2602.20751},
      archivePrefix={arXiv},
      primaryClass={cs.CL},
      url={https://arxiv.org/abs/2602.20751}, 
}

@misc{qiu2026Proxy-GRM,
      title={Rationale Matters: Learning Transferable Rubrics via Proxy-Guided Critique for VLM Reward Models}, 
      author={Weijie Qiu and Dai Guan and Junxin Wang and Zhihang Li and Yongbo Gai and Mengyu Zhou and Erchao Zhao and Xiaoxi Jiang and Guanjun Jiang},
      year={2026},
      eprint={2603.16600},
      archivePrefix={arXiv},
      primaryClass={cs.CV},
      url={https://arxiv.org/abs/2603.16600}, 
}

@misc{harada2025Reflect-and-Revise,
      title={Automated Refinement of Essay Scoring Rubrics for Language Models via Reflect-and-Revise}, 
      author={Keno Harada and Lui Yoshida and Takeshi Kojima and Yusuke Iwasawa and Yutaka Matsuo},
      year={2025},
      eprint={2510.09030},
      archivePrefix={arXiv},
      primaryClass={cs.CL},
      url={https://arxiv.org/abs/2510.09030}, 
}

@misc{bai2026iruler,
      title={iRULER: Intelligible Rubric-Based User-Defined LLM Evaluation for Revision}, 
      author={Jingwen Bai and Wei Soon Cheong and Philippe Muller and Brian Y Lim},
      year={2026},
      eprint={2602.12779},
      archivePrefix={arXiv},
      primaryClass={cs.HC},
      url={https://arxiv.org/abs/2602.12779}, 
}

@misc{li2026coreflect,
      title={CoReflect: Conversational Evaluation via Co-Evolutionary Simulation and Reflective Rubric Refinement}, 
      author={Yunzhe Li and Richie Yueqi Feng and Tianxin Wei and Chin-Chia Hsu},
      year={2026},
      eprint={2601.12208},
      archivePrefix={arXiv},
      primaryClass={cs.CL},
      url={https://arxiv.org/abs/2601.12208}, 
}

@misc{wang2025infimedorbit,
      title={InfiMed-ORBIT: Aligning LLMs on Open-Ended Complex Tasks via Rubric-Based Incremental Training}, 
      author={Pengkai Wang and Linus and Pengwei Liu and Zhijie Sang and Congkai Xie and Hongxia Yang},
      year={2025},
      eprint={2510.15859},
      archivePrefix={arXiv},
      primaryClass={cs.CL},
      url={https://arxiv.org/abs/2510.15859}, 
}

@misc{li2026evolm,
      title={EvoLM: Self-Evolving Language Models through Co-Evolved Discriminative Rubrics}, 
      author={Shuyue Stella Li and Rui Xin and Teng Xiao and Yike Wang and Rulin Shao and Zoey Hao and Melanie Sclar and Sewoong Oh and Faeze Brahman and Pang Wei Koh and Yulia Tsvetkov},
      year={2026},
      eprint={2605.03871},
      archivePrefix={arXiv},
      primaryClass={cs.AI},
      url={https://arxiv.org/abs/2605.03871}, 
}

@misc{ye2025selfrewardingrubric,
      title={Self-Rewarding Rubric-Based Reinforcement Learning for Open-Ended Reasoning}, 
      author={Zhiling Ye and Yun Yue and Haowen Wang and Xudong Han and Jiadi Jiang and Cheng Wei and Lei Fan and Jiaxin Liang and Shuowen Zhang and Ji Li and Chunxiao Guo and Jian Wang and Peng Wei and Jinjie Gu},
      year={2025},
      eprint={2509.25534},
      archivePrefix={arXiv},
      primaryClass={cs.CL},
      url={https://arxiv.org/abs/2509.25534}, 
}

@misc{wu2025rlac,
      title={RLAC: Reinforcement Learning with Adversarial Critic for Free-Form Generation Tasks}, 
      author={Mian Wu and Gavin Zhang and Sewon Min and Sergey Levine and Aviral Kumar},
      year={2025},
      eprint={2511.01758},
      archivePrefix={arXiv},
      primaryClass={cs.LG},
      url={https://arxiv.org/abs/2511.01758}, 
}

@misc{rao2026autorubric,
      title={Autorubric: Unifying Rubric-based LLM Evaluation}, 
      author={Delip Rao and Chris Callison-Burch},
      year={2026},
      eprint={2603.00077},
      archivePrefix={arXiv},
      primaryClass={cs.CL},
      url={https://arxiv.org/abs/2603.00077}, 
}

@misc{du2025mtalkbench,
      title={MTalk-Bench: Evaluating Speech-to-Speech Models in Multi-Turn Dialogues via Arena-style and Rubrics Protocols}, 
      author={Yuhao Du and Qianwei Huang and Guo Zhu and Zhanchen Dai and Shunian Chen and Qiming Zhu and Le Pan and Minghao Chen and Yuhao Zhang and Li Zhou and Benyou Wang and Haizhou Li},
      year={2025},
      eprint={2508.18240},
      archivePrefix={arXiv},
      primaryClass={cs.CL},
      url={https://arxiv.org/abs/2508.18240}, 
}

@misc{favero2026holistic,
      title={Beyond Holistic Scores: Automatic Trait-Based Quality Scoring of Argumentative Essays}, 
      author={Lucile Favero and Juan Antonio Pérez-Ortiz and Tanja Käser and Nuria Oliver},
      year={2026},
      eprint={2602.04604},
      archivePrefix={arXiv},
      primaryClass={cs.CL},
      url={https://arxiv.org/abs/2602.04604}, 
}

@inproceedings{Pathak_2025,
   title={Rubric Is All You Need: Improving LLM-Based Code Evaluation With Question-Specific Rubrics},
   url={http://dx.doi.org/10.1145/3702652.3744220},
   DOI={10.1145/3702652.3744220},
   booktitle={Proceedings of the 2025 ACM Conference on International Computing Education Research V.1},
   publisher={ACM},
   author={Pathak, Aditya and Gandhi, Rachit and Uttam, Vaibhav and Ramamoorthy, Arnav and Ghosh, Pratyush and Jindal, Aaryan Raj and Verma, Shreyash and Mittal, Aditya and Ased, Aashna and Khatri, Chirag and Nakka, Yashwanth and Devansh and Challa, Jagat Sesh and Kumar, Dhruv},
   year={2025},
   month=Aug, pages={181–195},
   collection={ICER ’25} }

@misc{zhang2026rethinking,
      title={Rethinking Atomic Decomposition for LLM Judges: A Prompt-Controlled Study of Reference-Grounded QA Evaluation}, 
      author={Xinran Zhang},
      year={2026},
      eprint={2603.28005},
      archivePrefix={arXiv},
      primaryClass={cs.CL},
      url={https://arxiv.org/abs/2603.28005}, 
}

@misc{mallinar2026scalable,
      title={A Scalable Framework for Evaluating Health Language Models}, 
      author={Neil Mallinar and A. Ali Heydari and Xin Liu and Anthony Z. Faranesh and Brent Winslow and Nova Hammerquist and Benjamin Graef and Cathy Speed and Mark Malhotra and Shwetak Patel and Javier L. Prieto and Daniel McDuff and Ahmed A. Metwally},
      year={2026},
      eprint={2503.23339},
      archivePrefix={arXiv},
      primaryClass={cs.AI},
      url={https://arxiv.org/abs/2503.23339}, 
}

@misc{hong2026rulers,
      title={RULERS: Locked Rubrics and Evidence-Anchored Scoring for Robust LLM Evaluation}, 
      author={Yihan Hong and Huaiyuan Yao and Bolin Shen and Wanpeng Xu and Hua Wei and Yushun Dong},
      year={2026},
      eprint={2601.08654},
      archivePrefix={arXiv},
      primaryClass={cs.CL},
      url={https://arxiv.org/abs/2601.08654}, 
}

@misc{yu2025DeCE,
      title={Beyond Pointwise Scores: Decomposed Criteria-Based Evaluation of LLM Responses}, 
      author={Fangyi Yu and Nabeel Seedat and Dasha Herrmannova and Frank Schilder and Jonathan Richard Schwarz},
      year={2025},
      eprint={2509.16093},
      archivePrefix={arXiv},
      primaryClass={cs.CL},
      url={https://arxiv.org/abs/2509.16093}, 
}

@misc{amin2026Co-Creation,
      title={LLM-as-a-Judge for Human-AI Co-Creation: A Reliability-Aware Evaluation Framework for Coding}, 
      author={Md Faizul Ibne Amin and Yutaka Watanobe and Daniel M. Muepu and Haruto Suzuki and Kenta Nanaumi and Md Mostafizer Rahman},
      year={2026},
      eprint={2604.27727},
      archivePrefix={arXiv},
      primaryClass={cs.SE},
      url={https://arxiv.org/abs/2604.27727}, 
}

@misc{huynh2026quantify,
      title={Quantifying the Statistical Effect of Rubric Modifications on Human-Autorater Agreement}, 
      author={Jessica Huynh and Alfredo Gomez and Athiya Deviyani and Renee Shelby and Jeffrey P. Bigham and Fernando Diaz},
      year={2026},
      eprint={2605.06283},
      archivePrefix={arXiv},
      primaryClass={cs.CL},
      url={https://arxiv.org/abs/2605.06283}, 
}

@misc{pan2026rubriceval,
      title={RubricEval: A Rubric-Level Meta-Evaluation Benchmark for LLM Judges in Instruction Following}, 
      author={Tianjun Pan and Xuan Lin and Wenyan Yang and Qianyu He and Shisong Chen and Licai Qi and Wanqing Xu and Hongwei Feng and Bo Xu and Yanghua Xiao},
      year={2026},
      eprint={2603.25133},
      archivePrefix={arXiv},
      primaryClass={cs.AI},
      url={https://arxiv.org/abs/2603.25133}, 
}

@misc{chen2026criterion,
      title={Criterion Validity of LLM-as-Judge for Business Outcomes in Conversational Commerce}, 
      author={Liang Chen and Qi Liu and Wenhuan Lin and Feng Liang},
      year={2026},
      eprint={2604.00022},
      archivePrefix={arXiv},
      primaryClass={cs.CL},
      url={https://arxiv.org/abs/2604.00022}, 
}

@misc{li2026evaluatingscoringbias,
      title={Evaluating Scoring Bias in LLM-as-a-Judge}, 
      author={Qingquan Li and Shaoyu Dou and Kailai Shao and Chao Chen and Haixiang Hu},
      year={2026},
      eprint={2506.22316},
      archivePrefix={arXiv},
      primaryClass={cs.CL},
      url={https://arxiv.org/abs/2506.22316}, 
}

@misc{xu2026ipointwise,
      title={Am I More Pointwise or Pairwise? Revealing Position Bias in Rubric-Based LLM-as-a-Judge}, 
      author={Yuzheng Xu and Tosho Hirasawa and Tadashi Kozuno and Yoshitaka Ushiku},
      year={2026},
      eprint={2602.02219},
      archivePrefix={arXiv},
      primaryClass={cs.CL},
      url={https://arxiv.org/abs/2602.02219}, 
}

@misc{li2025curseknowledge,
      title={Curse of Knowledge: When Complex Evaluation Context Benefits yet Biases LLM Judges}, 
      author={Weiyuan Li and Xintao Wang and Siyu Yuan and Rui Xu and Jiangjie Chen and Qingqing Dong and Yanghua Xiao and Deqing Yang},
      year={2025},
      eprint={2509.03419},
      archivePrefix={arXiv},
      primaryClass={cs.CL},
      url={https://arxiv.org/abs/2509.03419}, 
}

@misc{weng2026BeyondAccuracy,
      title={Beyond Accuracy: Policy Invariance as a Reliability Test for LLM Safety Judges}, 
      author={Shihao Weng and Yang Feng and Xiaofei Xie},
      year={2026},
      eprint={2605.06161},
      archivePrefix={arXiv},
      primaryClass={cs.AI},
      url={https://arxiv.org/abs/2605.06161}, 
}

@misc{mittal2026TRACE,
      title={Comparing Developer and LLM Biases in Code Evaluation}, 
      author={Aditya Mittal and Ryan Shar and Zichu Wu and Shyam Agarwal and Tongshuang Wu and Chris Donahue and Ameet Talwalkar and Wayne Chi and Valerie Chen},
      year={2026},
      eprint={2603.24586},
      archivePrefix={arXiv},
      primaryClass={cs.SE},
      url={https://arxiv.org/abs/2603.24586}, 
}

@misc{yang2026fairjudge,
      title={FairJudge: An Adaptive, Debiased, and Consistent LLM-as-a-Judge}, 
      author={Bo Yang and Lanfei Feng and Yunkui Chen and Yu Zhang and Xiao Xu and Shijian Li},
      year={2026},
      eprint={2602.06625},
      archivePrefix={arXiv},
      primaryClass={cs.CL},
      url={https://arxiv.org/abs/2602.06625}, 
}

@misc{chu2026GUIDE,
      title={Optimizing In-Context Demonstrations for LLM-based Automated Grading}, 
      author={Yucheng Chu and Hang Li and Kaiqi Yang and Yasemin Copur-Gencturk and Kevin Haudek and Joseph Krajcik and Jiliang Tang},
      year={2026},
      eprint={2603.00465},
      archivePrefix={arXiv},
      primaryClass={cs.AI},
      url={https://arxiv.org/abs/2603.00465}, 
}

@misc{deng2025rubricconditioned,
      title={Rubric-Conditioned LLM Grading: Alignment, Uncertainty, and Robustness}, 
      author={Haotian Deng and Chris Farber and Jiyoon Lee and David Tang},
      year={2025},
      eprint={2601.08843},
      archivePrefix={arXiv},
      primaryClass={cs.CL},
      url={https://arxiv.org/abs/2601.08843}, 
}

@misc{raghavendra2026agenticrubrics,
      title={Agentic Rubrics as Contextual Verifiers for SWE Agents}, 
      author={Mohit Raghavendra and Anisha Gunjal and Bing Liu and Yunzhong He},
      year={2026},
      eprint={2601.04171},
      archivePrefix={arXiv},
      primaryClass={cs.LG},
      url={https://arxiv.org/abs/2601.04171}, 
}

@misc{si2026ctx2skill,
      title={From Context to Skills: Can Language Models Learn from Context Skillfully?}, 
      author={Shuzheng Si and Haozhe Zhao and Yu Lei and Qingyi Wang and Dingwei Chen and Zhitong Wang and Zhenhailong Wang and Kangyang Luo and Zheng Wang and Gang Chen and Fanchao Qi and Minjia Zhang and Maosong Sun},
      year={2026},
      eprint={2604.27660},
      archivePrefix={arXiv},
      primaryClass={cs.AI},
      url={https://arxiv.org/abs/2604.27660}, 
}

@misc{fang2026rubricbasedonpolicy,
      title={Rubric-based On-policy Distillation}, 
      author={Junfeng Fang and Zhepei Hong and Mao Zheng and Mingyang Song and Gengsheng Li and Houcheng Jiang and Dan Zhang and Haiyun Guo and Xiang Wang and Tat-Seng Chua},
      year={2026},
      eprint={2605.07396},
      archivePrefix={arXiv},
      primaryClass={cs.LG},
      url={https://arxiv.org/abs/2605.07396}, 
}

@misc{yu2026thinkwithrubrics,
      title={Think-with-Rubrics: From External Evaluator to Internal Reasoning Guidance}, 
      author={Jiachen Yu and Zhihao Xu and Junjie Wang and Yujiu Yang},
      year={2026},
      eprint={2605.07461},
      archivePrefix={arXiv},
      primaryClass={cs.CL},
      url={https://arxiv.org/abs/2605.07461}, 
}

@misc{liu2026deltarubric,
      title={DeltaRubric: Generative Multimodal Reward Modeling via Joint Planning and Verification}, 
      author={Rui Liu and Dian Yu and Zhenwen Liang and Yucheng Shi and Tong Zheng and Runpeng Dai and Haitao Mi and Pratap Tokekar and Leoweiliang},
      year={2026},
      eprint={2605.09269},
      archivePrefix={arXiv},
      primaryClass={cs.CL},
      url={https://arxiv.org/abs/2605.09269}, 
}

@misc{nagar2026clrvoyance,
      title={CLR-voyance: Reinforcing Open-Ended Reasoning for Inpatient Clinical Decision Support with Outcome-Aware Rubrics}, 
      author={Aishik Nagar and Arun-Kumar Kaliya-Perumal and Yu-Hsuan Han and Andrew Sheng-Han Huang and Kristen Kee and Yushi Cao and Yiming Chen and Hongchao Jiang},
      year={2026},
      eprint={2605.09584},
      archivePrefix={arXiv},
      primaryClass={cs.CL},
      url={https://arxiv.org/abs/2605.09584}, 
}

@misc{wu2026amarismemory,
      title={AMARIS: A Memory-Augmented Rubric Improvement System for Rubric-Based Reinforcement Learning}, 
      author={Peilin Wu and Xinlu Zhang and Kun Wan and Wentian Zhao and Gang Wu and Xinya Du and Zhiyu Chen},
      year={2026},
      eprint={2605.18592},
      archivePrefix={arXiv},
      primaryClass={cs.LG},
      url={https://arxiv.org/abs/2605.18592}, 
}

@misc{levine2026rubricrefine,
      title={RubricRefine: Improving Tool-Use Agent Reliability with Training-Free Pre-Execution Refinement}, 
      author={Will LeVine and Brendan Evers and Sam Saltwick and Abhay Venkatesh},
      year={2026},
      eprint={2605.09730},
      archivePrefix={arXiv},
      primaryClass={cs.LG},
      url={https://arxiv.org/abs/2605.09730}, 
}

@misc{han2026swetrace,
      title={SWE-TRACE: Optimizing Long-Horizon SWE Agents Through Rubric Process Reward Models and Heuristic Test-Time Scaling}, 
      author={Hao Han and Jin Xie and Xuehao Ma and Weiquan Zhu and Ziyao Zhang and ZhiLiang Long and Hongkai Chen and Qingwen Ye},
      year={2026},
      eprint={2604.14820},
      archivePrefix={arXiv},
      primaryClass={cs.SE},
      url={https://arxiv.org/abs/2604.14820}, 
}

@misc{zhong2026draco,
      title={DRACO: a Cross-Domain Benchmark for Deep Research Accuracy, Completeness, and Objectivity}, 
      author={Joey Zhong and Hao Zhang and Clare Southern and Jeremy Yang and Thomas Wang and Kate Jung and Shu Zhang and Denis Yarats and Johnny Ho and Jerry Ma},
      year={2026},
      eprint={2602.11685},
      archivePrefix={arXiv},
      primaryClass={cs.LG},
      url={https://arxiv.org/abs/2602.11685}, 
}

@misc{dubois2025alpacaeval,
      title={Length-Controlled AlpacaEval: A Simple Way to Debias Automatic Evaluators}, 
      author={Yann Dubois and Balázs Galambosi and Percy Liang and Tatsunori B. Hashimoto},
      year={2025},
      eprint={2404.04475},
      archivePrefix={arXiv},
      primaryClass={cs.LG},
      url={https://arxiv.org/abs/2404.04475}, 
}

@misc{zhang2026chasingtail,
      title={Chasing the Tail: Effective Rubric-based Reward Modeling for Large Language Model Post-Training}, 
      author={Junkai Zhang and Zihao Wang and Lin Gui and Swarnashree Mysore Sathyendra and Jaehwan Jeong and Victor Veitch and Wei Wang and Yunzhong He and Bing Liu and Lifeng Jin},
      year={2026},
      eprint={2509.21500},
      archivePrefix={arXiv},
      primaryClass={cs.LG},
      url={https://arxiv.org/abs/2509.21500}, 
}

@inproceedings{lianget2025generative,
    title = "Generative Reward Modeling via Synthetic Criteria Preference Learning",
    author = "Liang, Xiaobo  and
      Zhang, Haoke  and
      Li, Juntao  and
      Chen, Kehai  and
      Zhu, Qiaoming  and
      Zhang, Min",
    editor = "Che, Wanxiang  and
      Nabende, Joyce  and
      Shutova, Ekaterina  and
      Pilehvar, Mohammad Taher",
    booktitle = "Proceedings of the 63rd Annual Meeting of the Association for Computational Linguistics (Volume 1: Long Papers)",
    month = jul,
    year = "2025",
    address = "Vienna, Austria",
    publisher = "Association for Computational Linguistics",
    url = "https://aclanthology.org/2025.acl-long.1297/",
    doi = "10.18653/v1/2025.acl-long.1297",
    pages = "26755--26769",
    ISBN = "979-8-89176-251-0"
}

@misc{jia2026openrs,
      title={Open Rubric System: Scaling Reinforcement Learning with Pairwise Adaptive Rubric}, 
      author={Ruipeng Jia and Yunyi Yang and Yuxin Wu and Yongbo Gai and Siyuan Tao and Mengyu Zhou and Jianhe Lin and Xiaoxi Jiang and Guanjun Jiang},
      year={2026},
      eprint={2602.14069},
      archivePrefix={arXiv},
      primaryClass={cs.CL},
      url={https://arxiv.org/abs/2602.14069}, 
}

@misc{tan2026papo,
      title={PAPO: Stabilizing Rubric Integration Training via Decoupled Advantage Normalization}, 
      author={Zelin Tan and Zhouliang Yu and Bohan Lin and Zijie Geng and Hejia Geng and Yudong Zhang and Mulei Zhang and Yang Chen and Shuyue Hu and Zhenfei Yin and Chen Zhang and Lei Bai},
      year={2026},
      eprint={2603.26535},
      archivePrefix={arXiv},
      primaryClass={cs.AI},
      url={https://arxiv.org/abs/2603.26535}, 
}

@misc{srivastava2025crome,
      title={Robust Reward Modeling via Causal Rubrics}, 
      author={Pragya Srivastava and Harman Singh and Rahul Madhavan and Gandharv Patil and Sravanti Addepalli and Arun Suggala and Rengarajan Aravamudhan and Soumya Sharma and Anirban Laha and Aravindan Raghuveer and Karthikeyan Shanmugam and Doina Precup},
      year={2025},
      eprint={2506.16507},
      archivePrefix={arXiv},
      primaryClass={cs.LG},
      url={https://arxiv.org/abs/2506.16507}, 
}

@misc{bhattarai2026rubricgroundedrl,
      title={Rubric-Grounded RL: Structured Judge Rewards for Generalizable Reasoning}, 
      author={Manish Bhattarai and Ismael Boureima and Nishath Rajiv Ranasinghe and Scott Pakin and Dan O'Malley},
      year={2026},
      eprint={2605.08061},
      archivePrefix={arXiv},
      primaryClass={cs.AI},
      url={https://arxiv.org/abs/2605.08061}, 
}

@misc{xie2026SRaR,
      title={Step-wise Rubric Rewards for LLM Reasoning}, 
      author={Weichu Xie and Haozhe Zhao and Wenpu Liu and Yongfu Zhu and Liang Chen and Minghao Ye and Zirong Chen and Yuqi Xu and Shuai Dong and Ziyue Wang and Xinbo Xu and Kean Shi and Ruoyu Wu and Xiaoying Zhang and Wenqi Shao and Baobao Chang and Nan Duan and Jiaqi Wang},
      year={2026},
      eprint={2605.17291},
      archivePrefix={arXiv},
      primaryClass={cs.LG},
      url={https://arxiv.org/abs/2605.17291}, 
}

@misc{shao2025deepseekmathv2,
      title={DeepSeekMath-V2: Towards Self-Verifiable Mathematical Reasoning}, 
      author={Zhihong Shao and Yuxiang Luo and Chengda Lu and Z. Z. Ren and Jiewen Hu and Tian Ye and Zhibin Gou and Shirong Ma and Xiaokang Zhang},
      year={2025},
      eprint={2511.22570},
      archivePrefix={arXiv},
      primaryClass={cs.AI},
      url={https://arxiv.org/abs/2511.22570}, 
}

@misc{xu2026rtt,
      title={Rubrics to Tokens: Bridging Response-level Rubrics and Token-level Rewards in Instruction Following Tasks}, 
      author={Tianze Xu and Yanzhao Zheng and Pengrui Lu and Lyumanshan Ye and Yong Wu and Zhentao Zhang and Yuanqiang Yu and Chao Ma and Jihuai Zhu and Pengfei Liu and Baohua Dong and Hangcheng Zhu and Ruohui Huang and Gang Yu},
      year={2026},
      eprint={2604.02795},
      archivePrefix={arXiv},
      primaryClass={cs.CL},
      url={https://arxiv.org/abs/2604.02795}, 
}

@misc{chen2026improvingdatareward,
      title={Improving Data and Reward Design for Scientific Reasoning in Large Language Models}, 
      author={Zijie Chen and Zhenghao Lin and Xiao Liu and Zhenzhong Lan and Yeyun Gong and Peng Cheng},
      year={2026},
      eprint={2602.08321},
      archivePrefix={arXiv},
      primaryClass={cs.CL},
      url={https://arxiv.org/abs/2602.08321}, 
}

@misc{jia2025writingzero,
      title={Writing-Zero: Bridge the Gap Between Non-verifiable Tasks and Verifiable Rewards}, 
      author={Ruipeng Jia and Yunyi Yang and Yongbo Gai and Kai Luo and Shihao Huang and Jianhe Lin and Xiaoxi Jiang and Guanjun Jiang},
      year={2025},
      eprint={2506.00103},
      archivePrefix={arXiv},
      primaryClass={cs.CL},
      url={https://arxiv.org/abs/2506.00103}, 
}

@misc{chen2025acerl,
      title={ACE-RL: Adaptive Constraint-Enhanced Reward for Long-form Generation Reinforcement Learning}, 
      author={Jianghao Chen and Wei Sun and Qixiang Yin and Zhixing Tan and Jiajun Zhang},
      year={2025},
      eprint={2509.04903},
      archivePrefix={arXiv},
      primaryClass={cs.CL},
      url={https://arxiv.org/abs/2509.04903}, 
}

@misc{bi2025RGRGRPO,
      title={Reward and Guidance through Rubrics: Promoting Exploration to Improve Multi-Domain Reasoning}, 
      author={Baolong Bi and Shenghua Liu and Yiwei Wang and Siqian Tong and Lingrui Mei and Yuyao Ge and Yilong Xu and Jiafeng Guo and Xueqi Cheng},
      year={2025},
      eprint={2511.12344},
      archivePrefix={arXiv},
      primaryClass={cs.AI},
      url={https://arxiv.org/abs/2511.12344}, 
}

@misc{zhou2026Bottleneck,
      title={Breaking the Exploration Bottleneck: Rubric-Scaffolded Reinforcement Learning for General LLM Reasoning}, 
      author={Yang Zhou and Sunzhu Li and Shunyu Liu and Wenkai Fang and Kongcheng Zhang and Jiale Zhao and Jingwen Yang and Yihe Zhou and Jianwei Lv and Tongya Zheng and Hengtong Lu and Wei Chen and Yan Xie and Mingli Song},
      year={2026},
      eprint={2508.16949},
      archivePrefix={arXiv},
      primaryClass={cs.LG},
      url={https://arxiv.org/abs/2508.16949}, 
}

@misc{he2025advancedif,
      title={AdvancedIF: Rubric-Based Benchmarking and Reinforcement Learning for Advancing LLM Instruction Following}, 
      author={Yun He and Wenzhe Li and Hejia Zhang and Songlin Li and Karishma Mandyam and Sopan Khosla and Yuanhao Xiong and Nanshu Wang and Xiaoliang Peng and Beibin Li and Shengjie Bi and Shishir G. Patil and Qi Qi and Shengyu Feng and Julian Katz-Samuels and Richard Yuanzhe Pang and Sujan Gonugondla and Hunter Lang and Yue Yu and Yundi Qian and Maryam Fazel-Zarandi and Licheng Yu and Amine Benhalloum and Hany Awadalla and Manaal Faruqui},
      year={2025},
      eprint={2511.10507},
      archivePrefix={arXiv},
      primaryClass={cs.CL},
      url={https://arxiv.org/abs/2511.10507}, 
}

@misc{li2026stitchcuda,
      title={StitchCUDA: An Automated Multi-Agents End-to-End GPU Programing Framework with Rubric-based Agentic Reinforcement Learning}, 
      author={Shiyang Li and Zijian Zhang and Winson Chen and Yuebo Luo and Mingyi Hong and Caiwen Ding},
      year={2026},
      eprint={2603.02637},
      archivePrefix={arXiv},
      primaryClass={cs.MA},
      url={https://arxiv.org/abs/2603.02637}, 
}

@misc{li2026rubricem,
      title={RubricEM: Meta-RL with Rubric-guided Policy Decomposition beyond Verifiable Rewards}, 
      author={Gaotang Li and Bhavana Dalvi Mishra and Zifeng Wang and Jun Yan and Yanfei Chen and Chun-Liang Li and Long T. Le and Rujun Han and George Lee and Hanghang Tong and Chen-Yu Lee and Tomas Pfister},
      year={2026},
      eprint={2605.10899},
      archivePrefix={arXiv},
      primaryClass={cs.CL},
      url={https://arxiv.org/abs/2605.10899}, 
}

@misc{yuan2025kardiar1,
      title={Kardia-R1: Unleashing LLMs to Reason toward Understanding and Empathy for Emotional Support via Rubric-as-Judge Reinforcement Learning}, 
      author={Jiahao Yuan and Zhiqing Cui and Hanqing Wang and Yuansheng Gao and Yucheng Zhou and Usman Naseem},
      year={2025},
      eprint={2512.01282},
      archivePrefix={arXiv},
      primaryClass={cs.CL},
      url={https://arxiv.org/abs/2512.01282}, 
}

@misc{su2026judger1,
      title={Enhancing Judgment Document Generation via Agentic Legal Information Collection and Rubric-Guided Optimization}, 
      author={Weihang Su and Xuanyi Chen and Yueyue Wu and Qingyao Ai and Yiqun Liu},
      year={2026},
      eprint={2605.02011},
      archivePrefix={arXiv},
      primaryClass={cs.CL},
      url={https://arxiv.org/abs/2605.02011}, 
}

@misc{xu2026wafersage,
      title={WaferSAGE: Large Language Model-Powered Wafer Defect Analysis via Synthetic Data Generation and Rubric-Guided Reinforcement Learning}, 
      author={Ke Xu and Zhongyuan Lian},
      year={2026},
      eprint={2604.27629},
      archivePrefix={arXiv},
      primaryClass={cs.AI},
      url={https://arxiv.org/abs/2604.27629}, 
}

@misc{gupta2026atlas,
      title={Scaling Agentic Capabilities, Not Context: Efficient Reinforcement Finetuning for Large Toolspaces}, 
      author={Karan Gupta and Pranav Vajreshwari and Yash Pandya and Raghav Magazine and Akshay Nambi and Ahmed Awadallah},
      year={2026},
      eprint={2603.06713},
      archivePrefix={arXiv},
      primaryClass={cs.LG},
      url={https://arxiv.org/abs/2603.06713}, 
}

@misc{zhao2026askaskbench,
      title={When and What to Ask: AskBench and Rubric-Guided RLVR for LLM Clarification}, 
      author={Jiale Zhao and Ke Fang and Lu Cheng},
      year={2026},
      eprint={2602.11199},
      archivePrefix={arXiv},
      primaryClass={cs.CL},
      url={https://arxiv.org/abs/2602.11199}, 
}

@misc{chen2026rucl,
      title={RuCL: Stratified Rubric-Based Curriculum Learning for Multimodal Large Language Model Reasoning}, 
      author={Yukun Chen and Jiaming Li and Longze Chen and Ze Gong and Jingpeng Li and Zhen Qin and Hengyu Chang and Ancheng Xu and Zhihao Yang and Hamid Alinejad-Rokny and Qiang Qu and Bo Zheng and Min Yang},
      year={2026},
      eprint={2602.21628},
      archivePrefix={arXiv},
      primaryClass={cs.CL},
      url={https://arxiv.org/abs/2602.21628}, 
}

@misc{tian2026autorubric,
      title={Auto-Rubric as Reward: From Implicit Preferences to Explicit Multimodal Generative Criteria}, 
      author={Juanxi Tian and Fengyuan Liu and Jiaming Han and Yilei Jiang and Yongliang Wu and Yesheng Liu and Haodong Li and Furong Xu and Wanhua Li},
      year={2026},
      eprint={2605.08354},
      archivePrefix={arXiv},
      primaryClass={cs.AI},
      url={https://arxiv.org/abs/2605.08354}, 
}

@misc{feng2025rubricrl,
      title={RubricRL: Simple Generalizable Rewards for Text-to-Image Generation}, 
      author={Xuelu Feng and Yunsheng Li and Ziyu Wan and Zixuan Gao and Junsong Yuan and Dongdong Chen and Chunming Qiao},
      year={2025},
      eprint={2511.20651},
      archivePrefix={arXiv},
      primaryClass={cs.CV},
      url={https://arxiv.org/abs/2511.20651}, 
}

@misc{yu2026rdpo,
      title={Visual Preference Optimization with Rubric Rewards}, 
      author={Ya-Qi Yu and Fangyu Hong and Xiangyang Qu and Hao Wang and Gaojie Wu and Qiaoyu Luo and Nuo Xu and Huixin Wang and Wuheng Xu and Yongxin Liao and Zihao Chen and Haonan Li and Ziming Li and Dezhi Peng and Minghui Liao and Jihao Wu and Haoyu Ren and Dandan Tu},
      year={2026},
      eprint={2604.13029},
      archivePrefix={arXiv},
      primaryClass={cs.CV},
      url={https://arxiv.org/abs/2604.13029}, 
}

@misc{gallego2025configurable,
      title={Configurable Preference Tuning with Rubric-Guided Synthetic Data}, 
      author={Víctor Gallego},
      year={2025},
      eprint={2506.11702},
      archivePrefix={arXiv},
      primaryClass={cs.CL},
      url={https://arxiv.org/abs/2506.11702}, 
}

@misc{huang2026Beyond,
      title={Beyond Verifiable Rewards: Rubric-Based GRM for Reinforced Fine-Tuning SWE Agents}, 
      author={Jiawei Huang and Qingping Yang and Renjie Zheng and Jiaze Chen},
      year={2026},
      eprint={2604.16335},
      archivePrefix={arXiv},
      primaryClass={cs.LG},
      url={https://arxiv.org/abs/2604.16335}, 
}

@misc{siro2026GER,
      title={Learning to Judge: LLMs Designing and Applying Evaluation Rubrics}, 
      author={Clemencia Siro and Pourya Aliannejadi and Mohammad Aliannejadi},
      year={2026},
      eprint={2602.08672},
      archivePrefix={arXiv},
      primaryClass={cs.CL},
      url={https://arxiv.org/abs/2602.08672}, 
}

@misc{wei2025conceptbased,
      title={Concept-based Rubrics Improve LLM Formative Assessment and Data Synthesis}, 
      author={Yuchen Wei and Dennis Pearl and Matthew Beckman and Rebecca J. Passonneau},
      year={2025},
      eprint={2504.03877},
      archivePrefix={arXiv},
      primaryClass={cs.LG},
      url={https://arxiv.org/abs/2504.03877}, 
}

@misc{ding2026ripd,
      title={Rubrics as an Attack Surface: Stealthy Preference Drift in LLM Judges}, 
      author={Ruomeng Ding and Yifei Pang and He Sun and Yizhong Wang and Zhiwei Steven Wu and Zhun Deng},
      year={2026},
      eprint={2602.13576},
      archivePrefix={arXiv},
      primaryClass={cs.CR},
      url={https://arxiv.org/abs/2602.13576}, 
}

@misc{jayarao2026explicitreasoning,
      title={Explicit Reasoning Makes Better Judges: A Systematic Study on Accuracy, Efficiency, and Robustness}, 
      author={Pratik Jayarao and Himanshu Gupta and Neeraj Varshney and Chaitanya Dwivedi},
      year={2026},
      eprint={2509.13332},
      archivePrefix={arXiv},
      primaryClass={cs.AI},
      url={https://arxiv.org/abs/2509.13332}, 
}

@misc{ikezogwo2026rubricsfail,
      title={When Rubrics Fail: Error Enumeration as Reward in Reference-Free RL Post-Training for Virtual Try-On}, 
      author={Wisdom Ikezogwo and Mehmet Saygin Seyfioglu and Ranjay Krishna and Karim Bouyarmane},
      year={2026},
      eprint={2603.05659},
      archivePrefix={arXiv},
      primaryClass={cs.CV},
      url={https://arxiv.org/abs/2603.05659}, 
}

@misc{lin2026jade,
      title={JADE: Expert-Grounded Dynamic Evaluation for Open-Ended Professional Tasks}, 
      author={Lanbo Lin and Jiayao Liu and Tianyuan Yang and Li Cai and Yuanwu Xu and Lei Wei and Sicong Xie and Guannan Zhang},
      year={2026},
      eprint={2602.06486},
      archivePrefix={arXiv},
      primaryClass={cs.AI},
      url={https://arxiv.org/abs/2602.06486}, 
}

@misc{zheng2026benchbench,
      title={BenchBench: Benchmarking Automated Benchmark Generation}, 
      author={Yandan Zheng and Haoran Luo and Zhenghong Lin and Wenjin Liu and Luu Anh Tuan},
      year={2026},
      eprint={2603.20807},
      archivePrefix={arXiv},
      primaryClass={cs.CL},
      url={https://arxiv.org/abs/2603.20807}, 
}

@misc{zhao2026reportlogic,
      title={ReportLogic: Evaluating Logical Quality in Deep Research Reports}, 
      author={Jujia Zhao and Zhaoxin Huan and Zihan Wang and Xiaolu Zhang and Jun Zhou and Suzan Verberne and Zhaochun Ren},
      year={2026},
      eprint={2602.18446},
      archivePrefix={arXiv},
      primaryClass={cs.CL},
      url={https://arxiv.org/abs/2602.18446}, 
}

@misc{li2026ueval,
      title={UEval: A Benchmark for Unified Multimodal Generation}, 
      author={Bo Li and Yida Yin and Wenhao Chai and Xingyu Fu and Zhuang Liu},
      year={2026},
      eprint={2601.22155},
      archivePrefix={arXiv},
      primaryClass={cs.CV},
      url={https://arxiv.org/abs/2601.22155}, 
}

@misc{starace2025paperbench,
      title={PaperBench: Evaluating AI's Ability to Replicate AI Research}, 
      author={Giulio Starace and Oliver Jaffe and Dane Sherburn and James Aung and Jun Shern Chan and Leon Maksin and Rachel Dias and Evan Mays and Benjamin Kinsella and Wyatt Thompson and Johannes Heidecke and Amelia Glaese and Tejal Patwardhan},
      year={2025},
      eprint={2504.01848},
      archivePrefix={arXiv},
      primaryClass={cs.AI},
      url={https://arxiv.org/abs/2504.01848}, 
}

@misc{chen2026presentbench,
      title={PresentBench: A Fine-Grained Rubric-Based Benchmark for Slide Generation}, 
      author={Xin-Sheng Chen and Jiayu Zhu and Pei-lin Li and Hanzheng Wang and Shuojin Yang and Meng-Hao Guo},
      year={2026},
      eprint={2603.07244},
      archivePrefix={arXiv},
      primaryClass={cs.CV},
      url={https://arxiv.org/abs/2603.07244}, 
}

@misc{pancholi2026tabxeval,
      title={TabXEval: Why this is a Bad Table? An eXhaustive Rubric for Table Evaluation}, 
      author={Vihang Pancholi and Jainit Bafna and Tejas Anvekar and Manish Shrivastava and Vivek Gupta},
      year={2026},
      eprint={2505.22176},
      archivePrefix={arXiv},
      primaryClass={cs.CL},
      url={https://arxiv.org/abs/2505.22176}, 
}

@misc{yang2026healthscore,
      title={Health-SCORE: Towards Scalable Rubrics for Improving Health-LLMs}, 
      author={Zhichao Yang and Sepehr Janghorbani and Dongxu Zhang and Jun Han and Qian Qian and Andrew Ressler II and Gregory D. Lyng and Sanjit Singh Batra and Robert E. Tillman},
      year={2026},
      eprint={2601.18706},
      archivePrefix={arXiv},
      primaryClass={cs.AI},
      url={https://arxiv.org/abs/2601.18706}, 
}

@misc{shi2026plawbench,
      title={PLawBench: A Rubric-Based Benchmark for Evaluating LLMs in Real-World Legal Practice}, 
      author={Yuzhen Shi and Huanghai Liu and Yiran Hu and Gaojie Song and Xinran Xu and Yubo Ma and Tianyi Tang and Li Zhang and Qingjing Chen and Di Feng and Wenbo Lv and Weiheng Wu and Kexin Yang and Sen Yang and Wei Wang and Rongyao Shi and Yuanyang Qiu and Yuemeng Qi and Jingwen Zhang and Xiaoyu Sui and Yifan Chen and Yi Zhang and An Yang and Bowen Yu and Dayiheng Liu and Junyang Lin and Weixing Shen and Bing Zhao and Charles L. A. Clarke and Hu Wei},
      year={2026},
      eprint={2601.16669},
      archivePrefix={arXiv},
      primaryClass={cs.CL},
      url={https://arxiv.org/abs/2601.16669}, 
}

@misc{yu2026aigrading,
      title={Evaluating AI Grading on Real-World Handwritten College Mathematics: A Large-Scale Study Toward a Benchmark}, 
      author={Zhiqi Yu and Xingping Liu and Haobin Mao and Mingshuo Liu and Long Chen and Jack Xin and Yifeng Yu},
      year={2026},
      eprint={2603.00895},
      archivePrefix={arXiv},
      primaryClass={cs.LG},
      url={https://arxiv.org/abs/2603.00895}, 
}

@misc{indel2026trainingdata,
      title={Training data generation for context-dependent rubric-based short answer grading}, 
      author={Pavel Šindelář and Dávid Slivka and Christopher Bouma and Filip Prášil and Ondřej Bojar},
      year={2026},
      eprint={2603.28537},
      archivePrefix={arXiv},
      primaryClass={cs.CL},
      url={https://arxiv.org/abs/2603.28537}, 
}

@misc{zhang2026fire,
      title={FIRE: A Comprehensive Benchmark for Financial Intelligence and Reasoning Evaluation}, 
      author={Xiyuan Zhang and Huihang Wu and Jiayu Guo and Zhenlin Zhang and Yiwei Zhang and Liangyu Huo and Xiaoxiao Ma and Jiansong Wan and Xuewei Jiao and Yi Jing and Jian Xie},
      year={2026},
      eprint={2602.22273},
      archivePrefix={arXiv},
      primaryClass={cs.AI},
      url={https://arxiv.org/abs/2602.22273}, 
}

@misc{yuksel2026agentic,
      title={Agentic AI for Human Resources: LLM-Driven Candidate Assessment}, 
      author={Kamer Ali Yuksel and Abdul Basit Anees and Ashraf Elneima and Sanjika Hewavitharana and Mohamed Al-Badrashiny and Hassan Sawaf},
      year={2026},
      eprint={2603.26710},
      archivePrefix={arXiv},
      primaryClass={cs.IR},
      url={https://arxiv.org/abs/2603.26710}, 
}

@misc{sun2026comai,
      title={CoMAI: A Collaborative Multi-Agent Framework for Robust and Equitable Interview Evaluation}, 
      author={Gengxin Sun and Ruihao Yu and Liangyi Yin and Yunqi Yang and Bin Zhang and Zhiwei Xu},
      year={2026},
      eprint={2603.16215},
      archivePrefix={arXiv},
      primaryClass={cs.MA},
      url={https://arxiv.org/abs/2603.16215}, 
}

@misc{zhang2026resource,
      title={Resources for Automated Evaluation of Assistive RAG Systems that Help Readers with News Trustworthiness Assessment}, 
      author={Dake Zhang and Mark D. Smucker and Charles L. A. Clarke},
      year={2026},
      eprint={2602.24277},
      archivePrefix={arXiv},
      primaryClass={cs.IR},
      url={https://arxiv.org/abs/2602.24277}, 
}

@misc{winata2025datasheet,
      title={Datasheets Aren't Enough: DataRubrics for Automated Quality Metrics and Accountability}, 
      author={Genta Indra Winata and David Anugraha and Emmy Liu and Alham Fikri Aji and Shou-Yi Hung and Aditya Parashar and Patrick Amadeus Irawan and Ruochen Zhang and Zheng-Xin Yong and Jan Christian Blaise Cruz and Niklas Muennighoff and Seungone Kim and Hanyang Zhao and Sudipta Kar and Kezia Erina Suryoraharjo and M. Farid Adilazuarda and En-Shiun Annie Lee and Ayu Purwarianti and Derry Tanti Wijaya and Monojit Choudhury},
      year={2025},
      eprint={2506.01789},
      archivePrefix={arXiv},
      primaryClass={cs.LG},
      url={https://arxiv.org/abs/2506.01789}, 
}

@misc{watsondaniels2026scrub,
      title={SCRuB: Social Concept Reasoning under Rubric-Based Evaluation}, 
      author={Jamelle Watson-Daniels and Himaghna Bhattacharjee and Skyler Wang and Brandon Handoko and Antonio Li and Anaelia Ovalle and Mahesh Pasupuleti and Candace Ross and Vidya Sarma and Arjun Subramonian and Karen Ullrich and Will van der Vaart and Yijing Xin and Maximilian Nickel},
      year={2026},
      eprint={2605.06444},
      archivePrefix={arXiv},
      primaryClass={cs.AI},
      url={https://arxiv.org/abs/2605.06444}, 
}

@misc{bai2025qwen25vl,
      title={Qwen2.5-VL Technical Report}, 
      author={Shuai Bai and Keqin Chen and Xuejing Liu and Jialin Wang and Wenbin Ge and Sibo Song and Kai Dang and Peng Wang and Shijie Wang and Jun Tang and Humen Zhong and Yuanzhi Zhu and Mingkun Yang and Zhaohai Li and Jianqiang Wan and Pengfei Wang and Wei Ding and Zheren Fu and Yiheng Xu and Jiabo Ye and Xi Zhang and Tianbao Xie and Zesen Cheng and Hang Zhang and Zhibo Yang and Haiyang Xu and Junyang Lin},
      year={2025},
      eprint={2502.13923},
      archivePrefix={arXiv},
      primaryClass={cs.CV},
      url={https://arxiv.org/abs/2502.13923}, 
}

@misc{li2026ares,
      title={ARES: Automated Rubric Synthesis for Scalable LLM Reinforcement Learning}, 
      author={Xiaoyuan Li and Keqin Bao and Moxin Li and Yubo Ma and Yichang Zhang and Wenjie Wang and Fuli Feng and Dayiheng Liu},
      year={2026},
      eprint={2605.23454},
      archivePrefix={arXiv},
      primaryClass={cs.CL},
      url={https://arxiv.org/abs/2605.23454}, 
}

@misc{wang2026mira,
      title={MIRA: Mid-training Rubric Anchoring for Source-Aware Data Selection}, 
      author={Haowen Wang and Yaxin Du and Jian Yang and Jiajun Wu and Shukai Liu and Yuxuan Zhang and Pingjie Wang and Siheng Chen and Tuney Zheng and Ming Zhou and Xianglong Liu and Bryan Dai},
      year={2026},
      eprint={2605.30288},
      archivePrefix={arXiv},
      primaryClass={cs.AI},
      url={https://arxiv.org/abs/2605.30288}, 
}

@misc{qiu2026parl,
      title={Preference-Aware Rubric Learning for Personalized Evaluation}, 
      author={Yilun Qiu and Xiaoyan Zhao and Yang Zhang and Yuxin Chen and Cilin Yan and Jiayin Cai and Xiaolong Jiang and Yao Hu and Yoko Yamakata and Tat-Seng Chua},
      year={2026},
      eprint={2605.31545},
      archivePrefix={arXiv},
      primaryClass={cs.CL},
      url={https://arxiv.org/abs/2605.31545}, 
}

@misc{mei2026DRRubric,
      title={Deep Research as Rubric for Reinforcement Learning}, 
      author={Wangyi Mei and Zhouhong Gu and Zhenhan Bai and Yin Cai and Lefan Zhang and Zhenxin Ding and Bo Chen and Yan Gao and Yi Wu and Yao Hu and Jiaqing Liang and Deqing Yang},
      year={2026},
      eprint={2606.01091},
      archivePrefix={arXiv},
      primaryClass={cs.CL},
      url={https://arxiv.org/abs/2606.01091}, 
}

@misc{zhu2026deeprubric,
      title={DEEPRUBRIC: Evidence-Tree Rubric Supervision for Efficient Reinforcement Learning of Deep Research Agents}, 
      author={Minghang Zhu and Chuyang Wei and Junhao Xu and Yilin Cheng and Zhumin Chen and Jiyan He},
      year={2026},
      eprint={2606.17029},
      archivePrefix={arXiv},
      primaryClass={cs.CL},
      url={https://arxiv.org/abs/2606.17029}, 
}

@misc{zhang2026qubric,
      title={QUBRIC: Co-Designing Queries and Rubrics for RL Beyond Verifiable Rewards}, 
      author={Rongzhi Zhang and Rui Feng and Zhihan Zhang and Jingfeng Yang and Qingyu Yin and Xin Liu and Zixuan Zhang and Priyanka Nigam and Bing Yin and Tuo Zhao and Chao Zhang},
      year={2026},
      eprint={2606.03968},
      archivePrefix={arXiv},
      primaryClass={cs.CL},
      url={https://arxiv.org/abs/2606.03968}, 
}

@misc{xu2026rubricasexpert,
      title={Rubric-as-Experts: Case-Specific MQM Rubrics for Translation Quality Evaluation}, 
      author={Weilu Xu and Yunzhi Shen and Xinye Wang and Ranfei Dang and Shujian Huang},
      year={2026},
      eprint={2606.21559},
      archivePrefix={arXiv},
      primaryClass={cs.CL},
      url={https://arxiv.org/abs/2606.21559}, 
}

@misc{zhang2026rubricstree,
      title={RubricsTree: Scalable and Evolving Open-Ended Evaluation of Personal Health Agents across Health Memory and Medical Skills}, 
      author={Weizhi Zhang and Zechen Li and Hamid Palangi and Ben Graef and A. Ali Heydari and Simon A. Lee and Salman Rahman and Ray Luo and Zeinab Esmaeilpour and Erik Schenck and Chloe Zhang and Yamin Li and Menglian Zhou and Philip S. Yu and Daniel McDuff and Lindsey Sunden and Mark Malhotra and Shwetak Patel and Ahmed A. Metwally},
      year={2026},
      eprint={2606.18203},
      archivePrefix={arXiv},
      primaryClass={cs.CL},
      url={https://arxiv.org/abs/2606.18203}, 
}

@misc{sun2026SVR,
      title={Support Vector Rubrics: Closing the Gap Between Self-Generated and Human Rubrics}, 
      author={Mengyuan Sun and Yu Li and Zhuohao Yu and Shikun Zhang and Wei Ye},
      year={2026},
      eprint={2606.08077},
      archivePrefix={arXiv},
      primaryClass={cs.CL},
      url={https://arxiv.org/abs/2606.08077}, 
}

@misc{yoshida2026feedbacktorubrics,
      title={Feedback-to-Rubrics: Can We Learn Expert Criteria from Inline Comments?}, 
      author={Kotaro Yoshida and So Kuroki and Yuki Imajuku and Taishi Nakamura and Ryunosuke Iwai and Haruki Goda and Takuya Akiba},
      year={2026},
      eprint={2605.29857},
      archivePrefix={arXiv},
      primaryClass={cs.LG},
      url={https://arxiv.org/abs/2605.29857}, 
}

@misc{wang2026generating,
      title={Generating and Refining Dynamic Evaluation Rubrics for LLM-as-a-Judge}, 
      author={Zijie Wang and Eduardo Blanco},
      year={2026},
      eprint={2605.30568},
      archivePrefix={arXiv},
      primaryClass={cs.CL},
      url={https://arxiv.org/abs/2605.30568}, 
}

@misc{jiang2026rubricarrow,
      title={RUBRIC-ARROW: Alternating Pointwise Rubric Reward Modeling for LLM Post-training in Non-verifiable Domains}, 
      author={Haoxiang Jiang and Zihan Dong and Tianci Liu and Wanying Wang and Ran Xu and Tony Yu and Linjun Zhang and Haoyu Wang},
      year={2026},
      eprint={2605.29156},
      archivePrefix={arXiv},
      primaryClass={cs.LG},
      url={https://arxiv.org/abs/2605.29156}, 
}

@misc{zhang2026llmevallogic,
      title={LLMEval-Logic: A Solver-Verified Chinese Benchmark for Logical Reasoning of LLMs with Adversarial Hardening}, 
      author={Ming Zhang and Qiyuan Peng and Yinxi Wei and Yujiong Shen and Kexin Tan and Yuhui Wang and Zhenghao Xiang and Junjie Ye and Zhangyue Yin and Zhiheng Xi and Shihan Dou and Tao Gui and Maxm Pan and Ruizhi Yang and Qi Zhang and Xuanjing Huang},
      year={2026},
      eprint={2605.19597},
      archivePrefix={arXiv},
      primaryClass={cs.CL},
      url={https://arxiv.org/abs/2605.19597}, 
}

@misc{lim2026reliableexpress,
      title={Reliable to Expressive: A Curriculum for Rubric-Following Safety Judges}, 
      author={Yongtaek Lim and Hyeji Choi and Minwoo Kim},
      year={2026},
      eprint={2606.09165},
      archivePrefix={arXiv},
      primaryClass={cs.AI},
      url={https://arxiv.org/abs/2606.09165}, 
}

@misc{kang2026coreact,
      title={Co-ReAct: Rubrics as Step-Level Collaborators for ReAct Agents}, 
      author={Jiazheng Kang and Bowen Zhang and Zixin Song and Jiangwang Chen and Xiao Yang and Da Zhu and Guanjun Jiang},
      year={2026},
      eprint={2605.23590},
      archivePrefix={arXiv},
      primaryClass={cs.AI},
      url={https://arxiv.org/abs/2605.23590}, 
}

@misc{ye2026rubricguidedprocess,
      title={Rubric-Guided Process Reward for Stepwise Model Routing}, 
      author={Shenghao Ye and Yu Guo and Zhengheng Li and Shuangwu Chen and Jian Yang},
      year={2026},
      eprint={2605.29310},
      archivePrefix={arXiv},
      primaryClass={cs.AI},
      url={https://arxiv.org/abs/2605.29310}, 
}

@misc{wang2026learnableassessmentskill,
      title={Learnable Assessment Skills for LLM-based Automated Scoring: Rubric Construction via Iterative Optimization}, 
      author={Yun Wang and Xin Xia and Xuansheng Wu and Xiaoming Zhai and Ninghao Liu},
      year={2026},
      eprint={2605.29274},
      archivePrefix={arXiv},
      primaryClass={cs.CL},
      url={https://arxiv.org/abs/2605.29274}, 
}

@misc{yue2026rubricsexplorat,
      title={Beyond Rubrics: Exploration-Guided Evaluation Skills for Reward Modeling}, 
      author={Xing Yue and Linjuan Wu and Daoxin Zhang and Yongliang Shen and Weiming Lu},
      year={2026},
      eprint={2606.07040},
      archivePrefix={arXiv},
      primaryClass={cs.CL},
      url={https://arxiv.org/abs/2606.07040}, 
}

@misc{long2025vista,
      title={VISTA: A Test-Time Self-Improving Video Generation Agent}, 
      author={Do Xuan Long and Xingchen Wan and Hootan Nakhost and Chen-Yu Lee and Tomas Pfister and Sercan \textsuperscript{\"{O}}. Ar\i k},
      year={2025},
      eprint={2510.15831},
      archivePrefix={arXiv},
      primaryClass={cs.CV},
      url={https://arxiv.org/abs/2510.15831}, 
}

@misc{yan2026DuMate-DeepResearch,
      title={DuMate-DeepResearch: An Auditable Multi-Agent System with Recursive Search and Rubric-Grounded Reasoning}, 
      author={Lingyong Yan and Can Xu and Yukun Zhao and Wenxuan Li and Qingyang Chen and Jiulong Wu and Wenli Song and Xiangnan Li and Weixian Shi and Yiqun Chen and Xuchen Ma and Yuchen Li and Jiashu Zhao and Shuaiqiang Wang and Jianmin Wu and Dawei Yin},
      year={2026},
      eprint={2606.07299},
      archivePrefix={arXiv},
      primaryClass={cs.AI},
      url={https://arxiv.org/abs/2606.07299}, 
}

@misc{yu2026RLR3,
      title={Reinforcement Learning with Robust Rubric Rewards}, 
      author={Ya-Qi Yu and Hao Wang and Fangyu Hong and Xiangyang Qu and Gaojie Wu and Qiaoyu Luo and Nuo Xu and Huixin Wang and Wuheng Xu and Yongxin Liao and Zihao Chen and Haonan Li and Ziming Li and Dezhi Peng and Minghui Liao and Jihao Wu and Haoyu Ren and Dandan Tu},
      year={2026},
      eprint={2605.30244},
      archivePrefix={arXiv},
      primaryClass={cs.CV},
      url={https://arxiv.org/abs/2605.30244}, 
}

@misc{lin2026longtracerl,
      title={LongTraceRL: Learning Long-Context Reasoning from Search Agent Trajectories with Rubric Rewards}, 
      author={Nianyi Lin and Jiajie Zhang and Lei Hou and Juanzi Li},
      year={2026},
      eprint={2605.31584},
      archivePrefix={arXiv},
      primaryClass={cs.CL},
      url={https://arxiv.org/abs/2605.31584}, 
}

@misc{rezaei2026RGSD,
      title={Rubric-Guided Self-Distillation: Post-Training Without Rubric Verifiers}, 
      author={MohammadHossein Rezaei and Anas Mahmoud and Zihao Wang and Utkarsh Tyagi and Advait Gosai and Razvan-Gabriel Dumitru and Aakash Sabharwal and Bing Liu and Yunzhong He},
      year={2026},
      eprint={2606.12507},
      archivePrefix={arXiv},
      primaryClass={cs.LG},
      url={https://arxiv.org/abs/2606.12507}, 
}

@misc{gu2026RCSD,
      title={Rethinking Reward Supervision: Rubric-Conditioned Self-Distillation}, 
      author={Siyi Gu and Jialin Chen and Sophia Zhou and Arman Cohan and Rex Ying},
      year={2026},
      eprint={2606.19327},
      archivePrefix={arXiv},
      primaryClass={cs.AI},
      url={https://arxiv.org/abs/2606.19327}, 
}

@misc{wang2026CHERRL,
      title={Reproducing, Analyzing, and Detecting Reward Hacking in Rubric-Based Reinforcement Learning}, 
      author={Xuekang Wang and Zhuoyuan Hao and Shuo Hou and Hao Peng and Juanzi Li and Xiaozhi Wang},
      year={2026},
      eprint={2606.04923},
      archivePrefix={arXiv},
      primaryClass={cs.LG},
      url={https://arxiv.org/abs/2606.04923}, 
}

@misc{loye2026RUBAS,
      title={RUBAS: Rubric-Based Reinforcement Learning for Agent Safety}, 
      author={Xian Qi Loye and Qinglin Su and Zhexin Zhang and Shiyao Cui and Qi Zhu and Fei Mi and Hongning Wang and Minlie Huang},
      year={2026},
      eprint={2606.04051},
      archivePrefix={arXiv},
      primaryClass={cs.LG},
      url={https://arxiv.org/abs/2606.04051}, 
}

@misc{kao2026AutoRubric-T2I,
      title={AutoRubric-T2I: Robust Rule-Based Reward Model for Text-to-Image Alignment}, 
      author={Kuei-Chun Kao and Daixuan Huo and Yuanhao Ban and Cho-Jui Hsieh},
      year={2026},
      eprint={2605.17602},
      archivePrefix={arXiv},
      primaryClass={cs.AI},
      url={https://arxiv.org/abs/2605.17602}, 
}

@misc{li2026anyaudiojudge,
      title={AnyAudio-Judge: A Dynamic Rubric-Based Benchmark and Evaluator for Audio Instruction Following}, 
      author={Haitao Li and Tian Tan and Yuguang Yang and Shan Yang and Xie Chen},
      year={2026},
      eprint={2606.03116},
      archivePrefix={arXiv},
      primaryClass={eess.AS},
      url={https://arxiv.org/abs/2606.03116}, 
}

@misc{guan2026evorubrics,
      title={EvoRubric: Self-Evolving Rubric-Driven RL for Open-Ended Generation}, 
      author={Xin Guan and Xiaomeng Hu and Shen Huang and Zhenyi Wang and Bo Zhang and Zijian Li and Pengjun Xie and Bo Liu and Jiuxin Cao},
      year={2026},
      eprint={2605.29847},
      archivePrefix={arXiv},
      primaryClass={cs.CL},
      url={https://arxiv.org/abs/2605.29847}, 
}

@misc{liu2026arbor,
      title={ARBOR: Online Process Rewards via a Reusable Rubric Buffer for Search Agents}, 
      author={Zheng Liu and Longxiang Zhang and Xintong Wang and Zhiang Xu and Shaoxiong Zhan and Xin Shan and Wen Huang and Tao Dai and Shu-Tao Xia and Chengfu Huo and Liang Ding},
      year={2026},
      eprint={2606.03239},
      archivePrefix={arXiv},
      primaryClass={cs.CL},
      url={https://arxiv.org/abs/2606.03239}, 
}

@misc{tian2026arco,
      title={ARCO: Adaptive Rubric with Co-Evolution for Multi-Step LLM-Based Agents}, 
      author={Zihang Tian and Jingsen Zhang and Rui Li and Xiaohe Bo and Yuanzi Li and Xu Chen},
      year={2026},
      eprint={2606.21262},
      archivePrefix={arXiv},
      primaryClass={cs.AI},
      url={https://arxiv.org/abs/2606.21262}, 
}

@misc{lv2026gear,
      title={Mitigating False Credit Propagation: Probabilistic Graphical Reward Aggregation for Rubric-Based Reinforcement Learning}, 
      author={Can Lv and Mingju Chen and Heng Chang and Shiji Zhou},
      year={2026},
      eprint={2606.03361},
      archivePrefix={arXiv},
      primaryClass={cs.LG},
      url={https://arxiv.org/abs/2606.03361}, 
}

@misc{huang2026focalreward,
      title={Focal Reward: Balanced Reinforcement Learning under Rubric-Based Rewards}, 
      author={Yu Huang and Zihua Zhao and Zhaoxin Huan and Wanli Gu and Feng Hong and Xinmu Ge and Lin Yuan and Weichang Wu and Qiang Hu and Xiaolu Zhang and Jun Zhou and Jiangchao Yao},
      year={2026},
      eprint={2605.26579},
      archivePrefix={arXiv},
      primaryClass={cs.LG},
      url={https://arxiv.org/abs/2605.26579}, 
}

@misc{tyagi2026pow3r,
      title={Not Every Rubric Teaches Equally: Policy-Aware Rubric Rewards for RLVR}, 
      author={Utkarsh Tyagi and Xingang Guo and MohammadHossein Rezaei and Daniel George and Anas Mahmoud and Jackson Lee and Bing Liu and Yunzhong He},
      year={2026},
      eprint={2605.20164},
      archivePrefix={arXiv},
      primaryClass={cs.AI},
      url={https://arxiv.org/abs/2605.20164}, 
}

@misc{yang2026tournamentgrpo,
      title={Tournament-GRPO: Group-Wise Tournament Rewards for Reinforcement Learning in Open-Ended Long-Form Generation}, 
      author={Zixuan Yang and Yiqun Chen and Wei Yang and Erhan Zhang and Zihan Shen and Xiaochi Wei and Yan Gao and Yi Wu and Yao Hu and Jiaxin Mao},
      year={2026},
      eprint={2605.26958},
      archivePrefix={arXiv},
      primaryClass={cs.CL},
      url={https://arxiv.org/abs/2605.26958}, 
}

@misc{yang2026judgmentbench,
      title={JudgmentBench: Comparing Rubric and Preference Evaluation for Quality Assessment}, 
      author={Russell Yang and Ruishi Chen and Pierce Kelaita and Riya Ranjan and Sibo Ma and Charles Dickens and Matthew Guillod and Megan Ma and Julian Nyarko},
      year={2026},
      eprint={2605.25240},
      archivePrefix={arXiv},
      primaryClass={cs.CL},
      url={https://arxiv.org/abs/2605.25240}, 
}

@misc{xu2026lpeval,
      title={LP-Eval: Rubric and Dataset for Measuring the Quality of Legal Proposition Generation}, 
      author={Shanshan Xu and Johan Lindholm and Amogh Raina and Henrik Palmer Olsen and Daniel Hershcovich},
      year={2026},
      eprint={2605.19815},
      archivePrefix={arXiv},
      primaryClass={cs.CL},
      url={https://arxiv.org/abs/2605.19815}, 
}

@misc{chen2026lexrubric,
      title={LexRubric: A Rubric-Guided Diagnostic Benchmark for Open-Ended Legal Tasks}, 
      author={Yifan Chen and Haitao Li and Yiran Hu and Kaisong Song and Jun Lin and Yueyue Wu and Qingyao Ai and Min Zhang and Yiqun Liu},
      year={2026},
      eprint={2606.09389},
      archivePrefix={arXiv},
      primaryClass={cs.CL},
      url={https://arxiv.org/abs/2606.09389}, 
}

@misc{wang2026bigfinancebench,
      title={BigFinanceBench: A Workflow-Grounded Benchmark for Financial-Research Agents}, 
      author={Alex Wang and Georg Meinhardt and Jacob Katz and Joseph H. Kim and Pratyush K. Chaudhary and Chase Blagden and Eric Xu},
      year={2026},
      eprint={2606.03829},
      archivePrefix={arXiv},
      primaryClass={cs.AI},
      url={https://arxiv.org/abs/2606.03829}, 
}

@misc{zhao2026pancanbench,
      title={PanCanBench: A Comprehensive Benchmark for Evaluating Large Language Models in Pancreatic Oncology}, 
      author={Yimin Zhao and Sheela R. Damle and Simone E. Dekker and Scott Geng and Karly Williams Silva and Jesse J Hubbard and Manuel F Fernandez and Fatima Zelada-Arenas and Alejandra Alvarez and Brianne Flores and Alexis Rodriguez and Stephen Salerno and Carrie Wright and Zihao Wang and Pang Wei Koh and Jeffrey T. Leek},
      year={2026},
      eprint={2603.01343},
      archivePrefix={arXiv},
      primaryClass={cs.CL},
      url={https://arxiv.org/abs/2603.01343}, 
}

@misc{ma2026mmae,
      title={MMAE: A Massive Multitask Audio Editing Benchmark}, 
      author={Ziyang Ma and Ruiqi Yan and Ruiyang Xu and Jie Fang and Zhikang Niu and Yi-Wen Chao and Wenming Tu and Tianrui Wang and Auden and Qi Chen and Wenxi Chen and Jiaying Chi and Yanru Huo and Zixuan Jiang and Xiquan Li and Yalin Li and Junxi Liu and Minghao Liu and Binghao Qiang and Yijia Shan and Zheshu Song and Tian Tan and Zixiang Wang and Zeyu Xie and Zhifei Xie and Xiaoyu Xing and Qixiang Xu and Chen Yang and Guanrou Yang and Shan Yang and Yifan Yang and Steve Yves and Haotian Zhang and Haina Zhu and Kai Yu and Liefeng Bo and Eng-Siong Chng and Xie Chen},
      year={2026},
      eprint={2606.07229},
      archivePrefix={arXiv},
      primaryClass={cs.SD},
      url={https://arxiv.org/abs/2606.07229}, 
}

@misc{ahmadi2026improvingheart,
      title={Improving Heart-Focused Medical Question Answering in LLMs via Variance-Aware Rubric Rewards with GRPO}, 
      author={Arash Ahmadi and Parisa Masnadi and Sarah Sharif and Charles Nicholson and David Ebert and Mike Banad},
      year={2026},
      eprint={2606.05174},
      archivePrefix={arXiv},
      primaryClass={cs.CL},
      url={https://arxiv.org/abs/2606.05174}, 
}

@misc{zhao2026reccbm,
      title={REC-CBM: Rubric-Aware Error-Correction Concept Bottleneck Models for Trustworthy Open-Ended Grading}, 
      author={Chengshuai Zhao and Fan Zhang and Kumar Satvik Chaudhary and Yiwen Li and Lo Pang-Yun Ting and Ying-Chih Chen and Huan Liu},
      year={2026},
      eprint={2605.27402},
      archivePrefix={arXiv},
      primaryClass={cs.CY},
      url={https://arxiv.org/abs/2605.27402}, 
}

@misc{benhenda2026ipofinance,
      title={IPO Finance Agent: Evaluation of LLM Financial Analysts beyond Finance Agent v2, with Automated Rubric Generation -- the Case of the SpaceX (SPCX) IPO}, 
      author={Mostapha Benhenda},
      year={2026},
      eprint={2606.23032},
      archivePrefix={arXiv},
      primaryClass={cs.AI},
      url={https://arxiv.org/abs/2606.23032}, 
}

@misc{yang2026merit,
      title={MERIT: Matching Expertise via Rubric-Informed Training for Reviewer Assignment}, 
      author={Zixuan Yang and Yibo Zhao and Weicong Liu and Xiang Li},
      year={2026},
      eprint={2605.27865},
      archivePrefix={arXiv},
      primaryClass={cs.CL},
      url={https://arxiv.org/abs/2605.27865}, 
}

@misc{long2026a2rd,
      title={A$^2$RD: Agentic Autoregressive Diffusion for Long Video Consistency}, 
      author={Do Xuan Long and Yale Song and Min-Yen Kan and Tomas Pfister and Long T. Le},
      year={2026},
      eprint={2605.06924},
      archivePrefix={arXiv},
      primaryClass={cs.CV},
      url={https://arxiv.org/abs/2605.06924}, 
}

@misc{mehta2026complexconstrain,
      title={ComplexConstraints and Beyond: Expert Rubrics for RLVR}, 
      author={Sushant Mehta and Liudas Panavas and Suhaas Garre and Edwin Chen},
      year={2026},
      eprint={2606.09118},
      archivePrefix={arXiv},
      primaryClass={cs.AI},
      url={https://arxiv.org/abs/2606.09118}, 
}

@misc{wei2026perception,
      title={PerceptionRubrics: Calibrating Multimodal Evaluation to Human Perception}, 
      author={Yana Wei and Hongbo Peng and Yanlin Lai and Liang Zhao and Kangheng Lin and En Yu and Keyu Lv and Han Zhou and Yin Tang and Haodong Li and Mitt Huang and Hangyu Guo and Jianjian Sun and Zheng Ge and Xiangyu Zhang and Daxin Jiang and Vishal M. Patel},
      year={2026},
      eprint={2606.28322},
      archivePrefix={arXiv},
      primaryClass={cs.CV},
      url={https://arxiv.org/abs/2606.28322}, 
}

@misc{ding2026evorubrics,
      title={EvoRubrics: Dynamic Rubrics as Rewards via Adversarial Co-Evolution for LLM Reinforcement Learning}, 
      author={Hongxin Ding and Baixiang Huang and Yue Fang and Weibin Liao and Zheng Li and Jinyang Zhang and Zhijing Wu and Junfeng Zhao and Yasha Wang},
      year={2026},
      eprint={2606.23038},
      archivePrefix={arXiv},
      primaryClass={cs.LG},
      url={https://arxiv.org/abs/2606.23038}, 
}

@misc{peng2026llmasajudge,
      title={Can LLM-as-a-Judge Reliably Verify Rubrics in Agentic Scenarios?}, 
      author={Yangda Peng and Yunjia Qi and Hao Peng and Haotian Xia and Guanzhong He and Xintong Shi and Richeng Xuan and Songyuanyi Lu and Yixian Liu and Zhichao Hu and Yuhong Liu and Lei Hou and Bin Xu and Juanzi Li},
      year={2026},
      eprint={2606.29920},
      archivePrefix={arXiv},
      primaryClass={cs.CL},
      url={https://arxiv.org/abs/2606.29920}, 
}

@misc{gandhi2026ppteval,
      title={PPT-Eval: A Benchmark for Computer-Use Agents on PowerPoint Tasks}, 
      author={Apurva Gandhi and Vishwas Suryanarayanan and Raja Hasnain Anwar and Firoz Shaik and Shubhang Desai and Thong Q. Nguyen and Muhammad Taqi Raza and Vishal Chowdhary and Graham Neubig},
      year={2026},
      eprint={2606.31154},
      archivePrefix={arXiv},
      primaryClass={cs.LG},
      url={https://arxiv.org/abs/2606.31154}, 
}

\end{document}